\documentclass[10pt,journal,compsoc]{IEEEtran}
\ifCLASSOPTIONcompsoc
  \usepackage[nocompress]{cite}
\else
  \usepackage{cite}
\fi

\ifCLASSINFOpdf
  \usepackage[pdftex]{graphicx}
\else
  \usepackage[dvips]{graphicx}
\fi

\usepackage{amsmath}
\usepackage{amssymb}
\usepackage{amsfonts} 

\ifCLASSOPTIONcompsoc
  \usepackage[caption=false,font=footnotesize,labelfont=sf,textfont=sf]{subfig}
\else
  \usepackage[caption=false,font=footnotesize]{subfig}
\fi

\usepackage{url}
\usepackage{color}

\begin{document}
\title{A Background-Agnostic Framework with Adversarial Training for Abnormal Event Detection in Video}
%
%

\author{Mariana-Iuliana~Georgescu,
        {Radu Tudor}~Ionescu,~\IEEEmembership{Member,~IEEE,}
        {Fahad Shahbaz}~Khan,~\IEEEmembership{Member,~IEEE,}
        Marius~Popescu
        and~Mubarak~Shah,~\IEEEmembership{Fellow,~IEEE}

\IEEEcompsocitemizethanks{\IEEEcompsocthanksitem M.I. Georgescu, R.T. Ionescu and M. Popescu are with SecurifAI and the Department
of Computer Science, University of Bucharest, Romania. R.T. Ionescu is also affiliated with the Romanian Young Academy.\protect\\
E-mail: raducu.ionescu@gmail.com
\IEEEcompsocthanksitem F.S. Khan is with Mohamed bin Zayed University of Artificial Intelligence (MBZUAI), UAE.
\IEEEcompsocthanksitem M. Shah is with the Center for Research in Computer Vision (CRCV), Department of Computer Science, University of Central Florida, Orlando, FL, 32816.}
\thanks{Manuscript received April 19, 2005; revised August 26, 2015.}}

%
%

\markboth{IEEE Transactions on Pattern Analysis and Machine Intelligence,~Vol.~14, No.~8, August~2015}%
{Georgescu \MakeLowercase{\textit{et al.}}: Object-Centric Framework with Adversarial Training for Abnormal Event Detection}


\IEEEtitleabstractindextext{%
\begin{abstract}
Abnormal event detection in video is a complex computer vision problem that has attracted significant attention in recent years.
The complexity of the task arises from the commonly-adopted definition of an abnormal event, that is, a rarely occurring event that typically depends on the surrounding context. Following the standard formulation of abnormal event detection as outlier detection, we propose a background-agnostic framework that learns from training videos containing only normal events. Our framework is composed of an object detector, a set of appearance and motion auto-encoders, and a set of classifiers. Since our framework only looks at object detections, it can be applied to different scenes, provided that normal events are defined identically across scenes and that the single main factor of variation is the background. This makes our method background agnostic, as we rely strictly on objects that can cause anomalies, and not on the background.
To overcome the lack of abnormal data during training, we propose an adversarial learning strategy for the auto-encoders. We create a scene-agnostic set of out-of-domain pseudo-abnormal examples, which are correctly reconstructed by the auto-encoders before applying gradient ascent on the pseudo-abnormal examples. We further utilize the pseudo-abnormal examples to serve as abnormal examples when training appearance-based and motion-based binary classifiers to discriminate between normal and abnormal latent features and reconstructions. Furthermore, to ensure that the auto-encoders focus only on the main object inside each bounding box image, we introduce a branch that learns to segment the main object. We compare our framework with the state-of-the-art methods on four benchmark data sets, using various evaluation metrics. Compared to existing methods, the empirical results indicate that our approach achieves favorable performance on all data sets. In addition, we provide region-based and track-based annotations for two large-scale abnormal event detection data sets from the literature, namely ShanghaiTech and Subway.
\end{abstract}
}

\setlength{\abovedisplayskip}{3.0pt}
\setlength{\belowdisplayskip}{3.0pt}

\maketitle

\IEEEdisplaynontitleabstractindextext

\IEEEpeerreviewmaketitle

\IEEEraisesectionheading{\section{Introduction}\label{sec:introduction}}

\IEEEPARstart{A}{bnormal} events are defined as rare occurrences that deviate from the normal patterns observed in familiar events \cite{Ramachandra-PAMI-2020}. Considering the prior work on video anomaly detection \cite{Antic-ICCV-2011,Cheng-CVPR-2015,Dong-Access-2020,Hasan-CVPR-2016,Ionescu-CVPR-2019,Ionescu-WACV-2019,Kim-CVPR-2009,Lee-TIP-2019,Li-PAMI-2014,Liu-CVPR-2018,Lu-ICCV-2013,Luo-ICCV-2017,Mahadevan-CVPR-2010,Mehran-CVPR-2009,Park-CVPR-2020,Ramachandra-PAMI-2020,Ravanbakhsh-WACV-2018,Tang-PRL-2020,Wu-TNNLS-2019,Xu-CVIU-2017,Zhao-CVPR-2011,Zhang-PR-2020}, we can devise a non-exhaustive taxonomy of abnormal events that includes: appearance anomalies (for example, a car or a truck in a pedestrian area), short-term motion anomalies (for example, a person running or a person throwing an object), long-term motion anomalies (for instance, loitering) and group anomalies (for example, several people running inside a public building). These anomalies, which are commonly observed in publicly available data sets \cite{Adam-PAMI-2008,Lu-ICCV-2013,Luo-ICCV-2017,Mahadevan-CVPR-2010}, typically involve people and vehicles, which usually appear in surveillance videos captured in urban areas. This is because abnormal event detection is studied in the context of video surveillance.
Additionally, we should emphasize that the classification of an event as normal or abnormal always depends on the context. 
For instance, driving a truck on the street is considered normal, but, if the truck enters a pedestrian area, the event becomes abnormal. Considering the commonly-adopted definition of abnormal events and the reliance on context, it is difficult to obtain a sufficiently representative set of anomalies for all possible contexts, making traditional supervised methods less applicable to abnormal event detection. Therefore, the majority of anomaly detection methods proposed so far are based on outlier detection~\cite{Antic-ICCV-2011,Cheng-CVPR-2015,Dong-Access-2020,Hasan-CVPR-2016,Ionescu-CVPR-2019,Ionescu-WACV-2019,Kim-CVPR-2009,Lee-TIP-2019,Li-PAMI-2014,Liu-CVPR-2018,Lu-ICCV-2013,Luo-ICCV-2017,Mahadevan-CVPR-2010,Mehran-CVPR-2009,Park-CVPR-2020,Ramachandra-WACV-2020a,Ramachandra-WACV-2020b,Ravanbakhsh-WACV-2018,Ravanbakhsh-ICIP-2017,Sabokrou-IP-2017,Tang-PRL-2020,Wu-TNNLS-2019,Xu-CVIU-2017,Zhao-CVPR-2011,Zhang-PR-2020}, learning normality models from training videos containing only normal events. During inference, an event is labeled as abnormal if it deviates from the normality model. In general, existing abnormal event detection methods build the normality model using local features~\cite{Dutta-AAAI-2015,Ionescu-WACV-2019,Kim-CVPR-2009,Lu-ICCV-2013,Luo-ICCV-2017,Mahadevan-CVPR-2010,Sabokrou-IP-2017,Saligrama-CVPR-2012,Wu-TNNLS-2019,Zhang-PR-2016,Zhang-PR-2020}, global (frame-level) features~\cite{Dong-Access-2020,Lee-TIP-2019,Liu-CVPR-2018,Mehran-CVPR-2009,Park-CVPR-2020,Ramachandra-WACV-2020a,Ramachandra-WACV-2020b,Ravanbakhsh-WACV-2018,Ravanbakhsh-ICIP-2017,Smeureanu-ICIAP-2017,Tang-PRL-2020}, or both~\cite{Cheng-CVPR-2015,Cong-CVPR-2011,Hasan-CVPR-2016}. Such methods work well when training and testing are conducted on the same scene.
If we switch to a different scene at inference time, however, the methods based on local or global features tend to fail because the features used for the normality model are specific to the training scene. There are a few approaches that do not require any training data~\cite{Giorno-ECCV-2016,Ionescu-ICCV-2017,Liu-BMVC-2018,Pang-CVPR-2020}, instead employing different algorithms for change detection at test time.
Such methods can be considered scene-agnostic, but they typically obtain much lower performance levels compared to methods that rely on training data to build normality models. Other works, such as \cite{Sultani-CVPR-2018}, tackled frame anomaly detection as an action recognition task, considering only events that are always abnormal, irrespective of the context, e.g. arson attacks, burglaries or traffic accidents. By considering only generic abnormal events, the method developed by Sultani et al.~\cite{Sultani-CVPR-2018} is implicitly scene-agnostic. However, the method falls outside the commonly-accepted definition of abnormal events and is rather considered by others~\cite{Ramachandra-PAMI-2020} as an action recognition method.

A handful of methods \cite{Doshi-CVPRW-2020a,Doshi-CVPRW-2020b,Ionescu-CVPR-2019} also have the potential to become scene-agnostic, while still taking into account the reliance on context of anomalies. This is achieved by applying an object detector before feature extraction, allowing the model to learn the normality only with respect to the objects, while ignoring the background or other elements in the scene. However, such methods are affected by significant viewpoint changes, frame resolution differences, frame rate variations and different types of normal events across scenes. Such methods work well across different scenes, especially in case of background variations. Similar to the above mentioned works \cite{Doshi-CVPRW-2020a,Doshi-CVPRW-2020b,Ionescu-CVPR-2019}, we employ an object detector, analyzing the abnormality at the object level by extracting appearance or motion features to represent each object.
Different from preliminary works that rely on object detection \cite{Doshi-CVPRW-2020a,Doshi-CVPRW-2020b,Ionescu-CVPR-2019}, 
we are the first to present a cross-database (cross-domain) evaluation, demonstrating that our proposed method is not severely affected by scene variations. Remarkably, our cross-domain results on Avenue~\cite{Lu-ICCV-2013}, ShanghaiTech~\cite{Luo-ICCV-2017}, Subway~\cite{Adam-PAMI-2008} and UCSD Ped2~\cite{Mahadevan-CVPR-2010} surpass many of the recently reported in-domain results (see Section~\ref{sec_crossdb}).



We conduct experiments on four challenging benchmarks, namely Avenue~\cite{Lu-ICCV-2013}, ShanghaiTech~\cite{Luo-ICCV-2017}, Subway~\cite{Adam-PAMI-2008} and UCSD Ped2~\cite{Mahadevan-CVPR-2010}, reporting favorable performance levels compared to the state-of-the-art methods~\cite{Giorno-ECCV-2016,Dong-Access-2020,Doshi-CVPRW-2020a,Doshi-CVPRW-2020b,Hasan-CVPR-2016,Ionescu-ICCV-2017,Ionescu-CVPR-2019,Ionescu-WACV-2019,Lee-TIP-2019,Lee-ICASSP-2018,Liu-CVPR-2018,Liu-BMVC-2018,Lu-ICCV-2013,Luo-ICCV-2017,Nguyen-ICCV-2019,Pang-CVPR-2020,Park-CVPR-2020,Ramachandra-WACV-2020a,Ramachandra-WACV-2020b,Smeureanu-ICIAP-2017,Sultani-CVPR-2018,Tang-PRL-2020,Wu-TNNLS-2019,Kim-CVPR-2009, Mehran-CVPR-2009, Mahadevan-CVPR-2010,Zhang-PR-2016, Ravanbakhsh-ICIP-2017, Xu-CVIU-2017, Ravanbakhsh-WACV-2018, Gong-ICCV-2019, Ji-IJCNN-2020, Lu-ECCV-2020, Sun-ACMMM-2020, Wang-ACMMM-2020, Yu-ACMMM-2020, Zaheer-CVPR-2020}. While we report results in terms of the standard frame-level area under the curve (AUC) metric, we also report our performance levels in terms of the Region-Based Detection Criterion (RBDC) and Track-Based Detection Criterion (TBDC).
These criteria were recently introduced by Ramachandra et al.~\cite{Ramachandra-WACV-2020a}, who argued that the frame-level AUC is inadequate to fully evaluate abnormal event detection systems, essentially because it does not take into consideration spatial localization, counting a frame as a correct detection even when the pixels predicted as abnormal do not overlap with the ground-truth abnormal pixels. Since Ramachandra et al.~\cite{Ramachandra-WACV-2020a} did not provide the region-based and track-based annotations required to compute RBDC and TBDC on ShanghaiTech and Subway, we labeled these data sets ourselves. We release the annotations along with our open source code at: \url{https://github.com/lilygeorgescu/AED}.

\noindent
\textbf{Relation to preliminary CVPR 2019 version~\cite{Ionescu-CVPR-2019}.}
Since our method stems from the method proposed in~\cite{Ionescu-CVPR-2019}, we briefly present our preliminary work and explain our new design changes. Ionescu et al.~\cite{Ionescu-CVPR-2019} was the first work to propose an object-centric framework, employing a single-stage detection framework (SSD \cite{Liu-ECCV-2016} with Feature Pyramid Networks (FPN) \cite{Lin-CVPR-2017}) on each frame in order to extract objects of interest. Representative deep unsupervised features for normal objects are learned using three convolutional auto-encoders (CAEs), one for appearance and two for motion. Upon training the auto-encoders, the concatenated latent features are clustered using the k-means algorithm to obtain clusters representing various types of normality. For each normality cluster, an SVM classifier is trained to discriminate the corresponding cluster from the rest, using the one-versus-rest scheme. During inference, 
the one-versus-rest SVM is applied to obtain a normality score for each detected object. The maximum among the scores assigned by the one-versus-rest SVM with respect to each normality cluster is the normality score for a given test sample. While our preliminary framework attained state-of-the-art results at the time of publication~\cite{Ionescu-CVPR-2019}, we observed that the auto-encoders sometimes produce excessively good reconstructions for abnormal examples, resulting in a higher false negative rate. Furthermore, the previous framework contains multiple components that are not integrated into an end-to-end pipeline. In order to adopt an end-to-end processing pipeline in our current work, we remove the k-means clustering and the one-versus-rest classification steps, replacing them with three binary classifiers (two for motion and one for appearance) that are trained to discriminate between normal examples from the training set and pseudo-abnormal examples from a generic data set. Additionally, we make significant changes to the auto-encoders. First, we introduce skip connections and two decoder branches, one for adversarial training and one for object segmentation. To cope with the lack of anomalies during training, we introduce a generic set of out-of-domain data samples that play the role of abnormal samples, which we call ``pseudo-abnormal''. The pseudo-abnormal examples are database-agnostic, meaning that they can be used for any type of scene, as demonstrated throughout our experiments. Indeed, we utilize the same pseudo-abnormal examples for every data set that we experiment with. We emphasize that the auto-encoders are not supposed to reconstruct the pseudo-abnormal examples, i.e.~we expect good reconstructions for normal data samples only. To address the issue regarding the excessively good reconstructions for abnormal examples observed in the preliminary framework~\cite{Ionescu-CVPR-2019}, we perform adversarial training on the pseudo-abnormal examples to prevent the auto-encoders from generalizing to such data samples, inherently inducing the same behavior for abnormal examples. Indeed, adversarial training helps us to obtain poor reconstructions for abnormal objects, such as bicycles or cars in pedestrian areas. 
One last difference from our previous work~\cite{Ionescu-CVPR-2019} is to replace image gradients with optical flow. In summary, we propose the following changes with respect to our previous work~\cite{Ionescu-CVPR-2019}: $(i)$ to replace the k-means clustering step with binary classifiers that distinguish between normal and
pseudo-abnormal objects, $(ii)$ to integrate skip connections, $(iii)$ to add a segmentation decoder, $(iv)$ to perform adversarial training and $(v)$ to use optical flow instead of image gradients. We emphasize that the proposed changes lead to a significantly different model, bringing consistently superior performance over the preliminary work~\cite{Ionescu-CVPR-2019}, especially in terms of the recently-introduced RBDC and TBDC metrics~\cite{Ramachandra-WACV-2020a}.

In summary, our contribution is threefold:
\vspace{-0.18cm}
\begin{itemize}
    \item We propose a novel framework that leverages adversarial training in the context of abnormal event detection in video.
    \item We show that our framework is background agnostic by performing a series of cross-database experiments.
    \item We provide region-level and track-level annotations for ShanghaiTech and Subway, allowing future works to report results in terms of RBDC and TBDC.
\end{itemize}
\vspace{-0.18cm}

The rest of this paper is organized as follows. We present related work on abnormal event detection in video in Section~\ref{sec_RelatedWork}. Our method is described in detail in Section~\ref{sec_Method}. We present the anomaly detection experiments and results in Section~\ref{sec_Experiments}. Finally, our conclusions are drawn in Section~\ref{sec_Conclusion}.

\vspace{-0.2cm}
\section{Related Work}
\label{sec_RelatedWork}

Most of the recent works treat abnormal event detection as an outlier detection task~\cite{Antic-ICCV-2011,Cheng-CVPR-2015,Dong-Access-2020,Hasan-CVPR-2016,Ionescu-WACV-2019,Kim-CVPR-2009,Lee-TIP-2019,Li-PAMI-2014,Liu-CVPR-2018,Lu-ICCV-2013,Luo-ICCV-2017,Mahadevan-CVPR-2010,Mehran-CVPR-2009,Park-CVPR-2020,Ramachandra-WACV-2020a,Ramachandra-WACV-2020b,Ravanbakhsh-WACV-2018,Ravanbakhsh-ICIP-2017,Sabokrou-IP-2017,Tang-PRL-2020,Wu-TNNLS-2019,Xu-CVIU-2017,Zhao-CVPR-2011,Zhang-PR-2020, Dutta-AAAI-2015, Zhang-PR-2016, Cong-CVPR-2011, Ren-BMVC-2015}, learning a model using only normal data. Then, at inference time, the events that diverge from the normality model are labeled as abnormal. Existing abnormal event detection methods can be categorized into distance-based approaches \cite{Ionescu-CVPR-2019,Ionescu-WACV-2019,Ramachandra-WACV-2020a,Ramachandra-WACV-2020b,Ravanbakhsh-WACV-2018,Sabokrou-IP-2017,Sabokrou-CVIU-2018,Saligrama-CVPR-2012,Smeureanu-ICIAP-2017,Sun-PR-2017,Tran-BMVC-2017}, reconstruction-based models \cite{Cong-CVPR-2011,Gong-ICCV-2019,Hasan-CVPR-2016,Liu-CVPR-2018,Lu-ICCV-2013,Luo-ICCV-2017,Nguyen-ICCV-2019,Park-CVPR-2020,Ravanbakhsh-ICIP-2017,Tang-PRL-2020}, probabilistic models \cite{Adam-PAMI-2008,Antic-ICCV-2011,Cheng-CVPR-2015,Feng-NC-2017,Hinami-ICCV-2017,Kim-CVPR-2009,Mahadevan-CVPR-2010,Mehran-CVPR-2009,Wu-CVPR-2010} and change detection methods \cite{Giorno-ECCV-2016,Ionescu-ICCV-2017,Liu-BMVC-2018,Pang-CVPR-2020}. 
A handful of preliminary works \cite{Cheng-CVPR-2015, Cong-CVPR-2011, Dutta-AAAI-2015, Lu-ICCV-2013, Ren-BMVC-2015} proposed to build a dictionary of atoms representing normal events, labeling the events that are not represented in the dictionary as abnormal. For example, Dutta et al.~\cite{Dutta-AAAI-2015} proposed an approach that builds a model of familiar events from training data using a sparse coding objective. Then, the model is incrementally updated in an unsupervised manner as new patterns are observed in the test data.

Other recent approaches \cite{Hasan-CVPR-2016, Hinami-ICCV-2017, Liu-CVPR-2018, Luo-ICCV-2017, Ramachandra-WACV-2020b, Ravanbakhsh-WACV-2018, Ravanbakhsh-ICIP-2017, Sabokrou-IP-2017, Smeureanu-ICIAP-2017, Xu-CVIU-2017, Doshi-CVPRW-2020a, Doshi-CVPRW-2020b} employed deep learning in order to detect the anomalous frames in a video. For instance, Liu et al.~\cite{Liu-CVPR-2018} proposed to detect abnormal frames by predicting the next frame in the video, given the previous four frames. Their hypothesis is that an abnormal frame should be harder to predict than a normal one. Thus, the peak signal-to-noise ratio between the predicted frame and the original frame is expected to be lower for abnormal frames. More recently, Ramachandra et al.~\cite{Ramachandra-WACV-2020b} used video-patches to detect the anomalies in videos. They stored a set of exemplars for each region in the video frames, comparing each video patch from the test video to each exemplar from the corresponding region. The minimum distance is interpreted as the anomaly score. 
 
Similar to our work, which learns features in an unsupervised manner, there are a few works that employ unsupervised learning steps for abnormal event detection \cite{Dutta-AAAI-2015, Gong-ICCV-2019, Hasan-CVPR-2016, Ren-BMVC-2015, Xu-CVIU-2017}. There are also some works that are completely unsupervised \cite{Giorno-ECCV-2016, Ionescu-ICCV-2017, Liu-BMVC-2018, Pang-CVPR-2020}, requiring no training data to perform anomaly detection. Different from such works, our method employs an object detector trained with supervision. The object detector helps our method to become background-agnostic. Another way we introduce supervision into our framework is through pseudo-abnormal examples. Our auto-encoders learn to output poor reconstructions for pseudo-abnormal examples through adversarial training, while still being capable of reconstructing normal patterns. We also use the pseudo-abnormal examples to train binary classifiers on top of the auto-encoder features and reconstructions, in a supervised way.

More closely related to our work, some methods use auto-encoders in order to learn useful features \cite{Gong-ICCV-2019, Hasan-CVPR-2016, Sabokrou-IP-2017, Xu-CVIU-2017, Nguyen-ICCV-2019}. For example, Nguyen et al.~\cite{Nguyen-ICCV-2019} used an auto-encoder with two branches to detect anomalous frames. The input of the auto-encoder is the current frame of the video. The first branch predicts the frame intensity, while the second branch predicts the motion between the current frame and the next frame. Unlike the first branch, which is a standard auto-encoder, the second branch follows a U-Net architecture. Different from Nguyen et al.~\cite{Nguyen-ICCV-2019} and other methods based on auto-encoders, we train our auto-encoders on detected objects, which helps our method to better localize anomalies and to become background-agnostic. We also employ adversarial training, which enables us to obtain good reconstructions for normal objects only.

There are also some works~\cite{Lee-ICASSP-2018, Nguyen-ICCV-2019,Ravanbakhsh-ICIP-2017} that use Generative Adversarial Networks (GANs)~\cite{Goodfellow-NIPS-2014} in order to detect abnormal events in videos. For instance, Nguyen et al.~\cite{Nguyen-ICCV-2019} used a discriminator to distinguish between the generated optical flows and the real ones. Therefore, in order to reconstruct the motion, their network is also guided by the discriminator to produce realistic optical flows. Our method does not employ GANs, i.e.~we do not train a generator and a discriminator in an adversarial fashion. In our case, the adversarial training consists of propagating the reversed gradients from an adversarial decoder through the encoder, forcing the auto-encoder to output bad reconstructions for pseudo-abnormal examples. While we integrate a binary classifier after each auto-encoder, the respective classifier is only trained to discriminate between normal and pseudo-abnormal reconstructions (both kinds of examples being generated, not real), without interfering with the learning process of the auto-encoder. We refrain from referring to our binary classifiers as discriminators to avoid any confusion with GANs. Unlike these previous works~\cite{Lee-ICASSP-2018, Nguyen-ICCV-2019,Ravanbakhsh-ICIP-2017}, we are the first to perform adversarial training using a generic set of pseudo-abnormal examples.

We note that auto-encoders \cite{Kieu-IJCAI-2019,Kim-AISTATS-2020,Zhou-KDD-2017} and adversarial models \cite{Beggel-ECML-2019,Chen-CIKM-2020} have been applied to anomaly detection in other domains, e.g.~time series \cite{Kieu-IJCAI-2019}, images \cite{Beggel-ECML-2019,Kim-AISTATS-2020,Zhou-KDD-2017} or network traffic analysis \cite{Chen-CIKM-2020,Kim-AISTATS-2020}. Abnormal event detection is commonly viewed as an independent domain, requiring the design of specific methods that take into account both motion and appearance as well as other particularities of the video domain. Hence, we consider methods such as \cite{Beggel-ECML-2019,Chen-CIKM-2020,Kieu-IJCAI-2019,Kim-AISTATS-2020,Zhou-KDD-2017} to be distantly related.

\vspace{-0.2cm}
\section{Method}
\label{sec_Method}

\begin{figure*}[!t]
\begin{center}
\includegraphics[width=0.98\linewidth]{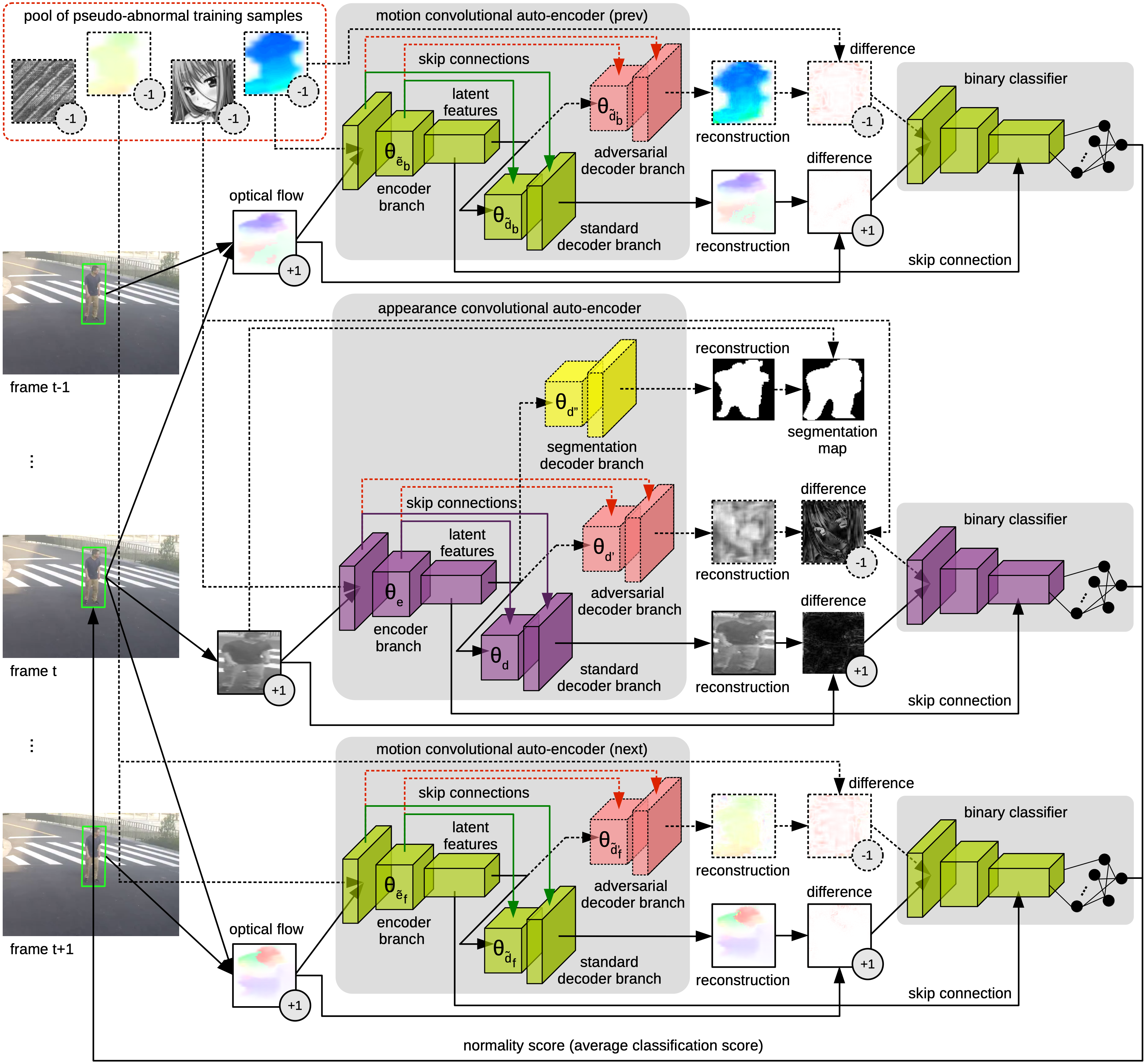}
\end{center}
\vspace{-0.5cm}
\caption{Our anomaly detection framework based on training convolutional auto-encoders with skip connections on top of object detections. In the learning phase, pseudo-abnormal examples are used to train the adversarial decoder branch using gradient ascent. The absolute differences between the inputs and the reconstructions are provided as input to a binary classifier corresponding to each convolutional auto-encoder. In the inference phase, we can label a test sample as abnormal if the average classification score is negative, i.e.~the sample is labeled as pseudo-abnormal. Components represented in dashed lines are removed during inference. Best viewed in color.}
\label{fig_pipeline}
\vspace{-0.2cm}
\end{figure*}

\subsection{Motivation}

We distinguish two realistic requirements that are especially desired when designing a framework for abnormal event detection in real-word scenarios. Based on these requirements, we introduce a set of design choices in our proposed framework.

The first requirement is to learn from training videos containing only normal events, deeming supervised learning methods needing both positive (normal) and negative (abnormal) samples unusable for abnormal event detection. Nonetheless, we believe that including any form of supervision is an important step towards obtaining better performance in practice. Motivated by this, we incorporate two approaches for including supervision into our framework. The first approach is to employ a single-shot object detector~\cite{Redmon-arXiv-2018}, which is trained with class and bounding-box supervision, in order to obtain object detections that are subsequently used throughout the rest of the processing pipeline. The second approach consists of gathering a large and generic pool of pseudo-abnormal examples, substituting the need for abnormal examples during training.

The second requirement is to apply the same model on multiple scenes with different backgrounds, eliminating the need to retrain the model for each and every scene. For example, a model that is trained on road traffic is expected to work well on multiple road traffic scenes. This has motivated us to design a background-agnostic framework. Since we employ an object detector and analyze abnormality at the object level, our framework is close to being background-agnostic. We take it a step further and equip our auto-encoders with a segmentation branch to focus on reconstructing only the corresponding segment in each bounding box, thus completely ignoring the background. We conduct cross-database experiments to demonstrate that our framework is indeed background-agnostic.

\vspace{-0.2cm}
\subsection{Overview}

Our abnormal event detection pipeline is illustrated in Figure~\ref{fig_pipeline}. An object detector is applied on the video, resulting in a set of object detections. For each object detection, optical flow maps are computed with respect to the previous and next frames. The object detections are given as input to an appearance auto-encoder, while the optical flow maps are given as input to two motion auto-encoders. The auto-encoders learn to reconstruct the detections and flow maps extracted from the input video. At the same time, the auto-encoders are prevented from learning to reconstruct examples from the pool of pseudo-abnormal training samples passing through the encoder and the adversarial decoder branch. The appearance auto-encoder has a third decoder branch that learns to reconstruct object segments. The segmentation branch helps the model to focus on the foreground object, ignoring the background inevitably caught inside each bounding box. The segmentation branch is also depicted in Figure~\ref{fig_pipeline}. The absolute differences between the inputs and reconstructions are subsequently used to learn three binary classifiers for discriminating between normal examples (labeled as positive) and pseudo-abnormal examples (labeled as negative). During inference, the average scores (class probabilities) provided by the three classifiers represent normality scores associated to the input object detections.

\vspace{-0.2cm}
\subsection{Object Detection and Preprocessing}

In this work, we employ a fast single-stage object detection framework, YOLOv3 \cite{Redmon-arXiv-2018}, that is pre-trained on MS COCO~\cite{Lin-ECCV-2014}. 
We do not fine-tune the object detector on abnormal event detection data sets due to the lack of bounding box annotations. Throughout our experiments, we did not observe any significant false negatives due to the employment of a pre-trained object detector. This is likely due to the fact that anomalous events are mostly associated with humans and their interaction with other humans or vehicles, e.g.~humans riding bikes, people fighting, people stealing or throwing backpacks and so on. Such categories (human, bike, backpack, etc.) are present in the MS COCO data set. Nevertheless, the detector can be retrained in case of any missing categories.  

We opted for YOLOv3 due to its combined advantage of superior detection performance (in terms of mean Average Precision) and high speed ($72$ frames per second on a single GPU).
We are interested in achieving an optimal trade-off between accuracy and speed, YOLOv3 being an excellent choice in this regard. The object detector is applied on each frame $t$, resulting in a set of bounding boxes at each frame. To obtain the input for the appearance auto-encoder, we crop the objects according to the detected bounding boxes, then, we convert the resulting image crops to grayscale. To obtain segmentation maps to be used as ground-truth labels during training time, we employ Mask R-CNN \cite{He-ICCV-2017}. Since Mask R-CNN is slower than YOLOv3, we only use it to obtain segments, necessary at training time, but not during inference. To obtain the motion representation corresponding to an object, we use optical flow. We compute the optical flow using the pre-trained version of SelFlow~\cite{Liu-CVPR-2019}, applying it on each tuple of three consecutive frames. More precisely, to compute the flow at frame $t$, the SelFlow network receives the frames $t - 1$, $t$ and $t + 1$. The resulting optical flow is composed of the forward and the backward optical flow maps. In order to extract the optical flow corresponding to each object in a frame $t$, we apply the bounding boxes detected by YOLOv3 to crop out the corresponding flow maps. As we obtain two flow maps from SelFlow, we employ two motion auto-encoders, one being used to encode the forward motion and the other to encode the backward motion. This is consistent with our preliminary framework~\cite{Ionescu-CVPR-2019}, which also employed two motion auto-encoders, although these were trained on top of image gradients instead of optical flow maps. Our motion auto-encoders are trained on tensors with two channels, one for the orientation and the other for the magnitude of the motion vectors.
 
\vspace{-0.2cm}
\subsection{Pseudo-Abnormal Examples}

We gathered a collection of pseudo-abnormal examples that are problem-agnostic, enabling us to use the same collection regardless of the data set. The purpose of using the pseudo-abnormal examples is twofold. First of all, pseudo-abnormal images force the auto-encoders to forget generic reconstruction patterns, projecting only normal objects to the learned manifold. Second of all, they represent an excellent way to fill in for the lack of abnormal data that would be required to train the binary classifiers using classic supervision.

The collection of pseudo-abnormal examples does not contain objects that can appear in real abnormal event detection scenarios, such as people, cars, bicycles, firearms and so on. Instead, it contains texture images \cite{Lazebnik-PAMI-2005}, flower images \cite{Nilsback-CVPR-2006}, anime images, butterfly images and some categories from Tiny ImageNet that are unrelated to abnormal event detection, such as \emph{hourglass} and \emph{acorn}. In total, the data set of pseudo-abnormal examples is composed of $66,\!918$ images. Since an object can be classified both as normal or abnormal, the distinction being based on the context, we completely refrain from adding any object class that is likely to appear in a video surveillance scene. To obtain pseudo-abnormal motion patterns, we compute optical flow maps on frame triplets selected at time $t-k$, $t$ and $t+k$ from the (normal) training video. We choose $k=\{3,4,5,6\}$ for the experiments, artificially magnifying the motion stored in the pseudo-abnormal optical flow maps.

\vspace{-0.2cm}
\subsection{Architecture}

Our neural architecture is composed of three independent processing streams (appearance, backward motion and forward motion), which are combined using a late fusion strategy. Each stream uses a convolutional auto-encoder (CAE) followed by a binary neural network classifier. The three auto-encoders share the same lightweight architecture, with only their input and trainable weights being different. One CAE takes as input cropped grayscale images of objects and learns to encode the appearance of objects in its latent space. The other two CAEs receive as input the orientation and the magnitude of the motion vectors stored in the optical flow maps, learning to represent motion in their latent spaces. The input size for the appearance CAE is $64\times64$, while the input size for the motion CAEs is $64\times64 \times 2$. 

Each encoder is composed of three convolutional (conv) layers, each followed by a max-pooling layer with a filter size of $2 \times 2$ applied at a stride of $2$. The conv layers are formed of $3 \times 3$ filters. Each conv layer is followed by Rectified Linear Units (ReLU)~\cite{Nair-ICML-2010} as the activation function. The first two conv layers consist of $32$ filters, while the third layer consists of $16$ filters. The latent representation is composed of $16$ activation maps of size $8\times8$.

Each decoder starts with an upsampling layer, increasing the spatial support of the activation maps by a factor of $2\times$. The upsampling operation is based on nearest neighbor interpolation. After upsampling, we apply a conv layer with $16$ filters of $3 \times 3$. The first upsampling and conv block is followed by another two upsampling and conv blocks. The last conv layer of an appearance decoder is formed of a single conv filter, while the last conv layer of a motion decoder is formed of two filters. In both cases, the number of filters in  the last conv layer is chosen such that the size of the output matches the size of the input.

We note that the appearance CAE incorporates one encoder $e$ and three decoder branches. The first decoder $d$ is used to reconstruct the normal objects, the second decoder $d'$ is used to decode the pseudo-abnormal objects and the third decoder $d''$ is used to generate a mask that segments the object and ignores the background of the input image. Different from the appearance CAE, the motion CAEs incorporate one encoder $\tilde{e}_*$ and only two decoder branches, one denoted by $\tilde{d}_*$, which reconstructs the normal objects, and the other denoted by $\tilde{d}'_*$, which decodes the pseudo-abnormal examples, where $*$ can be replaced with values from the set $\{b, f\}$, where $b$ represents the backward motion and $f$ represents the forward motion.
 
In order to distinguish between normal and pseudo-abnormal examples, we train binary classifiers using classic supervision. We designate a binary classifier for each of the three streams. The actual input of a binary classifier is the absolute difference between the input and the output of the corresponding CAE. In a set of preliminary experiments, we tried to use directly the reconstructions instead of the absolute differences, but the results were considerably worse. For the appearance CAE, the absolute difference is a matrix of $64\times64$ components, while for the motion CAEs, the absolute difference is a tensor of $64\times64\times2$ components. Since the inputs of the binary classifiers are of the same shape as the CAEs, we employ the same sequence of three conv and max-pooling layers as in the encoders. The resulting activation maps are subsequently passed through a neural network that follows the LeNet architecture \cite{lecun-bottou-ieee-1998}, except that the spatial support of the conv filters is always $3\times3$ and the pooling operation is max-pooling instead of average-pooling. In summary, a binary classifier is composed of five conv layers, a fully-connected layer and a Softmax classification layer.

We note that our architecture is also equipped with skip connections. Each conv layer in an encoder has skip connections to the corresponding conv layer in each standard or adversarial decoder that belongs to the same CAE as the encoder, following the U-Net architecture~\cite{Ronneberger-MICCAI-2015}. Instead of concatenating the features as in U-Net, we sum up the corresponding activation maps, as in ResNet~\cite{He-CVPR-2016}. Skip connections are also added between the last conv layer of an encoder and the third conv layer of the corresponding binary classifier. Since the first part of the binary classifier coincides with the encoder in terms of architecture, we can sum up the corresponding activation maps, both forming a tensor of $8\times 8\times 16$ components.

\vspace{-0.2cm}
\subsection{Training the Auto-Encoders}  

We train the auto-encoders using adversarial training, propagating the reversed gradients from the adversarial branch of each CAE through the corresponding encoder. Let $\theta_e$, $\theta_d$, $\theta_{d'}$ and $\theta_{d''}$ be the parameters of the appearance encoder $e$, the main appearance decoder $d$, the adversarial appearance decoder $d'$ and the segmentation decoder $d''$, respectively. Analogously, let $\theta_{\tilde{e}_*}$, $\theta_{\tilde{d}_*}$ and $\theta_{\tilde{d}'_*}$ be the parameters of the motion encoder $\tilde{e}_*$, the main motion decoder $\tilde{d}_*$ and the adversarial motion decoder $\tilde{d}'_*$, where $* \in \{b,f\}$. 
For the motion auto-encoders, the loss function for reconstructing an input optical flow $\tilde{x}_*$ of $h\times w\times c$ components is the pixel-wise mean squared error:
\begin{equation}\label{eq_mot_rec}
\mathcal{L}_{*\mbox{\scriptsize{mot-rec}}} (\tilde{x}_*,\hat{x}_*) = \frac{1}{h \cdot w \cdot c}\sum_{i=1}^{h}\sum_{j=1}^{w}\sum_{k=1}^{c} \left( \tilde{x}_{*ijk} - \hat{x}_{*ijk}  \right)^2,
\end{equation}
where $\hat{x}_*=\tilde{d}_*(\tilde{e}_*(\tilde{x}_*,\theta_{\tilde{e}_*}),\theta_{\tilde{d}_*})$ is the main output of a motion auto-encoder, $\forall\; * \in \{b,f\}$. We note that $h=w=64$, while $c=2$. Similarly, we define the loss for the adversarial branch of the motion CAEs as follows:
\begin{equation}\label{eq_mot_adv}
\mathcal{L}_{*\mbox{\scriptsize{mot-adv}}} (\tilde{x}_*,\bar{x}_*) = \frac{1}{h \cdot w \cdot c}\sum_{i=1}^{h}\sum_{j=1}^{w}\sum_{k=1}^{c} \left( \tilde{x}_{*ijk} - \bar{x}_{*ijk}  \right)^2,
\end{equation}
where $\bar{x}_*=\tilde{d}'_*(\tilde{e}_*(\tilde{x}_*,\theta_{\tilde{e}_*}),\theta_{\tilde{d}'_*})$ is the output of an adversarial motion decoder, $\forall\; * \in \{b,f\}$. In the above equations, it is important to highlight that the notations $\tilde{e}_*$, $\tilde{d}_*$ and $\tilde{d}'_*$ are interchangeably used to denote components of both motion auto-encoders. In other words, $*$ joins the notations $\tilde{e}_b$, $\tilde{d}_b$ and $\tilde{d}'_b$ designated for the components of the backward motion CAE and the notations $\tilde{e}_f$, $\tilde{d}_f$ and $\tilde{d}'_f$ designated for the components of the forward motion CAE. We note that the independent notations are used in Figure~\ref{fig_pipeline}.

For the appearance CAE, we also have the segmentation decoder $d''$, which is trained jointly with the principal decoder $d$. Hence, the loss function for reconstructing an input $x$ and a segmentation map $s$ is the sum of the pixel-wise mean squared error computed with respect to main decoder and the logistic error computed with respect to the segmentation decoder:
\begin{equation}\label{eq_app_rec}
\begin{split}
&\mathcal{L}_{\mbox{\scriptsize{app-rec}}} (x,\hat{x},s,\hat{s}) = \frac{1}{h \cdot w}\sum_{i=1}^{h}\sum_{j=1}^{w} \left( x_{ij} - \hat{x}_{ij}  \right)^2 +\\
&+ \frac{1}{h\!\cdot\!w}\sum_{i=1}^{h}\sum_{j=1}^{w} \!- s_{ij}\!\cdot\!\log(\hat{s}_{ij}) - (1 - s_{ij})\!\cdot\!\log(1- \hat{s}_{ij}) ,
\end{split}
\end{equation}
where $x$ is a grayscale image, $\hat{x}=d(e(x,\theta_e),\theta_d)$ is the main output of the auto-encoder and $\hat{s}=d''(e(x,\theta_e),\theta_{d''})$ is the output of the segmentation decoder. We notice that the segmentation mask is estimated from the input $x$, the ground-truth segmentation mask $s$ being used only for the comparison with the output $\hat{s}$. According to the described architecture, the size of both inputs is $h\!=\!w\!=\!64$. As the segmentation map is a binary map separating the foreground object from the background, we opted for the logistic loss on the segmentation branch. The loss for the adversarial branch of the appearance CAE is defined analogously to Equation~\eqref{eq_mot_adv}:
\begin{equation}\label{eq_app_adv}
\mathcal{L}_{\mbox{\scriptsize{app-adv}}} (x,\bar{x}) = \frac{1}{h \cdot w}\sum_{i=1}^{h}\sum_{j=1}^{w} \left( x_{ij} - \bar{x}_{ij} \right)^2,
\end{equation}
where $\bar{x}=d'(e(x,\theta_e),\theta_d')$ is the output of the adversarial decoder.  
Since our goal is to obtain poor reconstructions for the pseudo-abnormal examples, we train the decoders $d'$ and $\tilde{d}'_*$ as adversaries.

For the motion auto-encoders, the parameters $\theta_{\tilde{d}'_*}$ are updated to optimize the loss defined in Equation~\eqref{eq_mot_adv}, while the parameters $\theta_{\tilde{e}_*}$ of the encoder are updated to fool the decoder $\tilde{d}'_*$. This leads to the following update rules for the parameters:
\begin{equation}\label{eq_mot_update_d}
\theta_{\tilde{d}_*} \leftarrow \theta_{\tilde{d}_*} - \eta \cdot \frac{\partial \mathcal{L}_{*\mbox{\scriptsize{mot-rec}}}}{\partial \theta_{\tilde{d}_*}} ,
\end{equation}
\vspace{-0.2cm}
\begin{equation}\label{eq_mot_update_d_adv}
\theta_{\tilde{d}'_*} \leftarrow \theta_{\tilde{d}'_*} - \eta \cdot \frac{\partial \mathcal{L}_{*\mbox{\scriptsize{mot-adv}}}}{\partial \theta_{\tilde{d}'_*}} ,
\end{equation}
\vspace{-0.18cm}
\begin{equation}\label{eq_mot_update_e}
\theta_{\tilde{e}_*} \leftarrow \theta_{\tilde{e}_*} - \eta \frac{\partial \mathcal{L}_{*\mbox{\scriptsize{mot-rec}}}}{\partial \theta_{\tilde{e}_*}} + \eta \cdot \lambda \cdot \frac{\partial \mathcal{L}_{*\mbox{\scriptsize{mot-adv}}}}{\partial \theta_{\tilde{e}_*}} ,
\end{equation}
where $\eta$ is the learning rate and $\lambda$ is a weight for the reversed gradient. We note that the encoder is trained using gradient descent with respect to the main decoder and gradient ascent with respect to the adversarial decoder. To ensure convergence, $\lambda$ must be less than $1$ (otherwise, the gradient ascent step will be greater than the gradient descent step). As suggested in~\cite{McHardy-NAACL-2019}, we set $\lambda=0.2$ for our experiments. For the appearance auto-encoder, the parameter update rules are equivalent, while only adding the standard gradient descent for the segmentation decoder:
\begin{equation}\label{eq_app_update_d}
\theta_d \leftarrow \theta_d - \eta \cdot \frac{\partial \mathcal{L}_{\mbox{\scriptsize{app-rec}}}}{\partial \theta_d} ,
\end{equation}
\vspace{-0.16cm}
\begin{equation}\label{eq_app_update_d_adv}
\theta_{d'} \leftarrow \theta_{d'} - \eta \cdot \frac{\partial \mathcal{L}_{\mbox{\scriptsize{app-adv}}}}{\partial \theta_{d'}} ,
\end{equation}
\vspace{-0.22cm}
\begin{equation}\label{eq_app_update_d}
\theta_{d''} \leftarrow \theta_{d''} - \eta \cdot \frac{\partial \mathcal{L}_{\mbox{\scriptsize{app-rec}}}}{\partial \theta_{d''}} ,
\end{equation}
\vspace{-0.22cm}
\begin{equation}\label{eq_app_update_e}
\theta_{e} \leftarrow \theta_{e} - \eta \frac{\partial \mathcal{L}_{\mbox{\scriptsize{app-rec}}}}{\partial \theta_{e}} + \eta \cdot \lambda \cdot \frac{\partial \mathcal{L}_{\mbox{\scriptsize{app-adv}}}}{\partial \theta_{e}}.
\end{equation}
Once again, we set $\lambda=0.2$ for the experiments.
We train the auto-encoders using the Adam optimizer \cite{Kingma-ICLR-2014} with the learning rate $\eta=10^{-4}$, keeping the default values for the other parameters of Adam.

\vspace{-0.2cm}
\subsection{Training the Binary Classifiers}

After training the auto-encoders until convergence, we remove the additional decoder branches $d'$, $d''$ and $\tilde{d}'_*$, while freezing the parameters $\theta_e$, $\theta_d$, $\theta_{\tilde{e}_*}$ and $\theta_{\tilde{d}_*}$ of the remaining components. We then pass all the training examples, including the adversarial ones, through the main appearance and motion decoders, obtaining the final reconstructions for the training data. The next step is to compute the absolute difference between each input example and its corresponding reconstruction. For a tensor $x$ representing the absolute difference and the binary label $y$ associated to $x$, we employ the binary cross-entropy to train the classifiers:
\begin{equation}\label{eq_BCE_loss}
\mathcal{L}_{\mbox{\scriptsize{cross-entropy}}}(y, \hat{y}) = - y \cdot \log(\hat{y}) + (1 - y) \cdot \log(1 - \hat{y}),
\end{equation}
where $\hat{y}$ represents the prediction (class probability) for sample $x$. In order to use the cross-entropy loss, the normal examples, taken from the training video, are labeled with $y=1$ and the pseudo-abnormal examples are labeled with $y=0$. Our classifiers are optimized using Adam \cite{Kingma-ICLR-2014} with a learning rate of $10^{-3}$.

\vspace{-0.2cm}
\subsection{Inference}

During inference, we run the YOLOv3 detector to determine the bounding boxes of objects in the current frame. We then compute the optical flow maps for the entire frame. For each detected object, we apply the CAEs (without the additional decoder branches $d'$, $d''$ and $\tilde{d}'_*$) to obtain the appearance and motion reconstructions. Then, we compute the absolute differences and pass them to the binary classifiers. Because we have three binary classifiers, we obtain three class probabilities for each object. A class probability is interpreted as a normality score normalized between $0$ and $1$. The final anomaly score for an object $x$ is obtained by subtracting the average of the three normality scores from $1$, resulting in an anomaly score between $0$ and $1$:
\begin{equation}\label{eq_inference}
s(x) = 1 - \text{mean}\left(\hat{y}^{(i)}\right), \forall i \in \{1,2,3\},
\end{equation}
where $\hat{y}^{(i)}$ is a normality score provided by one of the three classifiers. We expect abnormal examples from the test video to be classified as pseudo-abnormal examples, having an abnormality score closer to $1$.

By reassembling the anomaly scores of the detected objects into an anomaly map for each frame, we obtain pixel-level anomaly detections. Hence, our framework can also perform anomaly localization. When the bounding boxes of two objects overlap, we keep the maximum anomaly score for the overlapping area. In order to make the pixel-level maps smoother, we apply a $3D$ mean filter. The frame-level anomaly score is obtained by taking the maximum inside the prediction map for the corresponding frame. We further apply a Gaussian filter to temporally smooth the frame-level anomaly scores.

\vspace{-0.2cm}
\section{Experiments}
\label{sec_Experiments}
 
\subsection{Data Sets}

We present results on four benchmark data sets, namely Avenue \cite{Lu-ICCV-2013}, ShanghaiTech \cite{Luo-ICCV-2017}, Subway \cite{Adam-PAMI-2008} and UCSD Ped2 \cite{Mahadevan-CVPR-2010}. Although UMN \cite{Mehran-CVPR-2009} is among the most widely-used abnormal event detection benchmarks, we consider this data set as being small and saturated. Hence, we do not conduct experiments on UMN.

\noindent
\textbf{Avenue.} The Avenue data set \cite{Lu-ICCV-2013} consists of 16 training videos and 21 test videos. The training videos have a total of $15,\!328$ frames, while the test videos have $15,\!324$ frames. The resolution of each video frame is $360 \times 640$ pixels. The original data set is annotated both at the pixel level and at the frame level. Region-level and track-level annotations are provided by Ramachandra et al.~\cite{Ramachandra-WACV-2020a}.

\noindent
\textbf{ShanghaiTech.} The ShanghaiTech Campus data set \cite{Luo-ICCV-2017} is one of the largest data sets for abnormal event detection. It contains $330$ training videos and $107$ test videos. The resolution of each video frame is $480 \times 856$ pixels and the data set has a total of $316,\!154$ frames. ShanghaiTech contains both frame-level and pixel-level annotations. We provide region-level and track-level annotations for ShanghaiTech.

\noindent
\textbf{Subway.} The Subway surveillance data set \cite{Adam-PAMI-2008} is formed of two videos, training and testing being conducted independently on the two videos. The length of one video (Entrance gate) is $96$ minutes and the length of the other (Exit gate) is $43$ minutes. The Entrance gate video has $144,\!251$ frames, while the Exit gate has $64,\!903$ frames. The resolution of each video frame is $384 \times 512$ pixels. For the Entrance video, we follow~\cite{Ionescu-WACV-2019, Cheng-CVPR-2015} and split the video in  $53\%$ frames for training and $47\%$ frames for testing. For the Exit video, similar to \cite{Lu-ICCV-2013, Ionescu-WACV-2019},  we use the first $15$ minutes for training and the rest of the video for testing. This data set contains only frame-level labels. We provide pixel-level, region-level and track-level annotations for both videos in the Subway data set. 
 
\noindent
\textbf{UCSD Ped2.} The UCSD Ped2 data set \cite{Mahadevan-CVPR-2010} is formed of 16 training videos and 12 test videos. The training videos have a total of $2,\!550$ frames, while the test videos have $2,\!010$ frames. The resolution of each video frame is $240 \times 360$ pixels. As for Avenue, region-level and track-level annotations are available due to Ramachandra et al.~\cite{Ramachandra-WACV-2020a}.
 
\begin{figure*}[!t]
\begin{center}
\includegraphics[width=0.98\linewidth]{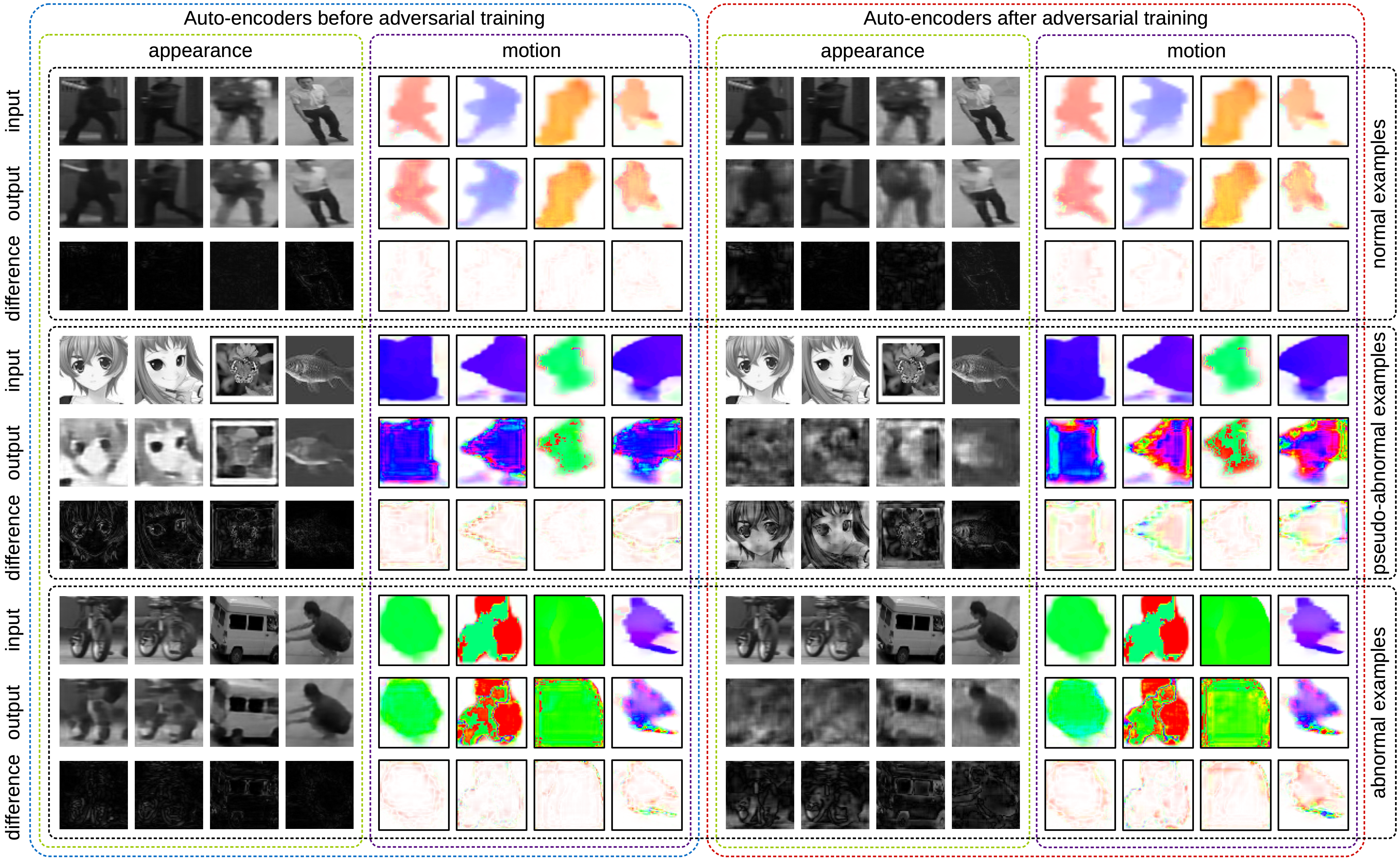}
\end{center}
\vspace{-0.5cm}
\caption{Normal (top), pseudo-abnormal (middle) and abnormal (bottom) examples and optical flow maps with reconstructions provided by the appearance and the motion convolutional auto-encoders, which are trained either without adversarial training (left) or with adversarial training (right). The auto-encoders provide worse reconstructions for pseudo-abnormal and abnormal examples after adversarial training, which is the desired effect. The normal and abnormal samples are selected from the Avenue~\cite{Lu-ICCV-2013} and the ShanghaiTech~\cite{Luo-ICCV-2017} test sets, while the pseudo-abnormal examples are selected from our pool of generic pseudo-abnormal examples. Best viewed in color.}
\label{fig_prelim}
\vspace{-0.2cm}
\end{figure*}

\vspace{-0.2cm}
\subsection{Evaluation}

As our first evaluation metric, we consider the area under the curve (AUC) computed with respect to the ground-truth frame-level annotations. At a given threshold, a frame is labeled as abnormal if at least one pixel inside the frame is abnormal. With some exceptions, we note that many previous works do not mention if the frame-level AUC is computed by $(i)$ concatenating all frames then computing the score (this is the micro-averaged AUC), or by $(ii)$ computing the frame-level AUC for each video, then averaging the resulting scores (this is the macro-averaged AUC). We therefore report both frame-level AUC measures. Additionally, we report the Region-Based Detection Criterion (RBDC) and the Track-Based Detection Criterion (TBDC), two new metrics introduced by Ramachandra et al.~\cite{Ramachandra-WACV-2020a}. In their work, Ramachandra et al.~\cite{Ramachandra-WACV-2020a} demonstrated that the frame-level AUC and the pixel-level AUC are not representative metrics to evaluate abnormal event detection frameworks, thus proposing RBDC and TBDC as alternative metrics. RBDC takes into consideration every region that is detected as abnormal. If the intersection over union of a ground-truth region and a predicted region is at least $\beta$, the predicted region is considered a true positive, otherwise it is a false positive.
TBDC takes into consideration the detection of tracks. A track is an abnormal event that occurs across several consecutive frames. Each track is formed of a set of regions. A track is considered detected if at least a fraction $\alpha$ of the ground-truth regions belonging to the track is detected. More details about these new evaluation metrics are presented in~\cite{Ramachandra-WACV-2020a,Ramachandra-PAMI-2020}.
Following Ramachandra et al.~\cite{Ramachandra-WACV-2020a}, we set $\alpha=0.1$, $\beta=0.1$ and compute the area under the ROC curve considering false positive rates that are less than or equal to $1$. 

As observed in~\cite{Ramachandra-WACV-2020a,Ramachandra-PAMI-2020}, a simple post-processing step can make the pixel-level AUC equal to the frame-level AUC. The post-processing step consists in labeling all pixels in a frame as abnormal if at least one pixel in the frame is abnormal, increasing the true positive rate without modifying the false positive rate. Hence, we refrain from independently reporting the pixel-level AUC.

\vspace{-0.2cm}
\subsection{Parameter and Implementation Details}

In the object detection stage, we employ the YOLOv3~\cite{Redmon-arXiv-2018} detector, which is pre-trained on the MS COCO data set \cite{Lin-ECCV-2014}. During training and inference, we keep the detections with a confidence level higher than $0.8$ for Avenue and ShanghaiTech. Because Subway and UCSD Ped2 have lower frame resolutions, we set the confidence level at $0.5$ for these two data sets. We employ the pre-trained Mask R-CNN~\cite{He-ICCV-2017} to obtain frame-level segmentation maps, and the pre-trained SelFlow~\cite{Liu-CVPR-2019} to obtain frame-level optical flow maps. The CAEs and the binary classifiers are trained from scratch in TensorFlow \cite{Abadi-OSDI-2016}. We train the CAEs for $20$ epochs with the learning rate set to $10^{-4}$, as in \cite{Ionescu-CVPR-2019}. The binary classifiers are trained for $30$ epochs with the learning rate set to $10^{-3}$. In each experiment, we set the mini-batch size to $64$ samples. 

\vspace{-0.2cm}
\subsection{Preliminary Results}

In Figure~\ref{fig_prelim}, we present a set of qualitative results to analyze the behavior of the appearance and motion auto-encoders with and without adversarial training. For each input sample, we provide the corresponding output and the absolute difference between input and output, respectively. 
We notice that, without adversarial training, the auto-encoders provide excessively good reconstructions for the abnormal examples, indicating that the auto-encoders generalize well to abnormal data samples coming from a very close distribution to the training data distribution. Unfortunately, in our application domain, we do not want the auto-encoders to generalize to abnormal examples. Instead, we would prefer to obtain visibly worse reconstructions for the abnormal examples, thus enabling the detection of such examples. This is the main motivation behind our decision to introduce adversarial training. We observe that, in general, the auto-encoders based on adversarial training provide worse reconstructions. However, we are not interested in the quality of the reconstructions, but in the difference of quality between reconstructions for normal samples and reconstructions for abnormal samples. Indeed, we observe that the auto-encoders based on adversarial training exhibit visibly worse reconstructions for abnormal examples than for normal examples, this being the desired effect. We notice the same effect on the pseudo-abnormal examples, only at a greater level. This happens because the pseudo-abnormal examples are used in the adversarial training procedure, while the abnormal examples are selected from the test set. In summary, we conclude that introducing adversarial training is helpful in achieving the desired goal, that of obtaining visibly better reconstructions for normal examples than for abnormal examples.

\vspace{-0.2cm}
\subsection{Quantitative Results on Avenue}

\begin{table}[!t]
\renewcommand{\arraystretch}{1.1} 
\caption{Micro-averaged AUC, macro-averaged AUC, RBDC and TBDC scores (in $\%$) of our approach compared to the state-of-the-art methods~\cite{Lu-ICCV-2013, Hasan-CVPR-2016, Giorno-ECCV-2016, Smeureanu-ICIAP-2017, Ionescu-ICCV-2017, Luo-ICCV-2017, Liu-CVPR-2018, Liu-BMVC-2018, Ionescu-WACV-2019, Doshi-CVPRW-2020a, Doshi-CVPRW-2020b, Nguyen-ICCV-2019, Tang-PRL-2020, Dong-Access-2020, Park-CVPR-2020, Wu-TNNLS-2019, Lee-ICASSP-2018, Lee-TIP-2019, Ramachandra-WACV-2020a, Ramachandra-WACV-2020b, Ionescu-CVPR-2019, Sun-ACMMM-2020, Wang-ACMMM-2020, Yu-ACMMM-2020} on the Avenue data set. When it is unclear if the reported frame-level AUC is micro-averaged or macro-averaged, we placed the score in the middle. All results are copied from the corresponding papers, except for those marked with asterisk (which are computed by ourselves using the official implementations). The best results are highlighted in bold.}
\label{table_avenue_sota}
\vspace{-0.3cm}
\centering
\begin{tabular}{|l|c|c|c|c|}
\hline
Method                & \multicolumn{2}{|c|}{AUC}          &  RBDC      & TBDC \\
 
\cline{2-3}                                                &      Micro   & Macro  & &     \\
\hline
Lu et al.~\cite{Lu-ICCV-2013}                   &     \multicolumn{2}{|c|}{$80.9$}     &       -         & -       \\
\hline
Hasan et al.~\cite{Hasan-CVPR-2016}             &      \multicolumn{2}{|c|}{$70.2$}    &       -         & -       \\
\hline 
Del Giorno et al.~\cite{Giorno-ECCV-2016}       &      $78.3$        & -               &       -         & -       \\
\hline 
Smeureanu et al.~\cite{Smeureanu-ICIAP-2017}    &      $84.6$        & -               &       -         & -       \\
\hline
Ionescu et al.~\cite{Ionescu-ICCV-2017}         &      $80.6$        & -               &       -         & -       \\
\hline
Luo et al.~\cite{Luo-ICCV-2017}                 &      $81.7$        & -               &       -         & -       \\
\hline
Liu et al.~\cite{Liu-CVPR-2018}                 &      $85.1$        &  $81.7$*     &       $19.59$*       & $56.01$*   \\
\hline
Liu et al.~\cite{Liu-BMVC-2018}                 &       $84.4$       & -               &       -         & -       \\
\hline 
Lee et al.~\cite{Lee-ICASSP-2018}               &      \multicolumn{2}{|c|}{$87.2$}    &         -       &  -       \\
\hline
Lee et al.~\cite{Lee-TIP-2019}                  &      \multicolumn{2}{|c|}{$90.0$}   &         -       &  -       \\
\hline
Ionescu et al.~\cite{Ionescu-WACV-2019}         &      $88.9$        &   -             &       -         & -       \\
\hline
Wu et al.~\cite{Wu-TNNLS-2019}                  &      \multicolumn{2}{|c|}{$86.6$}    &         -       &  -       \\
\hline
Nguyen et al.~\cite{Nguyen-ICCV-2019}           &      $86.9$        &      -          &         -       &  -       \\
\hline 
Ionescu et al.~\cite{Ionescu-CVPR-2019}         &      $87.4$*        & $\mathbf{90.4}$   &       $15.77$*   &  $27.01$* \\
\hline
Tang et al.~\cite{Tang-PRL-2020}                &       \multicolumn{2}{|c|}{$85.1$}    &         -       &  -       \\
\hline
Dong et al.~\cite{Dong-Access-2020}             &     \multicolumn{2}{|c|}{$84.9$}      &         -       &  -       \\
\hline
Park et al.~\cite{Park-CVPR-2020}               &     -              &       $88.5$    &         -       &  -       \\ 
\hline
Doshi et al.~\cite{Doshi-CVPRW-2020a,Doshi-CVPRW-2020b}           &       \multicolumn{2}{|c|}{$86.4$}    &       -         & -       \\
\hline
Ramachandra et al.~\cite{Ramachandra-WACV-2020a}&      \multicolumn{2}{|c|}{$72.0$}     &    $35.80$      &  $\mathbf{80.90}$ \\
\hline
Ramachandra et al.~\cite{Ramachandra-WACV-2020b}&      \multicolumn{2}{|c|}{$87.2$}     &    $41.20$      &    $78.60$ \\
\hline
Sun et al.~\cite{Sun-ACMMM-2020}                &     \multicolumn{2}{|c|}{$89.6$}      &         -       &  -       \\
\hline
Wang et al.~\cite{Wang-ACMMM-2020}               &     \multicolumn{2}{|c|}{$87.0$}      &         -       &  -       \\
\hline
Yu et al.~\cite{Yu-ACMMM-2020}                   &    $89.6$   &  -     &         -       &  -       \\
\hline
{Ours}              &      $\mathbf{92.3}$        &        $\mathbf{90.4}$   &       $\mathbf{65.05}$   &  $66.85$      \\ 
\hline
\end{tabular}
\vspace{-0.1cm}
\end{table}


We first compare our approach with several state-of-the-art methods \cite{Lu-ICCV-2013, Hasan-CVPR-2016, Giorno-ECCV-2016, Smeureanu-ICIAP-2017, Ionescu-ICCV-2017, Luo-ICCV-2017, Liu-CVPR-2018, Liu-BMVC-2018, Ionescu-WACV-2019, Doshi-CVPRW-2020a, Doshi-CVPRW-2020b, Nguyen-ICCV-2019, Tang-PRL-2020, Dong-Access-2020, Park-CVPR-2020, Wu-TNNLS-2019, Lee-ICASSP-2018, Lee-TIP-2019, Ramachandra-WACV-2020a, Ramachandra-WACV-2020b, Ionescu-CVPR-2019, Sun-ACMMM-2020, Wang-ACMMM-2020, Yu-ACMMM-2020} reporting results on the Avenue data set. The corresponding micro-averaged frame-level AUC, macro-averaged frame-level AUC, RBDC and TBDC scores are presented in Table~\ref{table_avenue_sota}. The existing methods attain frame-level AUC scores between $70.2\%$ and $89.6\%$. Notably, in terms of the frame-level AUC values, our method surpasses all exiting methods. With a micro-averaged AUC of $92.3\%$, our method is the only method surpassing the $90\%$ threshold on the Avenue data set by a certain margin. There are at least two works \cite{Park-CVPR-2020, Ionescu-CVPR-2019} that report the macro-averaged AUC. Our current approach obtains the same macro-averaged AUC as our preliminary method \cite{Ionescu-CVPR-2019}, but our micro-averaged AUC is almost $5\%$ higher than the result reported in~\cite{Ionescu-CVPR-2019}. We conjecture that our approach attains superior results compared to previous methods because $(i)$ it focuses specifically on detected objects, eliminating false positive events that are not caused by objects, and $(ii)$ it uses adversarial training to increase reconstruction errors for abnormal objects, reducing the number of false negatives.

In terms of RBDC and TBDC, only Ramachandra et al.~\cite{Ramachandra-WACV-2020b, Ramachandra-WACV-2020a} report the results on the Avenue data set. Additionally, we compute the RBDC and TBDC scores for our previous framework presented in~\cite{Ionescu-CVPR-2019} and for Liu et al.~\cite{Liu-CVPR-2018}. We should emphasize that the original implementation of Liu et al.~\cite{Liu-CVPR-2018} produces extremely low RBDC and TBDC scores (close to zero). We had to modify their framework by adding a post-processing step that removes abnormal regions smaller than a certain area, leading to the better RBDC and TBDC scores shown in Table~\ref{table_avenue_sota}. Compared to Ramachandra et al.~\cite{Ramachandra-WACV-2020b}, we obtain a considerable improvement of $23.85\%$ in terms of RBDC. We believe that this improvement is due to the fact that we detect objects in the scene, resulting in higher overlaps between our predicted regions and the ground-truth regions. In terms of TBDC, Ramachandra et al.~\cite{Ramachandra-WACV-2020a} attains the state-of-the-art result of $80.90\%$. Since RBDC takes into consideration every region from the ground-truth and since we surpass Ramachandra et al.~\cite{Ramachandra-WACV-2020a} by a large margin, we conjecture that our method is able to detect many more ground-truth regions with a lower rate of false positives. However, given the fact that our TBDC is lower than that of Ramachandra et al.~\cite{Ramachandra-WACV-2020a}, we conjecture that our method does not detect all the tracks in the ground-truth, given the maximum false positive rate of $1$. Finally, comparing our method to the preliminary version proposed in~\cite{Ionescu-CVPR-2019}, we report improvements of almost $50\%$ in terms of RBDC and almost $40\%$ in terms of TBDC, respectively. We therefore consider that the proposed framework is significantly better compared to its earlier version~\cite{Ionescu-CVPR-2019}. 

\begin{figure}[!t]
\begin{center}
\includegraphics[width=1.0\linewidth]{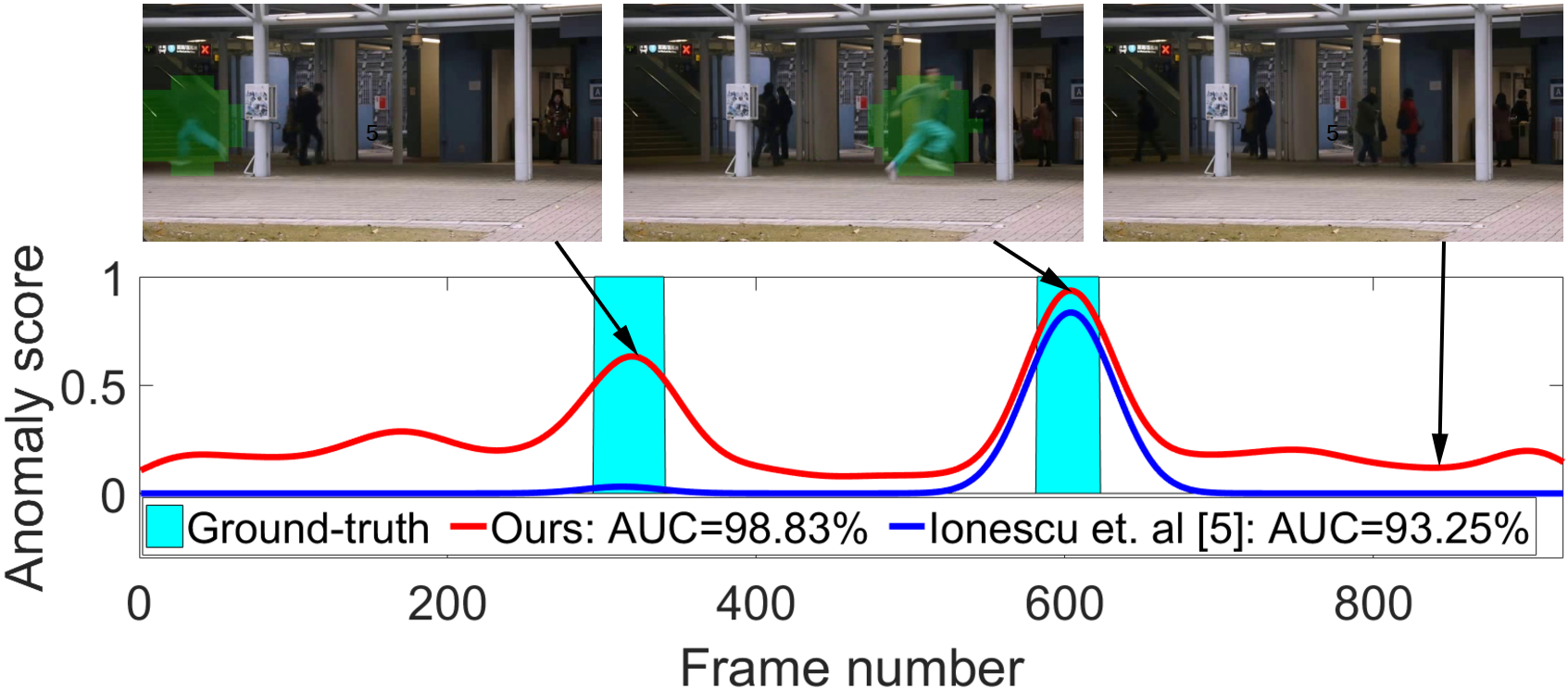}
\end{center}
\vspace{-0.5cm}
\caption{Frame-level anomaly scores (on the vertical axis) provided by our current approach versus the earlier version proposed in~\cite{Ionescu-CVPR-2019}, for test video 03 from Avenue~\cite{Lu-ICCV-2013}. Ground-truth abnormal events are represented in cyan, our scores are depicted in red and the scores of the earlier method are depicted in blue. Best viewed in color.}
\label{fig_avenue}
\vspace{-0.2cm}
\end{figure}

In Figure~\ref{fig_avenue}, we present the frame-level anomaly scores (corresponding to a frame-level AUC of $98.83\%$) produced by our method versus the anomaly scores (corresponding to a frame-level AUC of $93.25\%$) produced by our previous approach~\cite{Ionescu-CVPR-2019} on test video 03 from Avenue. According to the ground-truth labels, which are also illustrated in Figure~\ref{fig_avenue}, there are two abnormal events in the respective test video. Our approach is able to identify both events, without including any false positive detections, while our earlier approach~\cite{Ionescu-CVPR-2019} only identifies the second event. We would like to emphasize that adversarial training plays a key role in detecting the first abnormal event, which is missed by our preliminary framework \cite{Ionescu-CVPR-2019}.

\vspace{-0.2cm}
\subsection{Quantitative Results on ShanghaiTech}

\begin{table}[!t]
\renewcommand{\arraystretch}{1.1} 
\caption{Micro-averaged AUC, macro-averaged AUC, RBDC and TBDC scores  (in $\%$) of our approach compared to the state-of-the-art methods~\cite{Hasan-CVPR-2016, Luo-ICCV-2017, Liu-CVPR-2018, Sultani-CVPR-2018, Doshi-CVPRW-2020a, Doshi-CVPRW-2020b, Tang-PRL-2020, Dong-Access-2020, Park-CVPR-2020, Lee-TIP-2019, Ionescu-CVPR-2019, Sun-ACMMM-2020, Wang-ACMMM-2020, Yu-ACMMM-2020} on the ShanghaiTech data set. When it is unclear if the reported frame-level AUC is micro-averaged or macro-averaged, we placed the score in the middle. All results are copied from the corresponding papers, except for those marked with asterisk (which are computed by ourselves using the official implementations). The best results are highlighted in bold.}
\label{table_shanghai_sota}
\vspace{-0.3cm}
\centering
\begin{tabular}{|l|c|c|c|c|}
\hline
Method                & \multicolumn{2}{|c|}{AUC}          &  RBDC      & TBDC \\
 
\cline{2-3}                                                &      Micro   & Macro  & &     \\
\hline
Hasan et al.~\cite{Hasan-CVPR-2016}               &       \multicolumn{2}{|c|}{$60.9$}      &       -       & -       \\ 
\hline
Luo et al.~\cite{Luo-ICCV-2017}                   &      $68.0$        & -                  &       -       & -       \\ 
\hline
Liu et al.~\cite{Liu-CVPR-2018}                   &      $72.8$        & $80.6$*           &       $17.03$*       &  $54.23$*       \\ 
\hline 
Sultani et al.~\cite{Sultani-CVPR-2018}           &      -             &  $76.5$*            &       -       & -       \\ 
\hline
Lee et al.~\cite{Lee-TIP-2019}                    &     \multicolumn{2}{|c|}{$76.2$}        &         -     &  -       \\
\hline
Ionescu et al.~\cite{Ionescu-CVPR-2019}           &     $78.7$*         &        $84.9$      &       $20.65$* &  $44.54$* \\ 
\hline
Doshi et al.~\cite{Doshi-CVPRW-2020a,Doshi-CVPRW-2020b}             &      \multicolumn{2}{|c|}{$71.6$}       &       -       & -       \\
\hline 
Tang et al.~\cite{Tang-PRL-2020}                  &      \multicolumn{2}{|c|}{$73.0$}       &         -     &  -       \\
\hline
Dong et al.~\cite{Dong-Access-2020}               &      \multicolumn{2}{|c|}{$73.7$}       &         -     &  -       \\
\hline
Park et al.~\cite{Park-CVPR-2020}                 &     -              &       $72.8$       &         -     &  -       \\
\hline
Sun et al.~\cite{Sun-ACMMM-2020}                &     \multicolumn{2}{|c|}{$74.7$}      &         -       &  -       \\
\hline
Wang et al.~\cite{Wang-ACMMM-2020}               &     \multicolumn{2}{|c|}{$79.3$}      &         -       &  -       \\
\hline
Yu et al.~\cite{Yu-ACMMM-2020}                   &   $74.8$  &  -       &         -       &  -       \\
\hline 
{Ours}                                     &     $\mathbf{82.7}$         &        $\mathbf{89.3}$      &       $\mathbf{41.34}$ &  $\mathbf{78.79}$   \\
 
\hline
\end{tabular}
\vspace{-0.3cm}
\end{table}

\begin{table*}[!t]
\renewcommand{\arraystretch}{1.1} 
\caption{Frame-level AUC, RBDC and TBDC scores (in $\%$) of our approach compared to the state-of-the-art methods~\cite{Cheng-CVPR-2015, Hasan-CVPR-2016, Ionescu-WACV-2019, Giorno-ECCV-2016, Wu-TNNLS-2019, Saligrama-CVPR-2012, Pang-CVPR-2020, Cong-CVPR-2011, Ionescu-CVPR-2019, Ionescu-ICCV-2017} on the Subway data set. All results are copied from the corresponding papers, except for those marked with asterisk (which are computed by ourselves using the official implementations). The best results are highlighted in bold.}
\label{table_subway_sota}
\vspace{-0.3cm}
\centering
\begin{tabular}{|l|c|c|c|c|c|c|c|c|c|}
\hline 
           & \multicolumn{4}{|c|}{Exit} & \multicolumn{4}{|c|}{Entrance} \\
\cline{2-9} 
Method           & \multicolumn{2}{|c|}{AUC}    &  RBDC    & TBDC  & \multicolumn{2}{|c|}{AUC}     &  RBDC    & TBDC \\
 
\cline{2-3}     \cline{6-7}                      &     Old labels          &       New labels        &       &         &         Old labels          &       New labels        &                   &\\
\hline  
Cong et al.~\cite{Cong-CVPR-2011}                &     $83.0$        &       -          &       -           & -        &     $80.0$        &       -          &       -           & -       \\  
\hline
Saligrama et al.~\cite{Saligrama-CVPR-2012}      &     -          &       -          &       -           & -         &      $89.1$        &       -          &       -           & -       \\   
\hline 
Cheng et al.~\cite{Cheng-CVPR-2015}              &     -          &       -          &       -           & -         &      $92.7$        &       -          &       -           & -       \\   
\hline
Hasan et al.~\cite{Hasan-CVPR-2016}              &     $80.7$        &       -          &       -           & -        &     $\mathbf{94.3}$        &       -          &       -           & -       \\  
\hline
Del Giorno et al.~\cite{Giorno-ECCV-2016}        &     $82.4$        &       -          &       -           & -        &     $69.1$        &       -          &       -           & -       \\ 
\hline
Ionescu et al.~\cite{Ionescu-ICCV-2017} &     $86.3$        &       -          &       -           & -        &     $71.3$        &       -          &       -           & -       \\ 
\hline
Wu et al.~\cite{Wu-TNNLS-2019}                   &     $89.5$        &       -          &       -           & -        &     $91.1$        &       -          &       -           & -       \\    
\hline
Ionescu et al.~\cite{Ionescu-WACV-2019}          &     $\mathbf{95.1}$        &       $92.8$*       &       $23.85$*       & $52.62$*    &     $93.5$        &       $82.4$*       &       $23.15$*       & $49.90$*    \\  
\hline 
Ionescu et al.~\cite{Ionescu-CVPR-2019}          &     $92.8$*        &       $91.5$*       &       $43.07$*       & $62.34$*    &     $82.1$*        &       $84.3$*       &       $53.84$*       & $66.41$*    \\  
\hline 
Pang et al.~\cite{Pang-CVPR-2020}                &     $92.7$        &       -          &       -           & -        &     $88.1$        &       -          &       -           & -       \\    
\hline
{Ours}                                    &    $92.1$        &       $\mathbf{93.7}$       &       $\mathbf{47.95}$       &  $\mathbf{67.96}$   &     $87.6$       &       $\mathbf{92.2}$       &       $\mathbf{64.42}$       &  $\mathbf{76.97}$   \\
\hline
\end{tabular}
\vspace{-0.3cm}
\end{table*}

We further compare our method with the state-of-the-art approaches~\cite{Hasan-CVPR-2016, Luo-ICCV-2017, Liu-CVPR-2018, Sultani-CVPR-2018, Doshi-CVPRW-2020a, Doshi-CVPRW-2020b, Tang-PRL-2020, Dong-Access-2020, Park-CVPR-2020, Lee-TIP-2019, Ionescu-CVPR-2019, Sun-ACMMM-2020, Wang-ACMMM-2020, Yu-ACMMM-2020} on the ShanghaiTech data set, presenting the corresponding results in Table \ref{table_shanghai_sota}. The state-of-the-art results in terms of the micro-averaged and the macro-averaged frame-level AUC are attained by our previous method~\cite{Ionescu-CVPR-2019}. The method proposed in the current work outperforms its previous version~\cite{Ionescu-CVPR-2019} by a margin of $4\%$ in terms of the micro-averaged AUC and a margin of $4.4\%$ in terms of the macro-averaged AUC, respectively. All other methods attain lower frame-level AUC scores, ranging from $60.9\%$ to $79.3\%$. We note that there are no previous works reporting RBDC and TBDC scores on ShanghaiTech, since we are the first to provide the necessary region-level and track-level annotations for this data set. Nevertheless, we compute the RBDC and TBDC scores for our previous method proposed in~\cite{Ionescu-CVPR-2019} and for Liu et al.~\cite{Liu-CVPR-2018}. As on Avenue, we had to add a post-processing step to significantly improve the RBDC and TBDC scores of Liu et al.~\cite{Liu-CVPR-2018}. Our current method surpasses its earlier version by more than $20\%$ in terms of RBDC. The improvement is even higher in terms of TBDC, the difference being $34.25\%$ in favor of the current method. Compared to Liu et al.~\cite{Liu-CVPR-2018}, our improvements are higher than $24\%$ for both RBDC and TBDC metrics.
This demonstrates that our method is able to better localize the anomalies, having a lower rate of false positives per frame. Since ShanghaiTech is one of the largest anomaly detection data sets, we consider our results reported in Table \ref{table_shanghai_sota} as noteworthy.

\begin{figure}[!t]
\begin{center}
\includegraphics[width=1.0\linewidth]{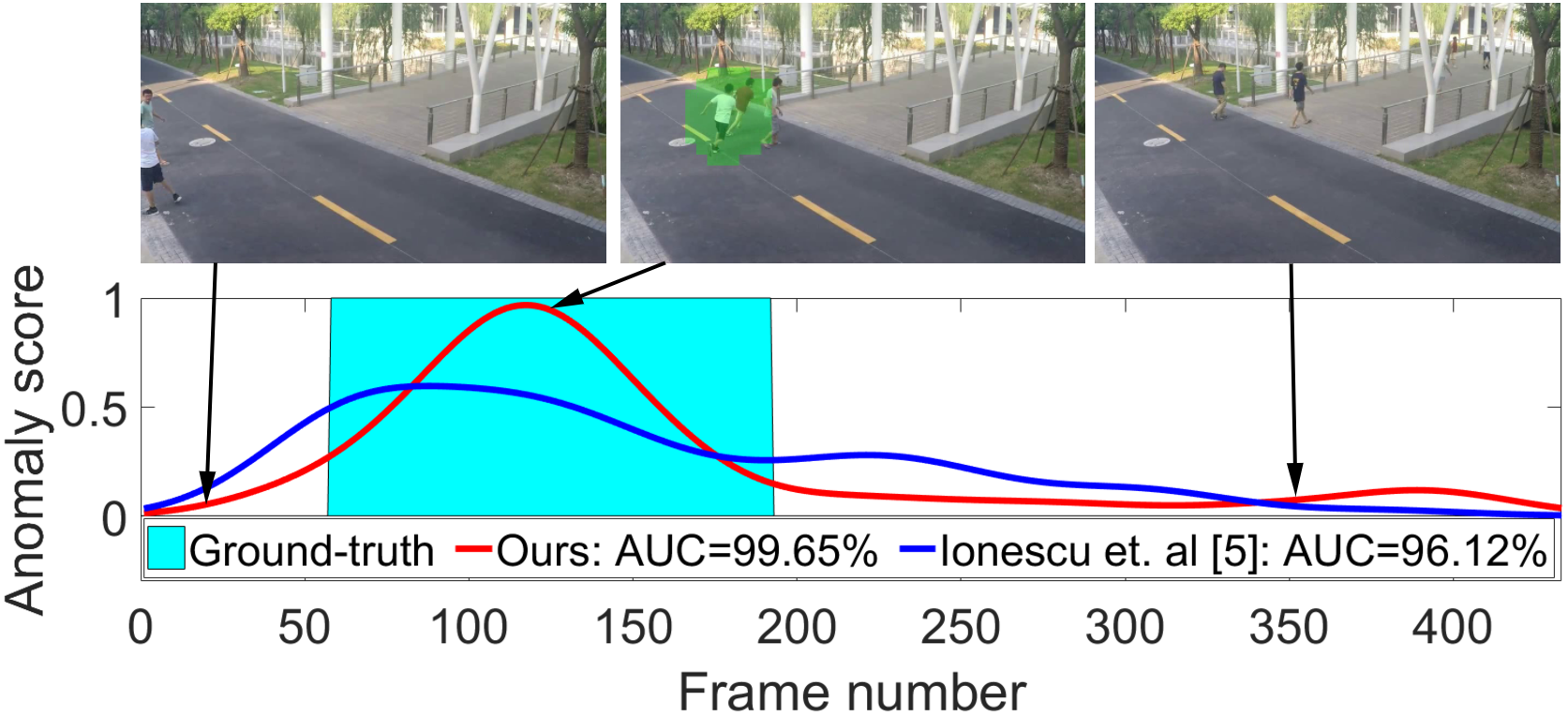}
\end{center}
\vspace{-0.5cm}
\caption{Frame-level anomaly scores (on the vertical axis) provided by our current approach versus the earlier version proposed in~\cite{Ionescu-CVPR-2019}, for test video 05\_0024 from ShanghaiTech~\cite{Luo-ICCV-2017}. Ground-truth abnormal events are represented in cyan, our scores are depicted in red and the scores of the earlier method are depicted in blue. Best viewed in color.}
\label{fig_shanghai}
\vspace{-0.2cm}
\end{figure}

In Figure~\ref{fig_shanghai}, we display the frame-level anomaly scores (corresponding to a frame-level AUC of $99.65\%$) of our method against the ground-truth labels on a ShanghaiTech test video with one abnormal event. On this video, we can clearly observe a strong correlation between our anomaly scores and the ground-truth labels. The preliminary framework proposed in~\cite{Ionescu-CVPR-2019} produces anomaly scores that are less correlated to the ground-truth, its frame-level AUC for the selected test video being $96.12\%$.

\vspace{-0.2cm}
\subsection{Quantitative Results on Subway}

The Subway data set was originally annotated at the frame-level only. Therefore, in order to determine the ground-truth regions and tracks, we first had to annotate the data set at the pixel-level. During the manual annotation process, our annotators observed that there are some frames that were labeled as abnormal, but those frames did not contain any abnormal objects or events. In these circumstances, many false positive frames were counted as correct detections in previous works, which is wrong. In order to rectify this problem, we also relabeled the data set at the frame-level based on the identified abnormal regions, such that, if there is an abnormal region in a frame, the respective frame is considered anomalous. We note that our new labels were subject to the agreement of two independent annotators. Hence, we consider our new labels to be more accurate that the original ones.

In order to compare our work with all previous works \cite{Cheng-CVPR-2015, Hasan-CVPR-2016, Ionescu-ICCV-2017, Ionescu-CVPR-2019, Ionescu-WACV-2019, Giorno-ECCV-2016, Wu-TNNLS-2019, Saligrama-CVPR-2012, Pang-CVPR-2020, Cong-CVPR-2011}, we report the performance obtained using the original frame labels. Since we have access to the exact implementation of the top scoring method on the Subway data set, namely that of Ionescu et al.~\cite{Ionescu-WACV-2019}, we were able to compute the frame-level AUC, the RBDC and TBDC scores for the respective method. In addition, we compute the results of our preliminary framework \cite{Ionescu-CVPR-2019} on Subway. The comparative results are reported in Table~\ref{table_subway_sota}. As the Subway data set contains only one testing video per scene, the micro-averaged frame-level AUC is equivalent to the macro-averaged frame-level AUC. 

\begin{figure}[!t]
\begin{center}
\includegraphics[width=1.0\linewidth]{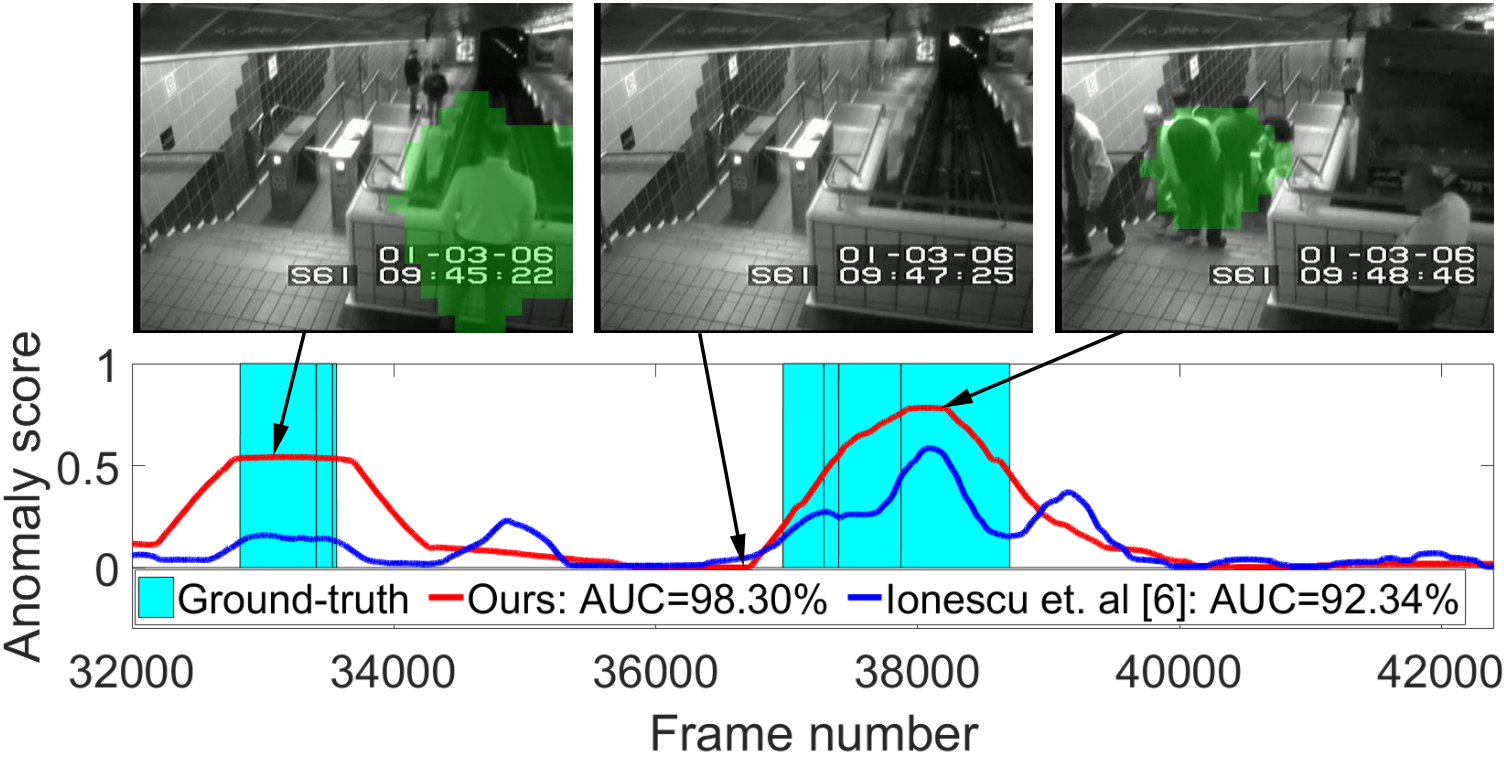}
\end{center}
\vspace{-0.5cm}
\caption{Frame-level anomaly scores (on the vertical axis) provided by our approach versus the approach of Ionescu et al.~\cite{Ionescu-WACV-2019}, for a chunk of video trimmed out from Subway Exit. Ground-truth abnormal events are represented in cyan, our scores are depicted in red and the scores of Ionescu et al.~\cite{Ionescu-WACV-2019} are depicted in blue. Best viewed in color.}
\label{fig_exit}
\vspace{-0.2cm}
\end{figure}

\noindent
\textbf{Results on Subway Exit.}
Considering the old labels, on the Exit video, the state-of-the-art result of $95.1\%$ in terms of the frame-level AUC is obtained by Ionescu et al.~\cite{Ionescu-WACV-2019}, our method being able to achieve the fourth best score of $92.1\%$. Even though our results are lower than those of Ionescu et al.~\cite{Ionescu-WACV-2019} on the original labels, when we switch to the new ones, we outperform their method by $0.9\%$. The same happens when we compare our current method with its previous version \cite{Ionescu-CVPR-2019}.
In terms of RBDC and TBDC, we surpass the state-of-the-art method of Ionescu et al.~\cite{Ionescu-WACV-2019} by very large margins, namely $24.1\%$ in terms of RBDC and $15.34\%$ in terms of TBDC, respectively. Our RBDC and TBDC scores are also higher than those of the preliminary version \cite{Ionescu-CVPR-2019}, the differences being around $5\%$ in favor of our method. 

We compare our frame-level anomaly scores (corresponding to a frame-level AUC of $98.30\%$) against the ground-truth labels on a chunk of the Exit test video in Figure~\ref{fig_exit}. There are several abnormal events, which seem to be grouped into two temporal clusters. Our approach correctly identifies both groups of abnormal events, without false positives. As reference, the frame-level scores produced by the method of Ionescu et al.~\cite{Ionescu-WACV-2019} are also included in Figure~\ref{fig_exit}. While the approach of Ionescu et al.~\cite{Ionescu-WACV-2019} seems to identify both clusters of abnormal events, it also produces high anomaly scores for some normal frames, thus having a higher false positive rate compared to our approach.

\noindent
\textbf{Results on Subway Entrance.}
Considering the frame-level AUC on the old labels for the Entrance video, it seems that our method is surpassed by many other approaches. Hasan et al.~\cite{Hasan-CVPR-2016} attained the state-of-the-art score of $94.3\%$, being closely followed by Ionescu et al.~\cite{Ionescu-WACV-2019} with a score of $93.5\%$. 
Although our frame-level AUC score is lower on the old labels, when considering the new labels, we report an improvement of over $10\%$ compared to Ionescu et al.~\cite{Ionescu-WACV-2019}. We also obtain superior results in terms of RBDC and TBDC, respectively. Our approach outperforms the method presented in~\cite{Ionescu-WACV-2019} by $41.27\%$ in terms of RBDC, being able to detect many more regions with a lower rate of false positives. With a TBDC of $76.97\%$, we surpass the method of Ionescu et al.~\cite{Ionescu-WACV-2019} by more than $27\%$. When compared to its previous version \cite{Ionescu-CVPR-2019}, the proposed framework attains superior results, regardless of the metric. Notably, we observe RBDC and TBDC improvements higher than $10\%$.

\begin{figure}[!t]
\begin{center}
\includegraphics[width=1.0\linewidth]{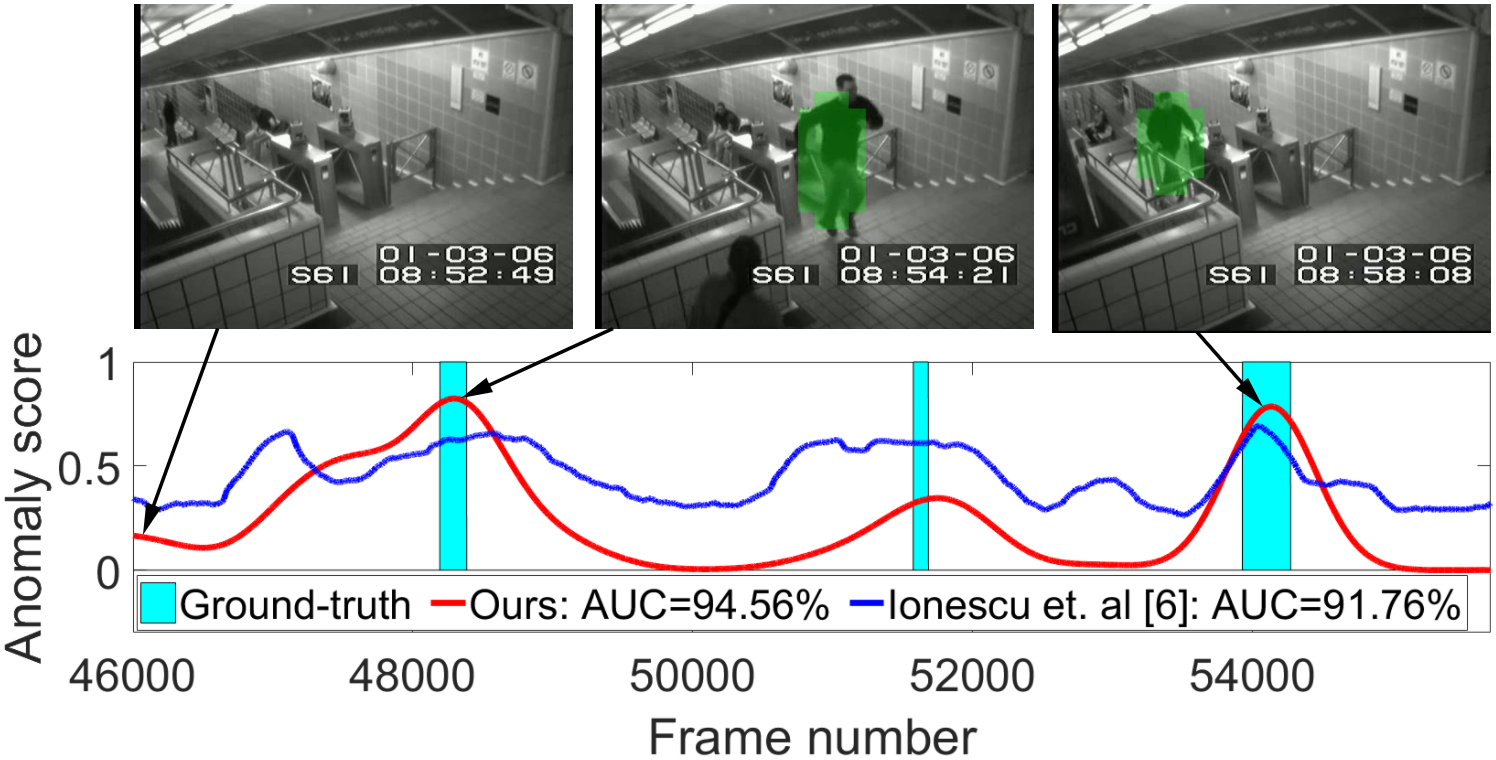}
\end{center}
\vspace{-0.5cm}
\caption{Frame-level anomaly scores (on the vertical axis) provided by our approach versus the approach of Ionescu et al.~\cite{Ionescu-WACV-2019}, for a chunk of video trimmed out from Subway Entrance. Ground-truth abnormal events are represented in cyan, our scores are depicted in red and the scores of Ionescu et al.~\cite{Ionescu-WACV-2019} are depicted in blue. Best viewed in color.}
\label{fig_entrance}
\vspace{-0.2cm}
\end{figure}

As for Subway Exit, we select a chunk of the Subway Entrance test video to compare our frame-level anomaly scores to those of Ionescu et al.~\cite{Ionescu-WACV-2019} as well as to the ground-truth labels, illustrating the comparison in Figure~\ref{fig_entrance}. There are three abnormal events in the selected chunk and our method provides peak anomaly scores for all three events, attaining a frame-level AUC of $94.56\%$ on the selected chunk of video. However, our method also provides high anomaly scores for some frames before the first abnormal event. Nevertheless, the method of Ionescu et al.~\cite{Ionescu-WACV-2019} exhibits high anomaly scores for many normal frames, thus having a much higher false positive rate. Consequently, the frame-level AUC obtained by Ionescu et al.~\cite{Ionescu-WACV-2019} is only $91.76\%$.

\subsection{Quantitative Results on UCSD Ped2}

\begin{table}[!t]
\renewcommand{\arraystretch}{1.1} 
\caption{Micro-averaged AUC, macro-averaged AUC, RBDC and TBDC scores (in $\%$) of our approach compared to the state-of-the-art methods~\cite{Kim-CVPR-2009, Mehran-CVPR-2009, Mahadevan-CVPR-2010, Hasan-CVPR-2016, Zhang-PR-2016, Ionescu-ICCV-2017, Luo-ICCV-2017, Ravanbakhsh-ICIP-2017, Xu-CVIU-2017, Liu-CVPR-2018, Liu-BMVC-2018, Ravanbakhsh-WACV-2018, Gong-ICCV-2019, Ionescu-CVPR-2019, Lee-TIP-2019, Nguyen-ICCV-2019, Dong-Access-2020, Doshi-CVPRW-2020a,Doshi-CVPRW-2020b, Ji-IJCNN-2020, Lu-ECCV-2020, Park-CVPR-2020, Ramachandra-WACV-2020a, Ramachandra-WACV-2020b, Tang-PRL-2020, Yu-ACMMM-2020, Zaheer-CVPR-2020} on the UCSD Ped2 data set. When it is unclear if the reported frame-level AUC is micro-averaged or macro-averaged, we placed the score in the middle. All results are copied from the corresponding papers, except for those marked with asterisk (which are computed by ourselves using the official implementations). The best results are highlighted in bold.}
\label{table_ped2_sota}
\vspace{-0.3cm}
\centering
\begin{tabular}{|l|c|c|c|c|}
\hline
Method                & \multicolumn{2}{|c|}{AUC}          &  RBDC      & TBDC \\

\cline{2-3}                                                &      Micro   & Macro  & &     \\
\hline
Kim et al.~\cite{Kim-CVPR-2009}                   &     \multicolumn{2}{|c|}{$69.3$}     &       -         & -       \\
\hline
Mehran et al.~\cite{Mehran-CVPR-2009}             &     \multicolumn{2}{|c|}{$55.6$}     &       -         & -       \\
\hline
Mahadevan et al.~\cite{Mahadevan-CVPR-2010}       &     \multicolumn{2}{|c|}{$82.9$}     &       -         & -       \\
\hline
Hasan et al.~\cite{Hasan-CVPR-2016}               &     \multicolumn{2}{|c|}{$90.0$}     &       -         & -       \\
\hline
Zhang et al.~\cite{Zhang-PR-2016}                 &     \multicolumn{2}{|c|}{$91.0$}     &       -         & -       \\
\hline
Ionescu et al.~\cite{Ionescu-ICCV-2017}           &    $82.2$  & -  &       -         & -       \\
\hline
Luo et al.~\cite{Luo-ICCV-2017}                   &     $92.2$  & -      &       -         & -       \\
\hline
Ravanbakhsh et al.~\cite{Ravanbakhsh-ICIP-2017}   &     \multicolumn{2}{|c|}{$93.5$}     &       -         & -       \\
\hline
Xu et al.~\cite{Xu-CVIU-2017}                     &     \multicolumn{2}{|c|}{$90.8$}     &       -         & -       \\
\hline
Liu et al.~\cite{Liu-CVPR-2018}                   &     $95.4$   &  $98.1$* &       $38.34$*         & $56.76$*       \\
\hline
Liu et al.~\cite{Liu-BMVC-2018}                   &     $87.5$   &   -     &       -         & -       \\
\hline
Ravanbakhsh et al.~\cite{Ravanbakhsh-WACV-2018}   &     \multicolumn{2}{|c|}{$88.4$}     &       -         & -       \\
\hline
Gong et al.~\cite{Gong-ICCV-2019}                 &     $94.1$  & -     &       -         & -       \\ 
\hline
Ionescu et al.~\cite{Ionescu-CVPR-2019}           &      $94.3$*        &        $97.8$   &       $52.76$*   &  $72.88$*      \\ 
\hline 
Lee et al.~\cite{Lee-TIP-2019}                    &     \multicolumn{2}{|c|}{$96.6$}     &       -         & -       \\
\hline 
Nguyen  et al.~\cite{Nguyen-ICCV-2019}              &     $96.2$  & - &       -         & -       \\
\hline
Dong  et al.~\cite{Dong-Access-2020}                &     \multicolumn{2}{|c|}{$95.6$}     &       -         & -       \\
\hline
Doshi  et al.~\cite{Doshi-CVPRW-2020a,Doshi-CVPRW-2020b}&     \multicolumn{2}{|c|}{$97.8$}  &       -         & -       \\
\hline
Ji  et al.~\cite{Ji-IJCNN-2020}                     &     \multicolumn{2}{|c|}{$98.1$}     &       -         & -       \\
\hline
Lu  et al.~\cite{Lu-ECCV-2020}                      &     \multicolumn{2}{|c|}{$96.2$}     &       -         & -       \\
\hline
Park  et al.~\cite{Park-CVPR-2020}                  &       -    &                  $97.0$ &        -         & -       \\
\hline
Ramachandra  et al.~\cite{Ramachandra-WACV-2020a}   &     \multicolumn{2}{|c|}{$88.3$}     &       $62.50$    &  $80.50$     \\
\hline
Ramachandra  et al.~\cite{Ramachandra-WACV-2020b}   &     \multicolumn{2}{|c|}{$94.0$}     &       $\mathbf{74.00}$    &  $89.30$     \\
\hline
Tang  et al.~\cite{Tang-PRL-2020}                   &     \multicolumn{2}{|c|}{$96.3$}     &       -         & -       \\
\hline
Yu  et al.~\cite{Yu-ACMMM-2020}                     &     $97.3$    & -   &       -         & -       \\
\hline
Zaheer  et al.~\cite{Zaheer-CVPR-2020}              &    $98.1$   &  -    &       -         & -       \\
\hline
{Ours}              &      $\mathbf{98.7}$        &        $\mathbf{99.7}$                &       $69.23$   &  $\mathbf{93.15}$      \\ 
\hline
\end{tabular}
\vspace{-0.3cm}
\end{table}

UCSD Ped2 is one of the most popular benchmarks in video anomaly detection, with a broad range of methods reporting results on it~\cite{Kim-CVPR-2009, Mehran-CVPR-2009, Mahadevan-CVPR-2010, Hasan-CVPR-2016, Zhang-PR-2016, Ionescu-ICCV-2017, Luo-ICCV-2017, Ravanbakhsh-ICIP-2017, Xu-CVIU-2017, Liu-CVPR-2018, Liu-BMVC-2018, Ravanbakhsh-WACV-2018, Gong-ICCV-2019, Ionescu-CVPR-2019, Lee-TIP-2019, Nguyen-ICCV-2019, Dong-Access-2020, Doshi-CVPRW-2020a,Doshi-CVPRW-2020b, Ji-IJCNN-2020, Lu-ECCV-2020, Park-CVPR-2020, Ramachandra-WACV-2020a, Ramachandra-WACV-2020b, Tang-PRL-2020, Yu-ACMMM-2020, Zaheer-CVPR-2020}. We compare the results of our approach with the results reported in literature in Table~\ref{table_ped2_sota}. Our preliminary approach \cite{Ionescu-CVPR-2019} and five other recent methods \cite{Doshi-CVPRW-2020a,Ji-IJCNN-2020,Park-CVPR-2020,Yu-ACMMM-2020,Zaheer-CVPR-2020} obtain frame-level scores higher than $97\%$. Nonetheless, our framework outperforms all these methods, reaching a micro-averaged frame-level AUC of $98.7\%$ and a macro-averaged frame-level AUC of $99.7\%$. In addition to~\cite{Ramachandra-WACV-2020a, Ramachandra-WACV-2020b}, we report the RBDC and TBDC scores for our previous method \cite{Ionescu-CVPR-2019} and for Liu et al.~\cite{Liu-CVPR-2018}. Once again, we had to add a post-processing step to significantly improve the RBDC and TBDC scores of Liu et al.~\cite{Liu-CVPR-2018}. In terms of RBDC, our method outperforms Liu et al.~\cite{Liu-CVPR-2018}, Ramachandra et al.~\cite{Ramachandra-WACV-2020a}, as well as our previous method \cite{Ionescu-CVPR-2019}. The same can be said about the TBDC scores of these methods. In terms of RBDC and TBDC, our method is somewhat comparable to that of Ramachandra et al.~\cite{Ramachandra-WACV-2020b}, the former method attaining a superior TBDC score, while the latter one yielding a better RBDC score. An advantage of our method is its superiority over Ramachandra et al.~\cite{Ramachandra-WACV-2020b} in terms of the frame-level AUC.

\vspace{-0.2cm}
\subsection{Cross-Database Quantitative Results}
\label{sec_crossdb}

In order to demonstrate that our framework is indeed scene-agnostic, we conduct cross-database experiments considering seven pairs of data sets, namely ShanghaiTech$\rightarrow$Avenue, Avenue$\rightarrow$ShanghaiTech, Avenue$\rightarrow$UCSD Ped2, ShanghaiTech$\rightarrow$UCSD Ped2, Avenue$\rightarrow$Subway Exit, UCSD Ped2$\rightarrow$Subway Exit and Avenue + UCSD Ped2$\rightarrow$Subway Exit. The corresponding results are presented in Table~\ref{tab_cross_db}. We consider our in-domain results as upper bounds for the cross-database results.

\noindent
\textbf{ShanghaiTech$\rightarrow$Avenue.} 
As expected, the performance of our method on Avenue is lower when training is performed on ShanghaiTech instead of Avenue. For example, the micro-averaged AUC score degrades from $92.3\%$ to $83.6\%$. Yet, this result is still better than many state-of-the-art results \cite{Lu-ICCV-2013, Hasan-CVPR-2016, Luo-ICCV-2017, Ramachandra-WACV-2020a} (see Table~\ref{table_avenue_sota}) that rely on models trained on the target data set (Avenue). Our cross-domain method also outperforms models that perform change detection at test time~\cite{Giorno-ECCV-2016,Ionescu-ICCV-2017}, which do not require any training data. In terms of RBDC, our cross-domain approach outperforms the in-domain method of Ramachandra et al.~\cite{Ramachandra-WACV-2020a} by $5.36\%$, being able to accurately detect the abnormal regions with a lower false positive rate. Remarkably, the TBDC score of our cross-domain method only drops by $4.09\%$ compared to our in-domain framework. This confirms that the abnormal tracks are still being accurately detected, even though we train our framework on scenes from a different data set.

\begin{table}[!t]
\renewcommand{\arraystretch}{1.1}
\setlength\tabcolsep{3.5pt}
\caption{Micro-averaged AUC, macro-averaged AUC, RBDC and TBDC scores (in $\%$) of our method for a series of cross-domain experiments. For each experiment, we specify the training and the test data sets.}
\label{tab_cross_db}
\vspace{-0.3cm}
\centering
\begin{tabular}{|l|c|c|c|c|}
\hline
Training$\rightarrow$Test         & \multicolumn{2}{|c|}{AUC}          &  RBDC      & TBDC \\
\cline{2-3} 
    & Micro         & Macro  & &     \\
\hline
Avenue$\rightarrow$Avenue       &      $92.3$        &        $90.4$   &       $65.05$   &  $66.85$      \\ 
\cline{2-5} 
ShanghaiTech$\rightarrow$Avenue       &      $83.6$        & 	      $81.0$   &       $46.56$	 &   $62.76$      \\
\hline
ShanghaiTech$\rightarrow$ShanghaiTech &     $82.7$         &        $89.3$   &       $41.34$   &  $78.79$   \\
\cline{2-5}
Avenue$\rightarrow$ShanghaiTech &      $76.3$        &      	$86.3$	&     $32.55$	 &   $63.89$       \\ 
\hline
UCSD Ped2$\rightarrow$UCSD Ped2    &      $98.7$        &        $99.7$    &       $69.23$   &  $93.15$      \\ 
\cline{2-5}
Avenue$\rightarrow$UCSD Ped2     &      $87.0$        &        $97.2$    &       $47.43$   &  $68.58$      \\ 
\cline{2-5}
Shanghai$\rightarrow$UCSD Ped2     &      $90.6$        &        $95.7$    &       $41.71$   &  $70.49$      \\ 
\hline
\hline
Training$\rightarrow$Test         & \multicolumn{2}{|c|}{AUC}          &  RBDC      & TBDC \\
\cline{2-3}
     & Old         & New  & &     \\
\hline
Subway Exit$\rightarrow$Subway Exit 	& $92.1$     &  $93.7$       	&  $47.95$       &  $67.96$ \\ 
\cline{2-5}
Avenue$\rightarrow$Subway Exit			& $92.1$		& $92.4$			&	$47.54$		& $63.93$ \\
\cline{2-5}
UCSD Ped2$\rightarrow$Subway Exit			& $92.8$		& $92.8$			&	$46.60$		& $65.73$ \\
\cline{2-5}
Avenue+UCSD Ped2$\rightarrow$Subway Exit			& $92.1$		& $92.3$			&	$47.50$		& $64.18$ \\
\hline
\end{tabular}
\vspace{-0.3cm}
\end{table}

\begin{figure*}[!t]
\begin{center}
\includegraphics[width=1.0\linewidth]{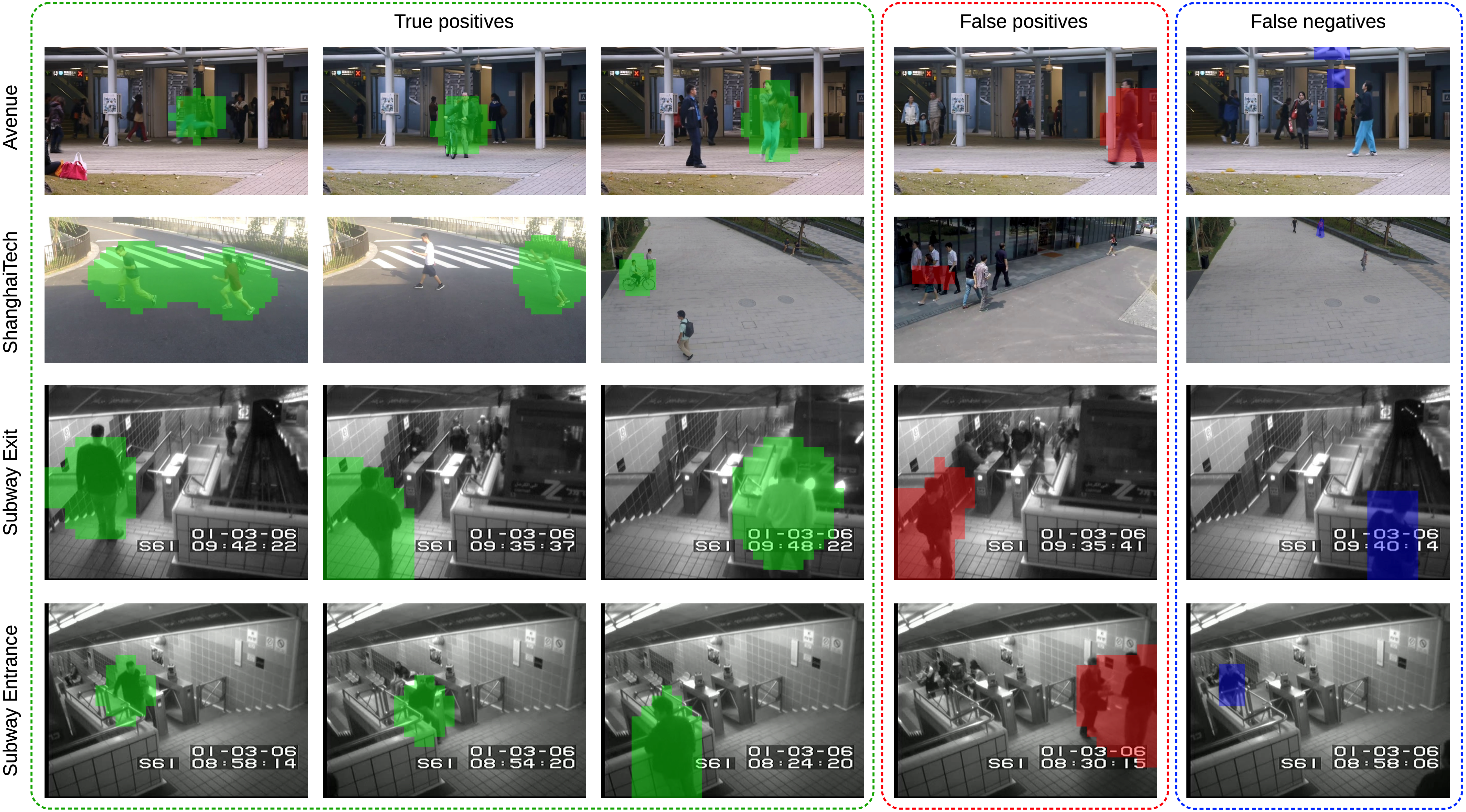}
\end{center}
\vspace{-0.5cm}
\caption{True positive (left) versus false positive (middle) and false negative (right) detections of our framework. Examples are selected from the Avenue~\cite{Lu-ICCV-2013} (first row), the ShanghaiTech~\cite{Luo-ICCV-2017} (second row), the Subway Exit~\cite{Adam-PAMI-2008} (third row) and the Subway Entrance~\cite{Adam-PAMI-2008} (fourth row) data sets. Best viewed in color.}
\label{fig_examples}
\vspace{-0.2cm}
\end{figure*}

\noindent
\textbf{Avenue$\rightarrow$ShanghaiTech.}
In the second cross-domain experiment, we train our framework on the Avenue data set, evaluating it on the ShanghaiTech data set. 
Even though ShanghaiTech is much larger than Avenue, we are still able to obtain compelling results. In terms of the micro-averaged frame-level AUC, our cross-domain method surpasses most of the state-of-the-art methods \cite{Hasan-CVPR-2016, Luo-ICCV-2017, Lee-TIP-2019, Doshi-CVPRW-2020a, Tang-PRL-2020,Dong-Access-2020} (see Table~\ref{table_shanghai_sota}), which were trained on the ShanghaiTech training set. In terms of RBDC, we observe a performance degradation of only $8.79\%$ for our cross-domain method compared to its in-domain version. In terms of TBDC, our performance decreases from $78.79\%$ to $63.89\%$, but we still outperform the state-of-the-art method of Ionescu et al.~\cite{Ionescu-CVPR-2019} by a significant margin of $19.35\%$. 

\noindent
\textbf{Avenue$\rightarrow$UCSD Ped2.}
When we train our method on Avenue and test it on UCSD Ped2, we observe significant performance drops in terms of RBDC and TBDC with respect to the in-domain baseline. The performance degradation is likely caused by camera viewpoint, frame rate and frame resolution discrepancies between Avenue and UCSD Ped2. Still, the RBDC and TBDC scores of our cross-domain framework are very close to those of the previous version of our in-domain method \cite{Ionescu-CVPR-2019} (see Table~\ref{table_ped2_sota}). Furthermore, the macro-averaged AUC only drops by $2.5\%$ with respect to the upper bound ($99.7\%$) provided by our in-domain method. In terms of the frame-level AUC scores, our cross-domain model outperforms in-domain methods such as \cite{Kim-CVPR-2009, Mehran-CVPR-2009, Mahadevan-CVPR-2010, Ionescu-ICCV-2017, Park-CVPR-2020} (see Table~\ref{table_ped2_sota}).

\noindent
\textbf{ShanghaiTech$\rightarrow$UCSD Ped2.}
When we train our framework on ShanghaiTech and test it on UCSD Ped2, we notice that our cross-domain method exhibits performance drops similar to those observed for the Avenue$\rightarrow$UCSD Ped2 experiment, as the differences between ShanghaiTech and UCSD Ped2 are roughly the same. This time, our micro-averaged frame-level AUC is slightly higher ($90.6\%$), leading to superior performance with respect to in-domain methods such as \cite{Kim-CVPR-2009, Mehran-CVPR-2009, Mahadevan-CVPR-2010, Hasan-CVPR-2016, Liu-BMVC-2018, Ionescu-ICCV-2017, Ravanbakhsh-WACV-2018,Ramachandra-WACV-2020a} (see Table~\ref{table_ped2_sota}).

\noindent
\textbf{Avenue$\rightarrow$Subway Exit.} In terms of frame rate and frame resolution, Avenue and Subway Exit are very well aligned. Hence, it comes as no surprise that, when we train our background-agnostic framework on Avenue and apply it on Subway Exit, we observe generally low performance reductions with respect to the in-domain framework, the only exception being the larger TBDC drop from $67.96\%$ to $63.93\%$. Still, the RBDC and TBDC scores reported for the Avenue$\rightarrow$Subway Exit experiment are higher than the scores of the state-of-the-art in-domain methods~\cite{Ionescu-CVPR-2019,Ionescu-WACV-2019}.

\noindent
\textbf{UCSD Ped2$\rightarrow$Subway Exit.} The frame resolution discrepancy between Subway Exit and UCSD Ped2 is only slightly larger than the discrepancy between Subway Exit and Avenue. It is therefore natural to expect minor variations between Avenue$\rightarrow$Subway Exit and UCSD Ped2$\rightarrow$Subway Exit. While the frame-level AUC is moderately higher when UCSD Ped2 is used as source data set, we attain a slightly better RBDC score when Avenue is used as source data set. We conclude that both Avenue and UCSD Ped2 data sets provide useful training data for Subway Exit.

\noindent
\textbf{Avenue + UCSD Ped2$\rightarrow$Subway Exit.} To study if we can gain additional performance by combining multiple source data sets, we considered an experiment where Avenue and UCSD Ped2 are jointly used as source data sets and Subway Exit is used as target data set. We observe insignificant performance changes when we put the source data sets together as opposed to considering them as independent source data sets. This result hints that trying to combine normal training data for multiple sources is not always helpful.

\noindent
\textbf{Overall.}
Since our cross-domain method is able to outperform many state-of-the-art methods \cite{Cong-CVPR-2011,Kim-CVPR-2009, Mehran-CVPR-2009, Mahadevan-CVPR-2010, Giorno-ECCV-2016, Ionescu-ICCV-2017, Ramachandra-WACV-2020a, Hasan-CVPR-2016, Luo-ICCV-2017, Lee-TIP-2019, Doshi-CVPRW-2020a, Tang-PRL-2020,Dong-Access-2020, Ionescu-CVPR-2019,Ionescu-WACV-2019,Wu-TNNLS-2019} that are trained on the target data sets, we conclude that our method is background-agnostic, providing good results even when testing is performed on scenes never seen during training. 

\vspace{-0.2cm}
\subsection{Qualitative Results}

Figure~\ref{fig_examples} illustrates a set of true positive, false positive and false negative abnormal event localizations from Avenue, ShanghaiTech, Subway Exit and Subway Entrance. We note that the anomaly maps overlapped over the presented frames were subject to a $3D$ mean filter, hence, the shapes of the detections do not coincide with object bounding boxes.

\noindent
\textbf{Avenue.}
The examples selected from the Avenue data set are illustrated on the first row. From left to right, the true positive detections are \emph{a person running}, \emph{a person walking besides a bike} and \emph{a person throwing an object}. The false positive example is \emph{a person entering the scene} from an unusual location (closer to the camera than usual). The false negative example is represented by \emph{papers thrown in the air}, which are not detected because \emph{paper} is not among the classes known by the pre-trained object detector.

\noindent
\textbf{ShanghaiTech.}
The examples from ShanghaiTech are illustrated on the second row of Figure~\ref{fig_examples}. The true positive abnormal events detected by our framework are (from left to right) \emph{two people running}, \emph{a person jumping} and \emph{a person riding a bike}. The false positive example consists of \emph{two people that are detected in the same bounding box} by the object detector, generating a very unusual motion. The false negative example is \emph{a person riding a skateboard}, which is not detected by YOLOv3 because the person is too small (just entering the scene from the far end).

\noindent
\textbf{Subway Exit.}
The examples from Subway Exit are presented on the third row. From left to right, the first two true positive detections represent \emph{a person walking in the wrong direction}, while the third true positive example is \emph{a person loitering}. The false positive example is \emph{a person crossing the scene from left to right}. The false negative example is \emph{a person loitering}.

\noindent
\textbf{Subway Entrance.}
The examples from Subway Entrance are presented on the last row of Figure~\ref{fig_examples}. From left to right, the first two true positive examples represent \emph{a person jumping over the gate}, while the third true positive example is \emph{a person loitering}. The false positive example is represented by \emph{two people interacting}, one of them facing the wrong direction (according to the definition of abnormal events for Subway Entrance~\cite{Adam-PAMI-2008}). The false negative example is \emph{a person walking in the wrong direction}.

\begin{table*}[!t]
\setlength\tabcolsep{0.8pt}
\renewcommand{\arraystretch}{1.1} 
\caption{Micro-averaged AUC, macro-averaged AUC, RBDC and TBDC scores (in $\%$) obtained by making gradual design changes to our original method presented in~\cite{Ionescu-CVPR-2019}, until the framework converges to our current proposal. Best results are highlighted in bold. Notations: MAE represents the MAE between the CAE input and output; AD is short for absolute differences between the CAE input and output; LF is short for the CAE latent features.}
\label{table_avenue_ablation}
\vspace{-0.3cm}
\centering
\begin{tabular}{|c|c|c|c|c|c|c|c|c|c|c|c|c|c|c|c|}
\hline 
Object        & CAE          & Motion      & CAE         & CAE & CAE & Anomaly          & \multicolumn{4}{|c|}{Avenue} & \multicolumn{4}{|c|}{ShanghaiTech} \\
\cline{8-15}

Detector & Input & Type & Adversarial & Segmentation & Skip & Detection                          & \multicolumn{2}{|c|}{AUC}  &  RBDC    & TBDC 
                                 & \multicolumn{2}{|c|}{AUC}  &  RBDC    & TBDC  \\
\cline{8-9}
\cline{12-13}

        & Type          &       & Branch & Branch & Connections & Method
                                 & Micro & Macro &      &                & Micro & Macro &      &   \\
\hline
SSD-FPN & fusion & gradients & & & & k-means+OVR SVM &     $87.4$     &   $90.4$  &  $15.77$   &  $27.01$      &  $78.70$   &  $84.90$  & 	$20.65$	 &  $44.54$  \\ 

\hline
SSD-FPN & motion & gradients & & & & MAE &     $77.8$     &   $78.8$  &  $16.12$   &  $30.17$      &  $72.20$   &  $79.30$  & 	$24.24$	 &  $49.69$  \\ 

SSD-FPN & appearance & - & & & & MAE &     $77.6$     &   $77.4$  &  $16.77$   &  $29.90$      &  $75.60$	 &  $81.90$  &	$26.30$	 &  $50.44$  \\ 

SSD-FPN & fusion & gradients & & & & MAE &     $78.3$     &   $79.2$  &  $16.42$   &  $30.01$      &  $72.40$	 &  $79.70$  & 	$24.35$	 &  $49.74$ \\ 
\hline 
YOLOv3 & motion & gradients & & & & MAE &     $72.4$     &   $74.6$  &  $47.88$   &  $48.34$      &  $65.60$	 &  $76.80$	 &  $19.11$  &	$44.72$ \\

YOLOv3 & appearance & - & & & & MAE &     $78.3$     &   $76.5$  &  $49.94$   &  $44.22$      &  $67.50$   & 	$78.40$  &	$21.69$  & 	$48.13$   \\  

YOLOv3 & fusion & gradients & & & & MAE &     $74.8$     &   $75.4$  &  $51.49$   &  $48.80$      &  $65.80$	 &  $77.70$	 &  $19.54$  & 	$45.16$ \\ 
\hline 
YOLOv3 & motion & SelFlow & & & & MAE &     $80.3$     &   $81.7$  &  $50.60$   &  $52.65$      &  $69.30$   & 	$79.80$  & 	$22.84$  &	$52.66$   \\  

YOLOv3 & fusion & SelFlow & & & & MAE &     $81.6$     &   $82.3$  &  $52.89$   &  $52.70$      &  $69.80$   &	$80.60$  & 	$22.90$  &	$52.70$\\ 
\hline
YOLOv3 & motion & SelFlow & \checkmark & & & MAE &     $77.6$     &   $81.3$  &  $44.96$   &  $49.66$      &  $69.00$   &	$80.50$  &	$23.23$  &	$52.62$  \\

YOLOv3 & appearance & - & \checkmark & & & MAE &     $76.7$     &   $79.7$  &  $48.84$   &  $45.28$      &  $67.30$	 &  $79.00$  & 	$21.48$	 &  $48.67$   \\

YOLOv3 & fusion & SelFlow & \checkmark & & & MAE &     $80.0$     &   $83.4$  &  $49.98$   &  $51.69$      &  $69.20$   &	$81.20$  &	$23.23$	 &  $52.68$ \\ 
\hline 
YOLOv3 & appearance & - & \checkmark & \checkmark & & MAE &     $77.3$     &   $79.5$  &  $49.86$   &  $43.45$      &  $67.90$   & 	$77.10$  &  $20.67$  & 	$48.29$   \\  

YOLOv3 & fusion & SelFlow & \checkmark & \checkmark & & MAE &     $79.9$     &   $83.7$  &  $50.72$   &  $51.38$      &  $69.20$   & 	$81.10$  &	$23.11$  &	$52.71$  \\ 
\hline
YOLOv3 & motion & SelFlow & \checkmark & & & classifier on AD &     $84.8$	  &   $85.7$  &  $46.30$   &  $62.54$      &  $78.80$   &	$87.90$  &	$34.79$  &	$72.97$    \\ 

YOLOv3 & appearance & - & \checkmark & \checkmark & & classifier on AD &     $79.7$ 	  &   $81.5$  &  $36.98$	&  $49.66$      &  $62.30$   &	$75.10$  &	$24.21$  &	$54.28$    \\  

YOLOv3 & fusion & SelFlow & \checkmark & \checkmark & & classifier on AD &     $86.2$	  &   $85.9$  &  $50.68$	&  $63.74$      &  $77.50$   &	$87.70$  & 	$33.78$  &	$69.79$ \\ 
\hline 
YOLOv3 & motion & SelFlow & \checkmark & & & classifier on LF &     $86.5$     &   $86.2$  &  $49.96$   &  $63.10$      &  $81.60$   &	$88.60$  &	$36.43$  & 	$74.31$   \\  
        
YOLOv3 & appearance & - & \checkmark & \checkmark & & classifier on LF &     $89.2$     &   $83.4$  &  $48.13$   &  $53.38$      &  $63.30$   &	$74.10$  &	$30.44$  &	$66.80$   \\ 

YOLOv3 & fusion & SelFlow & \checkmark & \checkmark & & classifier on LF &     $86.6$     &   $88.2$  &  $51.79$   &  $63.26$      &  $81.90$   &	$89.10$  &	$37.31$  & 	$75.33$ \\ 
\hline 
YOLOv3 & motion & SelFlow & \checkmark & & & classifier on AD+LF &     $86.1$     &   $82.5$  &  $45.29$   &  $65.02$      &  $80.00$   & 	$87.30$  &	$37.49$  &	$75.71$    \\

YOLOv3 & appearance & - & \checkmark & \checkmark & & classifier on AD+LF &     $87.4$     &   $83.3$  &  $49.15$   &  $53.23$      &  $64.80$	 &  $77.20$	 &  $31.00$  &  $68.35$   \\ 

YOLOv3 & fusion & SelFlow & \checkmark & \checkmark & & classifier on AD+LF  &     $91.6$     &   $\mathbf{90.6}$  &  $62.11$   &  $\mathbf{67.55}$      &  $82.40$   &	$88.90$	 &  $41.26$  &	$77.29$  \\ 
\hline  
SSD-FPN & fusion & SelFlow & \checkmark & \checkmark & & classifier on AD+LF  &     $82.6$	 & $80.7$  & $21.29$ &	 $34.12$  &   $78.40$ &	$81.70$	& $28.73$ &	$59.10$  \\ 
\hline 
YOLOv3 & motion & SelFlow & \checkmark & & concatenate & classifier on AD+LF   &     $87.4$	 &   $87.2$  &  $52.87$  &  $64.17$     &  $81.00$	&  $87.40$	& $37.24$	&  $75.54$   \\ 
        
YOLOv3 & appearance & - & \checkmark & \checkmark & concatenate & classifier on AD+LF    &    $81.7$ &	$83.7$	  & $46.08$	 &  $49.94$  &  $68.50$	 & $75.80$ &  $33.05$ & $72.37$  \\  
        
YOLOv3 & fusion & SelFlow & \checkmark & \checkmark & concatenate & classifier on AD+LF  &    $90.9$  &  $90.5$  & $64.59$	 &  $65.12$  &  $82.50$ &	$88.30$ &	$38.79$	 & $76.68$ \\
\hline  
YOLOv3 & motion & SelFlow & \checkmark & & sum & classifier on AD+LF   &     $84.8$     &   $84.0$  &  $46.58$   &  $63.57$      &  $81.40$   &	$88.10$  &	$37.78$  &	$76.05$   \\ 
        
YOLOv3 & appearance & - & \checkmark & \checkmark & sum & classifier on AD+LF         &     $83.0$     &   $82.4$  &  $39.57$   &  $52.83$      &  $63.90$   &	$73.50$  &	$33.72$  &	$69.31$   \\  
        
YOLOv3 & fusion & SelFlow & \checkmark & \checkmark & sum & classifier on AD+LF  &     $\mathbf{92.3}$     &   $90.4$  &  $\mathbf{65.05}$   &  $66.85$      &  $\mathbf{82.70}$   &	$\mathbf{89.30}$  & 	$\mathbf{41.34}$  & 	$\mathbf{78.79}$ \\ 
\hline 
\end{tabular}
\vspace{-0.3cm}
\end{table*}

\vspace{-0.2cm}
\subsection{Ablation Study}

We perform an ablation study on the Avenue and the ShanghaiTech data sets to emphasize the effect of each component over the overall performance of our framework. The ablation results are presented in Table~\ref{table_avenue_ablation}. The first row corresponds the method presented in~\cite{Ionescu-CVPR-2019}, which attains good AUC scores, yet seems to underperform in terms of RBDC and TBDC. Our first design change is to eliminate the k-means clustering and the one-versus-rest (OVR) SVM. The model is left with three auto-encoders, the anomaly score being computed as the mean absolute error (MAE) between the input and the reconstruction of each CAE. Although the AUC scores record significant drops with respect to the original method~\cite{Ionescu-CVPR-2019}, we observe some performance improvements in terms of RBDC and TBDC. Our next design change is the replacement of the SSD-FPN \cite{Lin-CVPR-2017} with YOLOv3~\cite{Redmon-arXiv-2018}. This change causes further performance drops on Avenue, in terms of AUC, and on ShanghaiTech, in terms of all metrics. However, YOLOv3 brings significant improvements on Avenue, in terms of RBDC and TBDC. Even though the results obtained with YOLOv3 \cite{Redmon-arXiv-2018} are not very encouraging, we have decided to continue the experiments with YOLOv3\cite{Redmon-arXiv-2018}, as it detects four times more objects than SSD-FPN, while having a lower false positive rate. The models presented so far use the image gradients as input to the motion auto-encoders, as proposed in~\cite{Ionescu-CVPR-2019}. Our next change is to replace image gradients with optical flow maps given by SelFlow~\cite{Liu-CVPR-2019}. This seems to be a very important design change, leading to significant performance gains with respect to all metrics.
We continue our ablation study by considering the integration of an adversarial branch in each auto-encoder, while also switching from conventional training to adversarial training. Through adversarial training, we obtained improvements of around $1\%$ in terms of the macro-averaged frame-level AUC on both data sets. In a similar manner, we updated the architecture and the loss of the appearance CAE such that the model can also output segmentation maps. This change seems to bring some slight improvements on Avenue, in terms of RBDC. Upon integrating the binary classifiers into our framework, we observe major improvements with respect to all performance metrics. We note that training the binary classifiers would not be possible without the adversarial component. Hence, the adversarial component plays an indirect yet important role in our framework, giving us a good reason to keep it. As input for the binary classifiers, we considered three options. Our first option is to use the absolute differences (AD) between inputs and reconstructions of the CAEs, obtaining improvements of more than $12\%$ in terms of TBDC and more than $6\%$ in terms of the micro-averaged frame-level AUC, on both data sets. Our second option, which is based on providing the latent features (LF) as input to the binary classifiers, leads to further performance gains. Our third option is to combine the absolute differences and the latent features. In the combination, the latent features are passed to the binary classifiers through some skip connections, as illustrated in Figure~\ref{fig_pipeline}. By combining the absolute differences and the latent features, we reach the state-of-the-art macro-averaged frame-level AUC of $90.6\%$ on the Avenue data set, surpassing the method presented in~\cite{Ionescu-CVPR-2019}. Interestingly, we note that replacing YOLOv3 back with SSD-FPN, while keeping all the other design changes presented so far, does not seem to be effective, confirming that YOLOv3 is a better choice in the end. Our next design change is to integrate skip connections into the CAEs, drawing our inspiration from U-Net~\cite{Ronneberger-MICCAI-2015}. This design change seems to degrade performance. Our last design change is to integrate skip connections by summing up the corresponding features instead of concatenating them as in U-Net. The resulting micro-averaged frame-level AUC scores of $92.3\%$ on the Avenue data set and $82.7\%$ on the ShanghaiTech data set are state-of-the-art.

\vspace{-0.2cm}
\subsection{Running Time}

Compared to our previous method described in~\cite{Ionescu-CVPR-2019}, one of the most important changes in terms of time is the replacement of SSD-FPN with YOLOv3. For optimal performance, we process the video in mini-batches of 64 frames. Even though YOLOv3~\cite{Redmon-arXiv-2018} is much faster than SSD \cite{Liu-ECCV-2016} with Feature Pyramid Networks \cite{Lin-CVPR-2017}, it still requires about $0.84$ seconds to process $64$ frames, thus running at $72$ frames per second (FPS). The slowest component of our framework is SelFlow~\cite{Liu-CVPR-2019}, which runs at $20$ FPS on mini-batches of $32$ frames. Due to the fact that our CAEs are very light, they require only $1.5$ milliseconds to extract the latent features and to output the reconstruction for one input image. Our binary classifiers are even faster, requiring less than $1$ millisecond to obtain the normality score for one object. Reassembling the anomaly scores of the detected objects into an anomaly map for each frame takes less than $1$ millisecond. Putting all the components together, our framework runs at $18$ FPS with a reasonable average of $5$ objects per frame. 
We note that the reported speed of $18$ FPS is for a sequential processing pipeline on a single thread. This is to fairly compare with other methods from the literature. However, we note that running the pipeline on two threads in parallel increases the speed to $24$ FPS, while still using a single GPU. Hence, our framework can process the video in real time using parallel processing. Nevertheless, we would like to note that $76\%$ of the processing time is spent computing the optical flow. Therefore, one way to further speed up the running time is to replace SelFlow with a faster optical flow predictor. The reported running times were measured on a GeForce GTX 3090 GPU with 24 GB of VRAM.

\vspace{-0.2cm}
\subsection{Impact of Adversarial Training}

Throughout the experiments presented above, we set $\lambda$ (the weight of the reversed gradient) to $0.2$, as recommended in~\cite{McHardy-NAACL-2019}. While this setting relinquishes us from the need to validate $\lambda$ on each data set, we recognize that it may also produce suboptimal results. In order to study the effect of $\lambda$ on the performance level, we present supplementary results on Avenue, ShanghaiTech, Subway Exit and UCSD Ped2, considering values of $\lambda$ between $0$ and $1$ taken at a regular step of $0.1$. In the subsequent experiments, we underline that $\lambda=0$ is equivalent to removing adversarial training altogether.

\begin{figure}[!t]
\begin{center}
\includegraphics[width=0.8\linewidth]{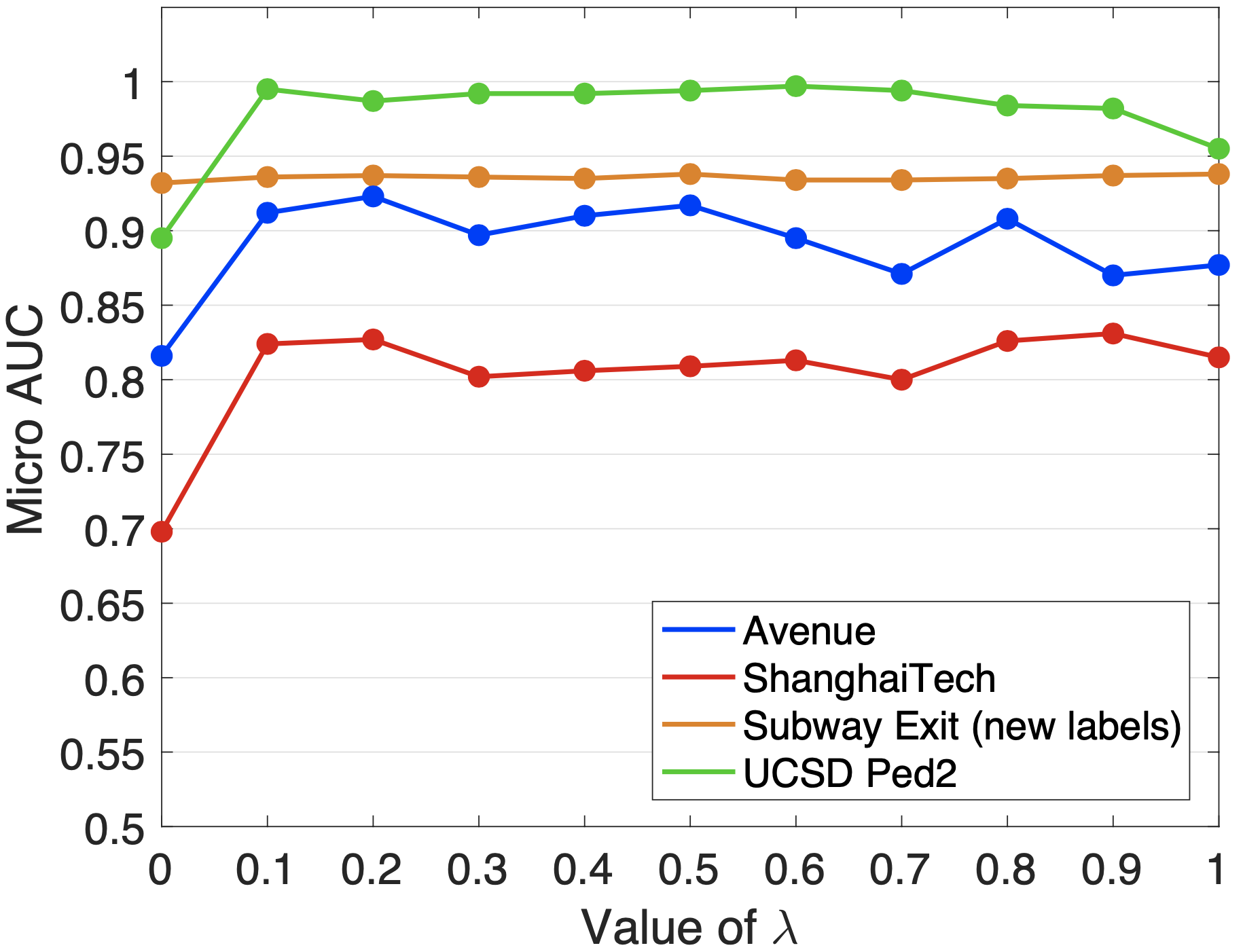}
\end{center}
\vspace{-0.5cm}
\caption{Effect of $\lambda$ (the weight of the reversed gradient) on the micro-averaged frame-level AUC. Best viewed in color.}
\label{fig_lambda_micro}
\vspace{-0.2cm}
\end{figure}

Figure~\ref{fig_lambda_micro} illustrates the effect of $\lambda$ on the micro-averaged frame-level AUC. We observe that $\lambda$ has almost no effect on the Subway Exit data set. For the other data sets, we discover that adversarial training, i.e. when $\lambda\geq0.1$, is instrumental, bringing significant performance gains. However, there is no agreement across data sets regarding the optimal value for $\lambda$. Indeed, $\lambda=0.9$ produces the best results ($83.1\%$) on ShanghaiTech, while $\lambda=0.6$ looks like a better choice for UCSD Ped2, leading to a micro-averaged AUC of $99.7\%$. The recommendation of McHardy et al.~\cite{McHardy-NAACL-2019} to use $\lambda=0.2$ provides the best results ($92.3\%$) only on Avenue.

\begin{figure}[!t]
\begin{center}
\includegraphics[width=0.8\linewidth]{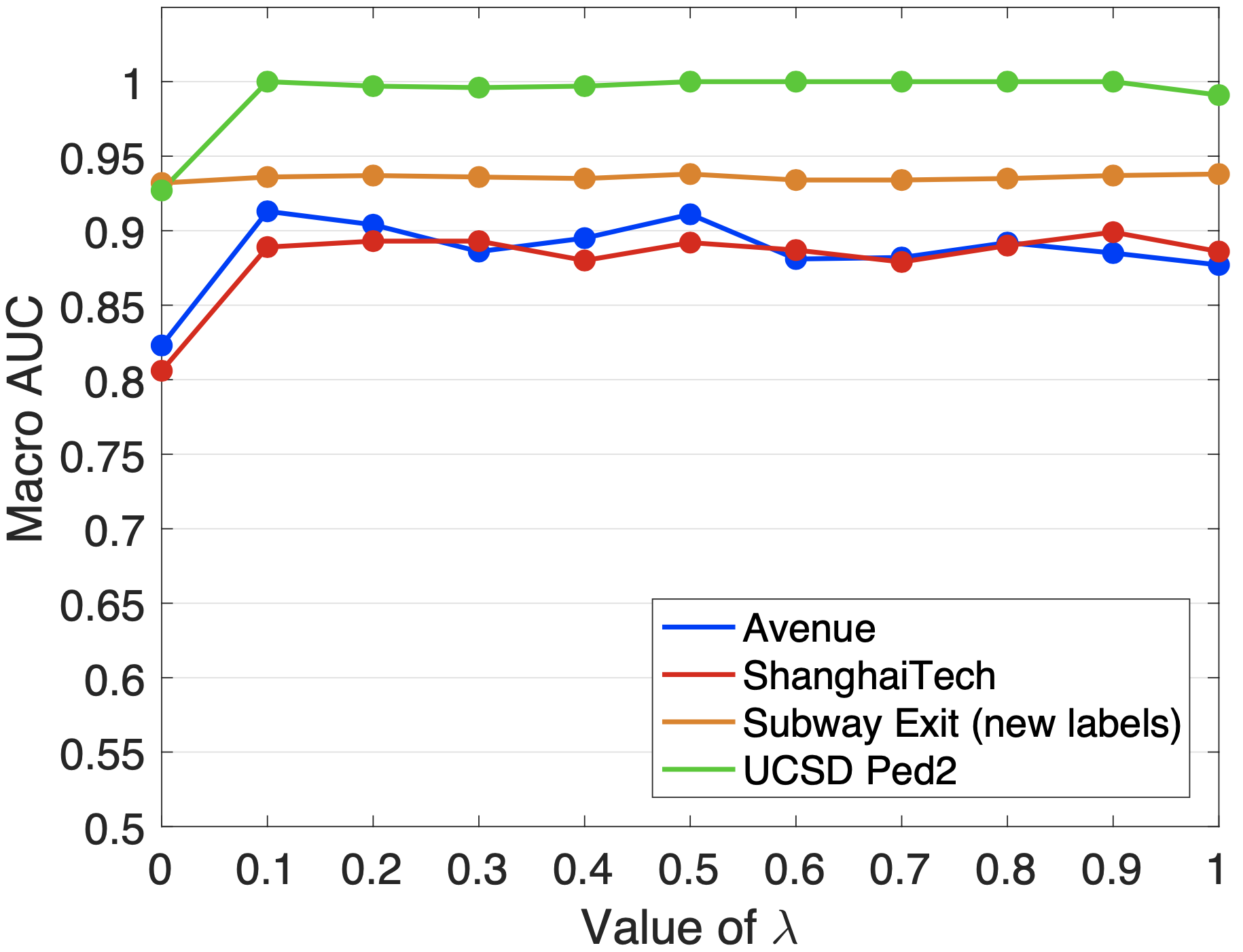}
\end{center}
\vspace{-0.5cm}
\caption{Effect of $\lambda$ (the weight of the reversed gradient) on the macro-averaged frame-level AUC. Best viewed in color.}
\label{fig_lambda_macro}
\vspace{-0.2cm}
\end{figure}

\begin{figure}[!t]
\begin{center}
\includegraphics[width=0.8\linewidth]{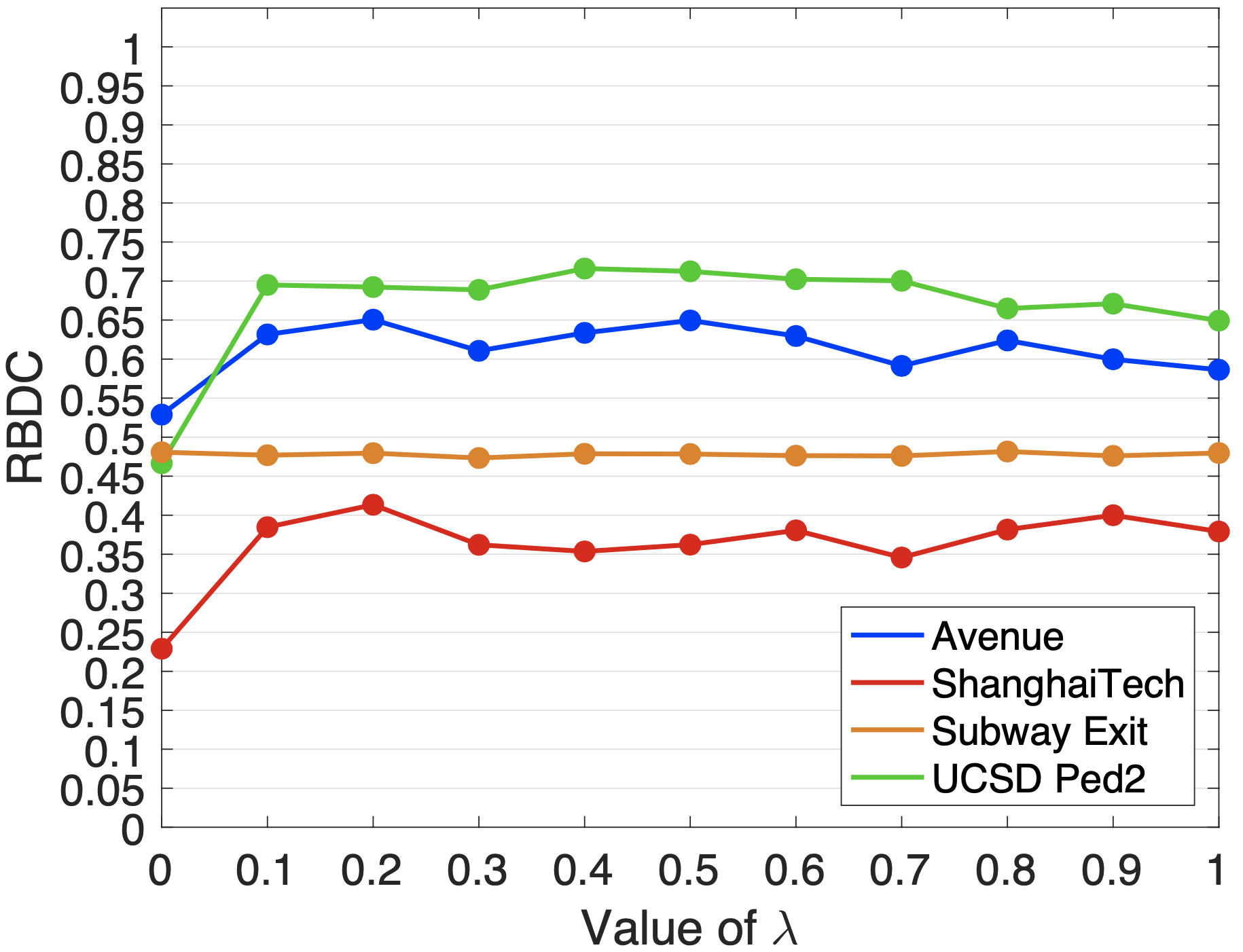}
\end{center}
\vspace{-0.5cm}
\caption{Effect of $\lambda$ (the weight of the reversed gradient) on RBDC. Best viewed in color.}
\label{fig_lambda_rbdc}
\vspace{-0.2cm}
\end{figure}

\begin{figure}[!t]
\begin{center}
\includegraphics[width=0.8\linewidth]{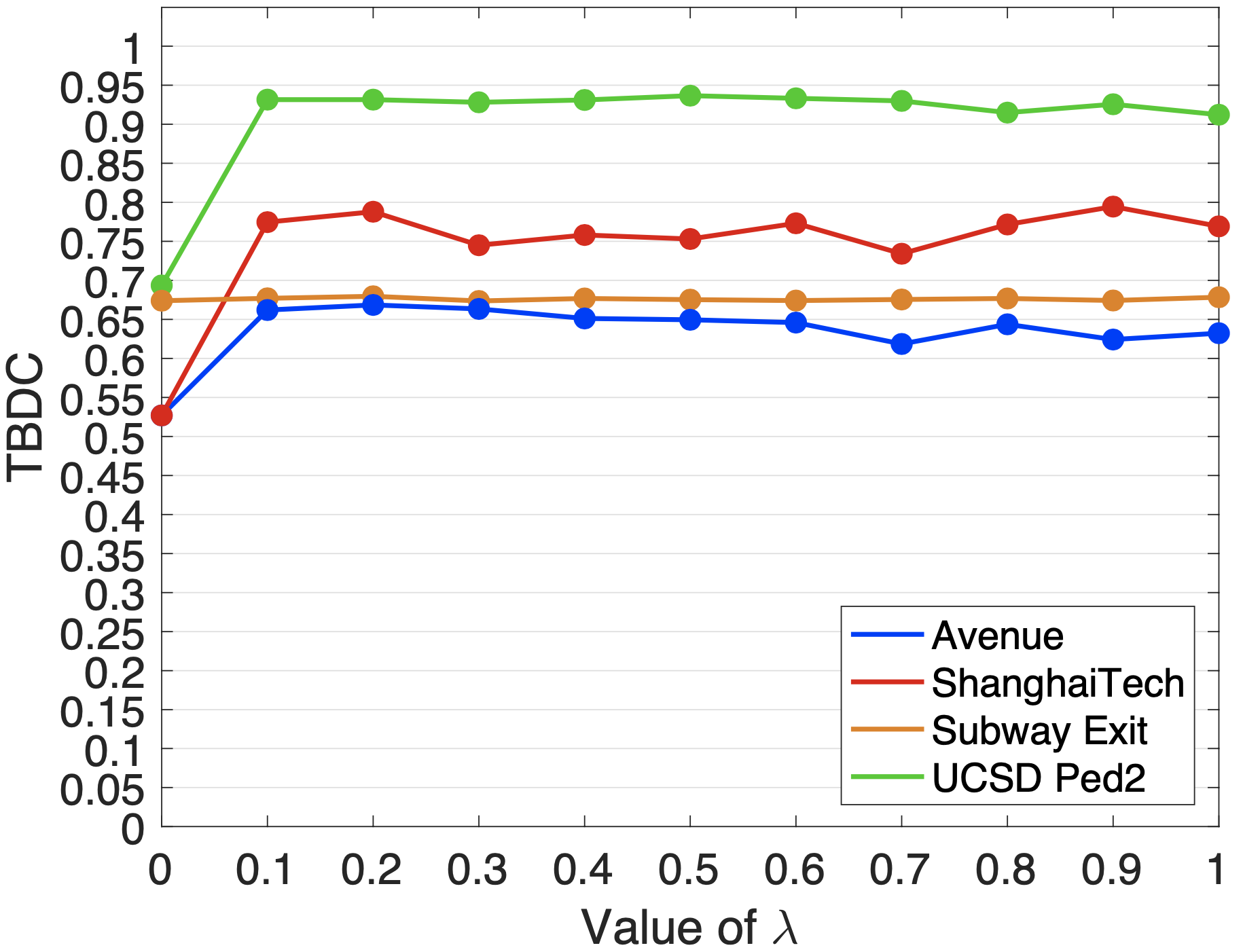}
\end{center}
\vspace{-0.5cm}
\caption{Effect of $\lambda$ (the weight of the reversed gradient) on TBDC. Best viewed in color.}
\label{fig_lambda_tbdc}
\vspace{-0.2cm}
\end{figure}

\begin{table*}[!t]
\setlength\tabcolsep{1.0pt}
\renewcommand{\arraystretch}{1.1} 
\caption{Micro-averaged AUC, macro-averaged AUC, RBDC and TBDC scores (in $\%$) with various design changes applied on our framework. Results are reported on the Avenue, ShanghaiTech, Subway Exit and UCSD Ped2 data sets.}
\label{tab_additional_ablation}
\vspace{-0.3cm}
\centering
\begin{tabular}{|l|c|c|c|c|c|c|c|c|c|c|c|c|c|c|c|c|} 
\hline 
            													& \multicolumn{4}{|c|}{Avenue}     					& \multicolumn{4}{|c|}{ShanghaiTech}					& \multicolumn{4}{|c|}{Subway Exit}						& \multicolumn{4}{|c|}{UCSD Ped2} \\
\cline{2-17}
Method            					& \multicolumn{2}{|c|}{AUC} & RBDC & TBDC & \multicolumn{2}{|c|}{AUC} & RBDC & TBDC & \multicolumn{2}{|c|}{AUC} & RBDC & TBDC & \multicolumn{2}{|c|}{AUC} & RBDC & TBDC\\
\cline{2-3}
\cline{6-7}
\cline{10-11}
\cline{14-15} 
       										& Micro	& Macro	&					&				 & Micro		& Macro	&					&				 & Old	& New	&					&				 & Micro	& Macro	&					&				\\
\hline
Proposed method         								& $92.3$	& $90.4$	& $65.05$	& $66.85$ 				& $82.7$	& $89.3$	& $41.34$	& $78.79$ 		
																	& $92.1$	& $93.7$	& $47.95$	& $67.96$ 				& $98.7$	& $99.7$	& $69.23$   & $93.15$ \\ 
\hline

$2\times$weight for appearance classifier & $92.1$	& $90.0$	& $66.77$	& $63.69$ 				& $82.3$	& $87.5$	& $42.41$	& $79.46$
																	& $92.1$	& $93.6$	& $47.82$	& $67.18$				& $98.9$	& $99.7$	& $69.06$	& $92.27$\\ 
\hline

$L_1$ loss for CAEs        								& $91.3$	& $90.0$	& $64.92$	& $64.54$ 				& $82.1$	& $88.8$	& $39.62$	& $78.11$	
																	& $91.5$	& $93.3$	& $47.70$	& $67.48$				& $99.4$	& $99.9$	& $69.36$	& $92.46$\\ 
\hline

RGB input for appearance CAE                		& $91.7$	& $90.3$	& $63.97$	& $65.48$ 				& $81.9$	& $89.5$	& $38.90$	& $78.63$	
																	& -	& -	& -	& -	& -	& -	& -	& -\\ 
\hline

Signed differences for classifiers             	& $89.5$	& $90.1$	& $61.98$	& $64.71$ 				& $81.7$	& $87.7$	& $39.73$	& $78.07$	
																	& $92.3$	& $93.3$	& $47.85$	& $67.12$				& $99.2$	& $99.6$	& $67.12$	& $92.37$\\ 
\hline
\end{tabular}
\vspace{-0.3cm}
\end{table*}

\begin{figure*}[!t]
\begin{center}
\includegraphics[width=0.6\linewidth]{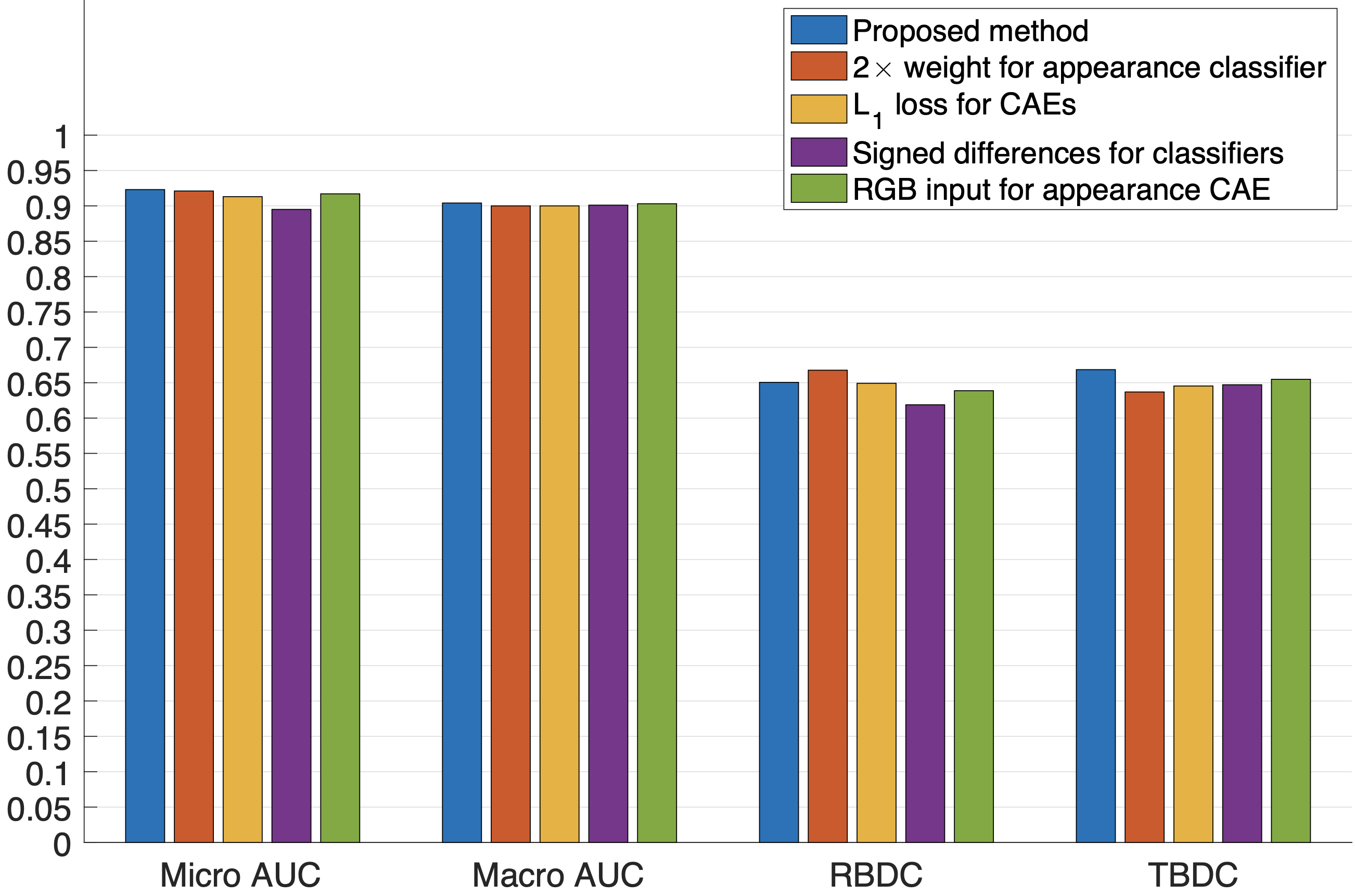}
\end{center}
\vspace{-0.5cm}
\caption{Micro-averaged AUC, macro-averaged AUC, RBDC and TBDC scores on Avenue with various design changes applied on our framework. Best viewed in color.}
\label{fig_choices_avenue}
\vspace{-0.2cm}
\end{figure*}

\begin{figure*}[!t]
\begin{center}
\includegraphics[width=0.6\linewidth]{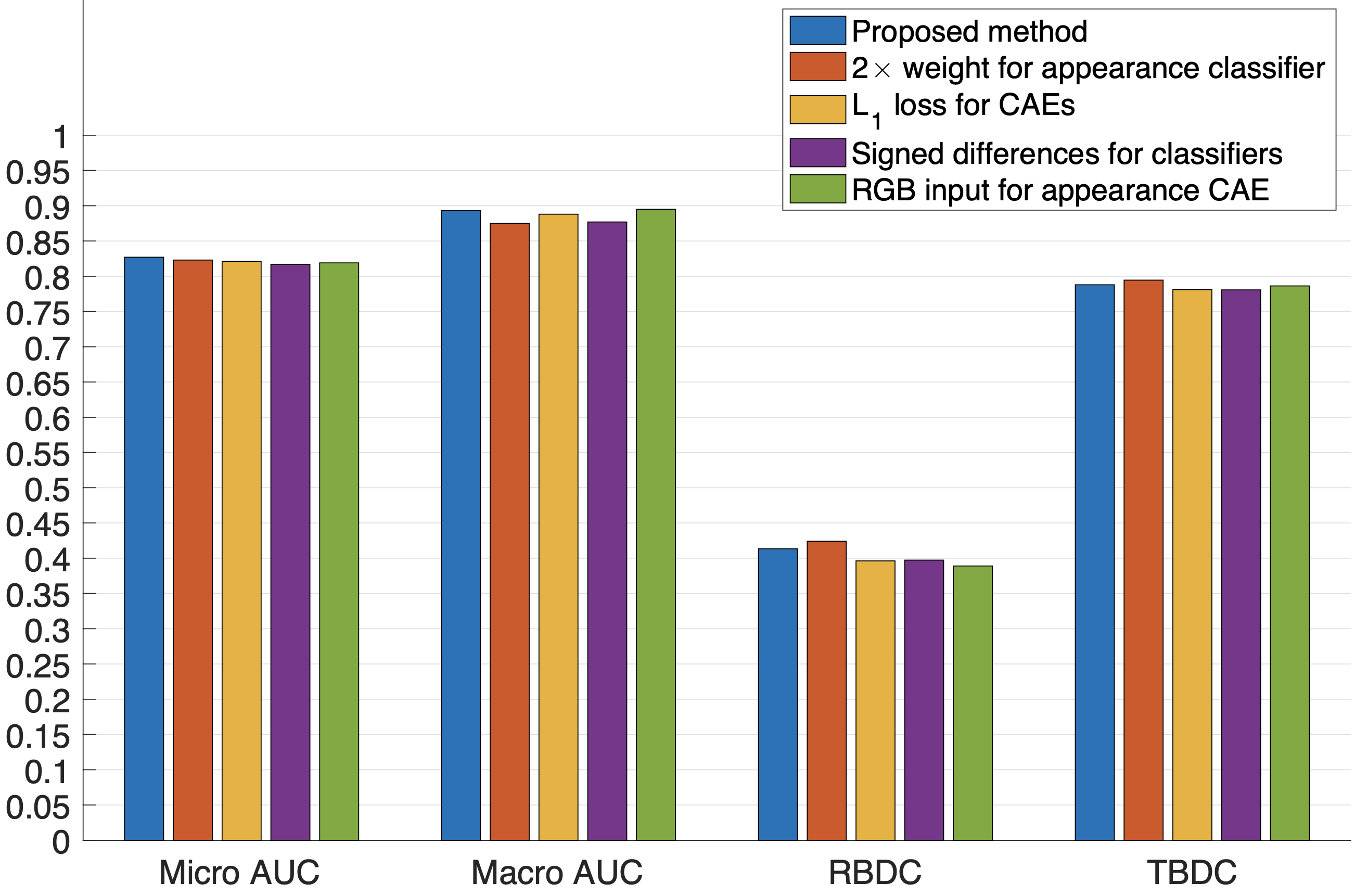}
\end{center}
\vspace{-0.5cm}
\caption{Micro-averaged AUC, macro-averaged AUC, RBDC and TBDC scores on ShanghaiTech with various design changes applied on our framework. Best viewed in color.}
\label{fig_choices_shanghaitech}
\vspace{-0.2cm}
\end{figure*}

Figure~\ref{fig_lambda_macro} shows the effect of $\lambda$ on the macro-averaged frame-level AUC. First, we note that the results on Subway Exit are the same as in Figure~\ref{fig_lambda_micro}, since the micro-averaged and the macro-averaged AUC are the same for this data set containing a single and very long test video. On UCSD Ped2, we notice that there are multiple values of $\lambda$ that produce a macro-averaged AUC of $100\%$. On ShanghaiTech, we obtain the top macro-averaged frame-level AUC of $89.9\%$ with $\lambda=0.9$, while on Avenue, $\lambda=0.1$ produces the top macro-averaged frame-level AUC of $91.3\%$.

The effect of $\lambda$ on RBDC is illustrated in Figure~\ref{fig_lambda_rbdc}. We underline that our default configuration, i.e. $\lambda=0.2$, produces optimal RBDC scores for both Avenue ($65.05\%$) and ShanghaiTech ($41.34\%$). On UCSD Ped2, we obtain the top RBDC score of $71.60\%$ with $\lambda=0.4$, while on Subway Exit, the RBDC scores are nearly constant, regardless of $\lambda$.

Figure~\ref{fig_lambda_tbdc} show the effect of $\lambda$ on TBDC. We observe that values of $\lambda\geq1$ produce fairly low TBDC fluctuations on Avenue, Subway Exit and UCSD Ped2. There are noticeable fluctuations on ShanghaiTech, where the $\lambda=0.9$ produces the highest TBDC score ($79.45\%$). We also underline that completely removing adversarial training ($\lambda=0$) produces significantly lower TBDC scores on Avenue, ShanghaiTech and UCSD Ped2.

Overall, we observe that the results are consistent across the four evaluation metrics. We conclude that the results discussed above indicate that it is better to tune the hyperparameter $\lambda$ on each data set, provided that a validation set would be available. Nevertheless, we emphasize that adversarial training brings major improvements, even when the value of $\lambda$ is not optimally chosen.

\begin{figure*}[!t]
\begin{center}
\includegraphics[width=0.6\linewidth]{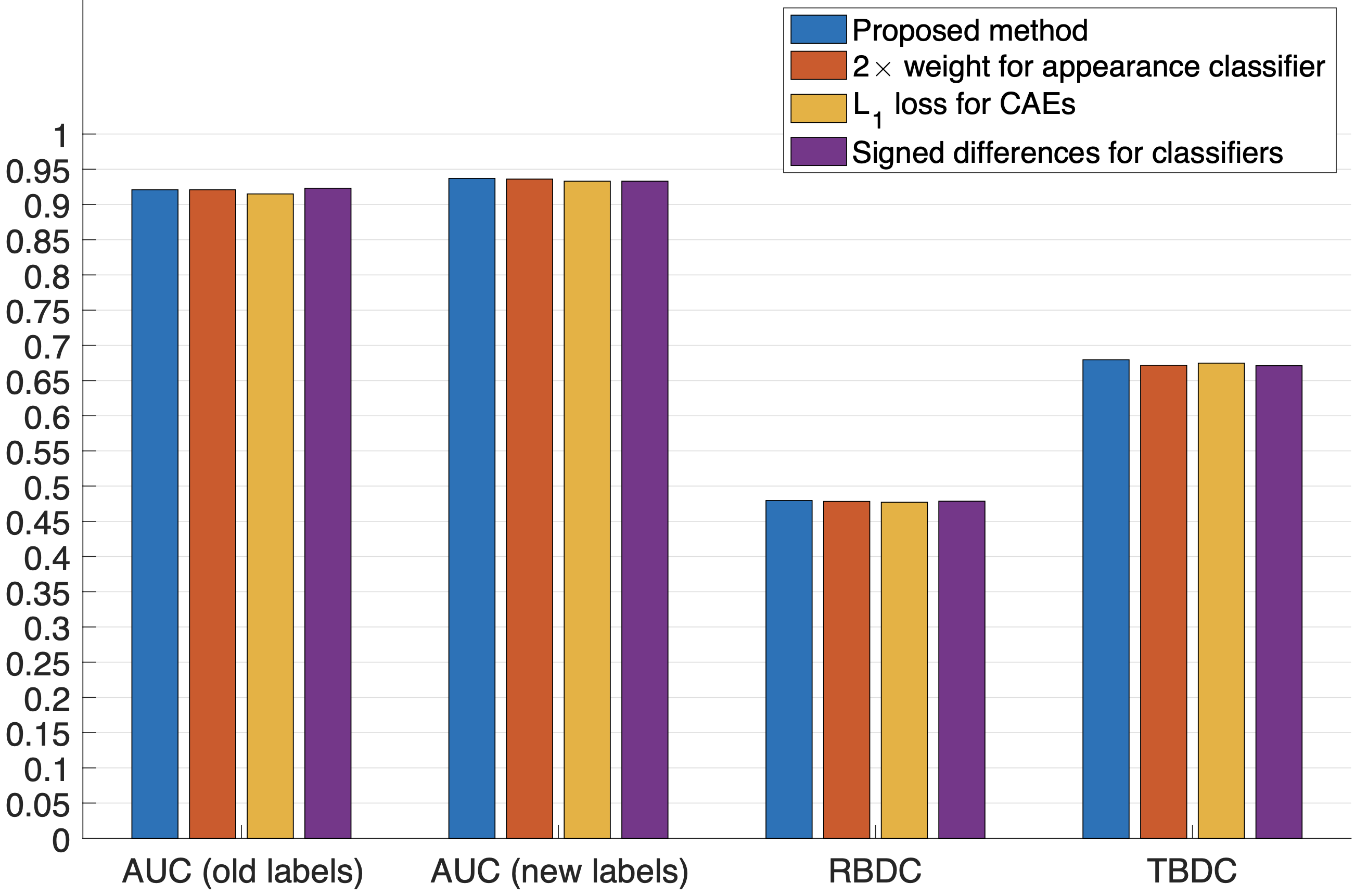}
\end{center}
\vspace{-0.5cm}
\caption{Micro-averaged AUC, macro-averaged AUC, RBDC and TBDC scores on Subway Exit with various design changes applied on our framework. Best viewed in color.}
\label{fig_choices_subway_exit}
\vspace{-0.2cm}
\end{figure*}

\begin{figure*}[!t]
\begin{center}
\includegraphics[width=0.6\linewidth]{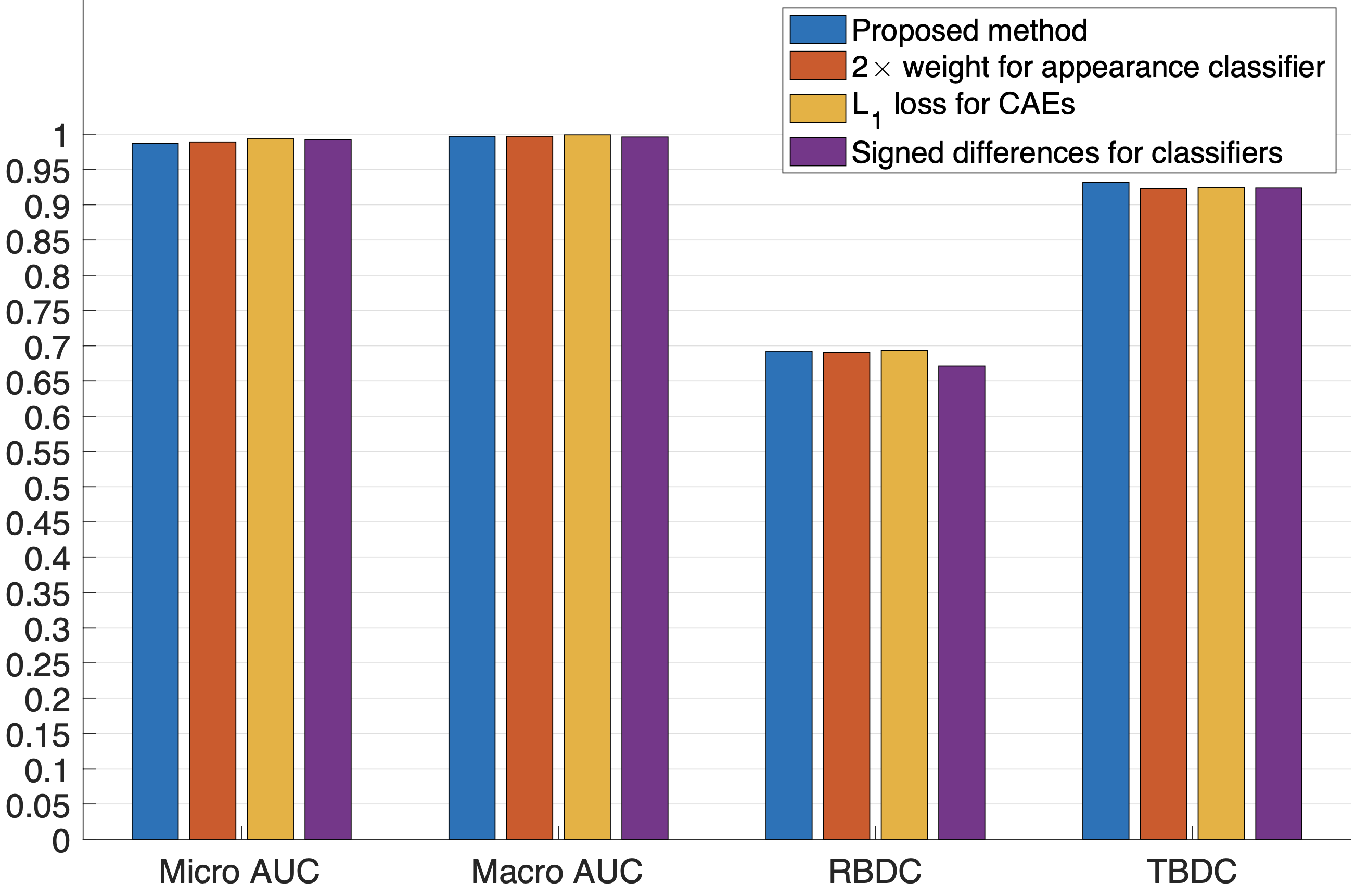}
\end{center}
\vspace{-0.5cm}
\caption{Micro-averaged AUC, macro-averaged AUC, RBDC and TBDC scores on UCSD Ped2 with various design changes applied on our framework. Best viewed in color.}
\label{fig_choices_ucsd_ped2}
\vspace{-0.2cm}
\end{figure*}

\vspace{-0.2cm}
\subsection{Testing Additional Design Choices}

In Table~\ref{tab_additional_ablation}, we present results considering additional design choices on four benchmarks: Avenue, ShanghaiTech, Subway Exit and UCSD Ped2. Another perspective of the results is provided with the bar charts illustrated in Figures~\ref{fig_choices_avenue}, \ref{fig_choices_shanghaitech}, \ref{fig_choices_subway_exit} and \ref{fig_choices_ucsd_ped2}.

In the proposed method, we computed the average of the normality scores of two motion classifiers and one appearance classifier, implicitly assigning a higher weight to motion. When we double the weight of the appearance classifier, we observe some noticeable improvements on Avenue (in terms of RBDC) and ShanghaiTech (in terms of RBDC and TBDC), the other positive or negative performance changes being minor.

In the proposed framework, we employed the $L_2$ loss to measure reconstruction errors. However, an alternative option is to use the $L_1$ loss. The results reported in Table~\ref{tab_additional_ablation} show generally lower performance levels for the $L_1$ loss on Avenue, ShanghaiTech and Subway Exit. The $L_1$ loss brings improvements only on UCSD Ped2. Overall, we believe that the $L_2$ loss is a better choice.

To keep the similar designed as in \cite{Ionescu-CVPR-2019} for our appearance CAE, we used a grayscale input. However, replacing the grayscale input with a color (RGB) input does not seem to bring any improvements for Avenue and ShanghaiTech. We underline that the videos in Subway and UCSD Ped2 are already grayscale, so it does not make sense to use auto-encoders with RGB input as this would not make any difference.

As input to our binary classifiers, we considered the absolute differences between inputs and reconstructions. Alternatively, the binary classifiers could also take the signed differences as input. We observe that the performance obtained with absolute differences is generally higher than using signed differences. 
Intuitively, for abnormal event detection, we believe that knowing if the difference is small or large is more important than knowing if the reconstruction contains pixels of lower or higher intensities than the input. Hence, the absolute difference, i.e.~the $L_1$ distance between inputs and reconstructions, is more representative.

\vspace{-0.2cm}
\subsection{Failure Cases}

\begin{figure}[!t]
\begin{center}
\includegraphics[width=1.0\linewidth]{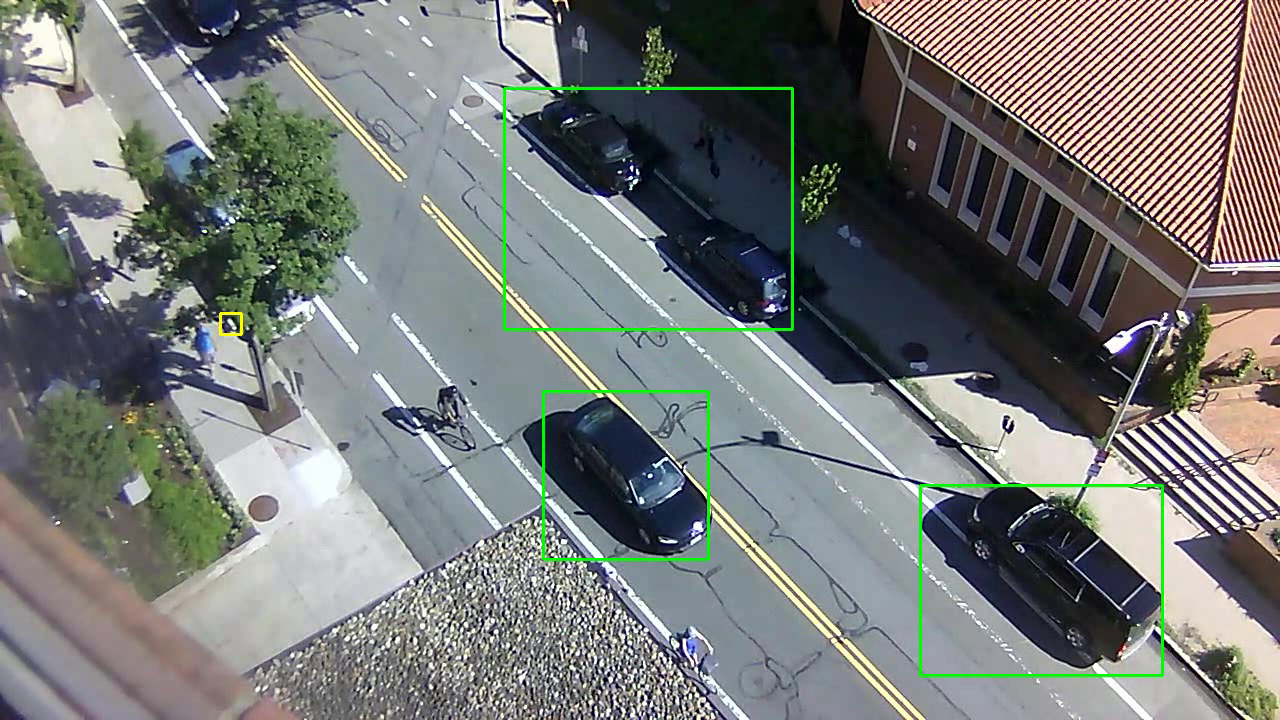}
.\vspace{-0.3cm}
\includegraphics[width=1.0\linewidth]{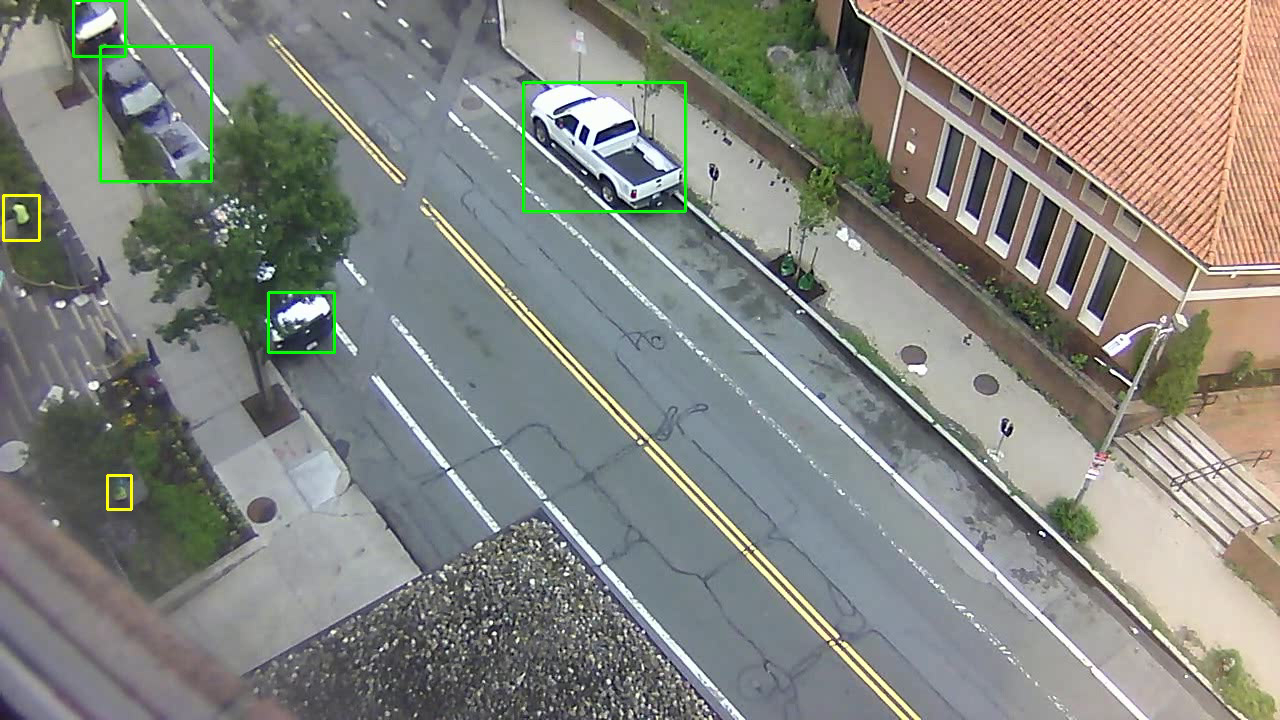}
\end{center}
\vspace{-0.5cm}
\caption{Two frames from Street Scene~\cite{Ramachandra-WACV-2020a} illustrating failure cases of our framework. Objects detected by YOLOv3 are surrounded by green bounding boxes, while abnormal objects are surrounded by yellow bounding boxes. There is no overlap between abnormal and detected objects. Best viewed in color.}
\label{fig_street_scene}
\vspace{-0.2cm}
\end{figure}

Besides Avenue, ShanghaiTech, Subway and UCSD Ped2, we also tried to apply our anomaly detection framework on the recently introduce Street Scene data set~\cite{Ramachandra-WACV-2020a}. Street Scene was relevant to us because it is a large data set and we would have been able to compare our method in terms of RBDC and TBDC with an existing work~\cite{Ramachandra-WACV-2020a}. Street Scene contains a single outdoor scene which is filmed from above, as shown in the frames illustrated in Figure~\ref{fig_street_scene}. Since the objects are relatively small and filmed from an atypical perspective, the pre-trained YOLOv3 detector fails to detect the objects of interest in most cases. In Figure~\ref{fig_street_scene}, we present two examples in which the YOLOv3 detections (surrounded by green bounding boxes) do not include the abnormal objects (surrounded by yellow bounding boxes), which are hard to see even with the naked eye. In such cases, our anomaly detection framework has no chance of detecting the anomalies. In order to demonstrate that it is impossible to surpass the state-of-the-art method~\cite{Ramachandra-WACV-2020a} due to the poor performance of the object detector, we compute the RBDC for the YOLOv3 detections obtained for the confidence level $0.5$, while assuming that the anomaly detection framework would output perfect results on the test set. In this setting, our RBDC score is $17.12\%$, which is $5\%$ under the RBDC score of $21\%$ reported in~\cite{Ramachandra-WACV-2020a}. We thus conclude that our framework fails to perform well when the objects of interest are too small or filmed in atypical perspectives. Certainly, the obvious solution is to train or fine-tune the object detector on the training video, but this would require manual labeling, which is not available for Street Scene or other video anomaly detection benchmarks.

We note that failures of the object detector can also affect our results on the other data sets. For example, on Avenue, the object detector does not detect papers or backpacks thrown in the air. Papers are not within the MS COCO object categories, while backpacks are not detected due to motion blur. However, the person performing the abnormal action (throwing papers or a backpack in the air) is detected and labeled as abnormal by our framework. This can generate false negatives when we consider the region-level or track-level metrics. However, the frame-level anomaly scores will remain unaffected in such cases.

In addition, we note that our sampling approach for pseudo-abnormal examples is based on the supposition that training and testing videos are typically collected at the same frame rate.
For example, if the test frame rate is significantly lower, it may cause normal events to be falsely reported as abnormal. This is another potential failure case, which we observed to cause performance drops in the cross-domain experiments involving the UCSD Ped2 data set as target. However, this should not represent a problem in practice, as long as we have control over the frame sampling rate.

As the chosen benchmarks do not contain location-based anomalies, we did not consider location information in our framework. To address this issue, one possible solution is to encode the location of object bounding boxes as pyramidal one-hot vectors, as proposed in~\cite{Ionescu-WACV-2019}. At a given level of the pyramid, the frame is divided into a number of spatial bins. Upon quantizing bounding boxes into bins, the one-hot encoding of a bounding box can be immediately obtained.

\vspace{-0.2cm}
\section{Conclusion}
\label{sec_Conclusion}

In this paper, we have presented a set of significant design changes to our abnormal event detection approach presented at CVPR 2019~\cite{Ionescu-CVPR-2019}. More specifically, we replaced the k-means clustering and the one-versus-rest SVM with a set of binary classifiers that learn from normal and pseudo-abnormal examples. Additionally, we modified the convolutional auto-encoders by adding adversarial and segmentation branches, as well as skip connections. Our design changes resulted in significant performance improvements in terms of both RBDC and TBDC. As a secondary contribution of our work, we release region-level and track-level annotations for the ShanghaiTech~\cite{Luo-ICCV-2017} and the Subway~\cite{Adam-PAMI-2008} data sets. Our experiments conducted on Avenue, ShanghaiTech, Subway and UCSD Ped2 indicate that our approach generally attains state-of-the-art results. We also demonstrated that our approach is background-agnostic, outperforming many approaches from the recent literature \cite{Giorno-ECCV-2016, Luo-ICCV-2017, Ramachandra-WACV-2020a, Hasan-CVPR-2016, Luo-ICCV-2017, Lee-TIP-2019, Doshi-CVPRW-2020a, Tang-PRL-2020,Dong-Access-2020, Ionescu-CVPR-2019}, when our method is tested on scenes not seen during training (unlike the related works that are trained and tested on the same scenes).

In future work, we aim to study new ways to improve the computational time of our abnormal event detection framework. More precisely, we will pursue faster object detection and optical flow estimation methods, which currently account for $95\%$ of our total processing time. 




\vspace{-0.2cm}
\ifCLASSOPTIONcompsoc
  \section*{Acknowledgments}
\else
  \section*{Acknowledgment}
\fi

The research leading to these results has received funding from the EEA Grants 2014-2021, under Project contract no. EEA-RO-NO-2018-0496. This article has also benefited from the support of the Romanian Young Academy, which is funded by Stiftung Mercator and the Alexander von Humboldt Foundation for the period 2020-2022. The work is also supported by starting grant (GR010) and VR starting grant (2016-05543).

\ifCLASSOPTIONcaptionsoff
  \newpage
\fi



%

\bibliographystyle{IEEEtran}
\bibliography{references}

\begin{IEEEbiography}[{\includegraphics[width=0.6in,height=0.78in,clip,keepaspectratio]{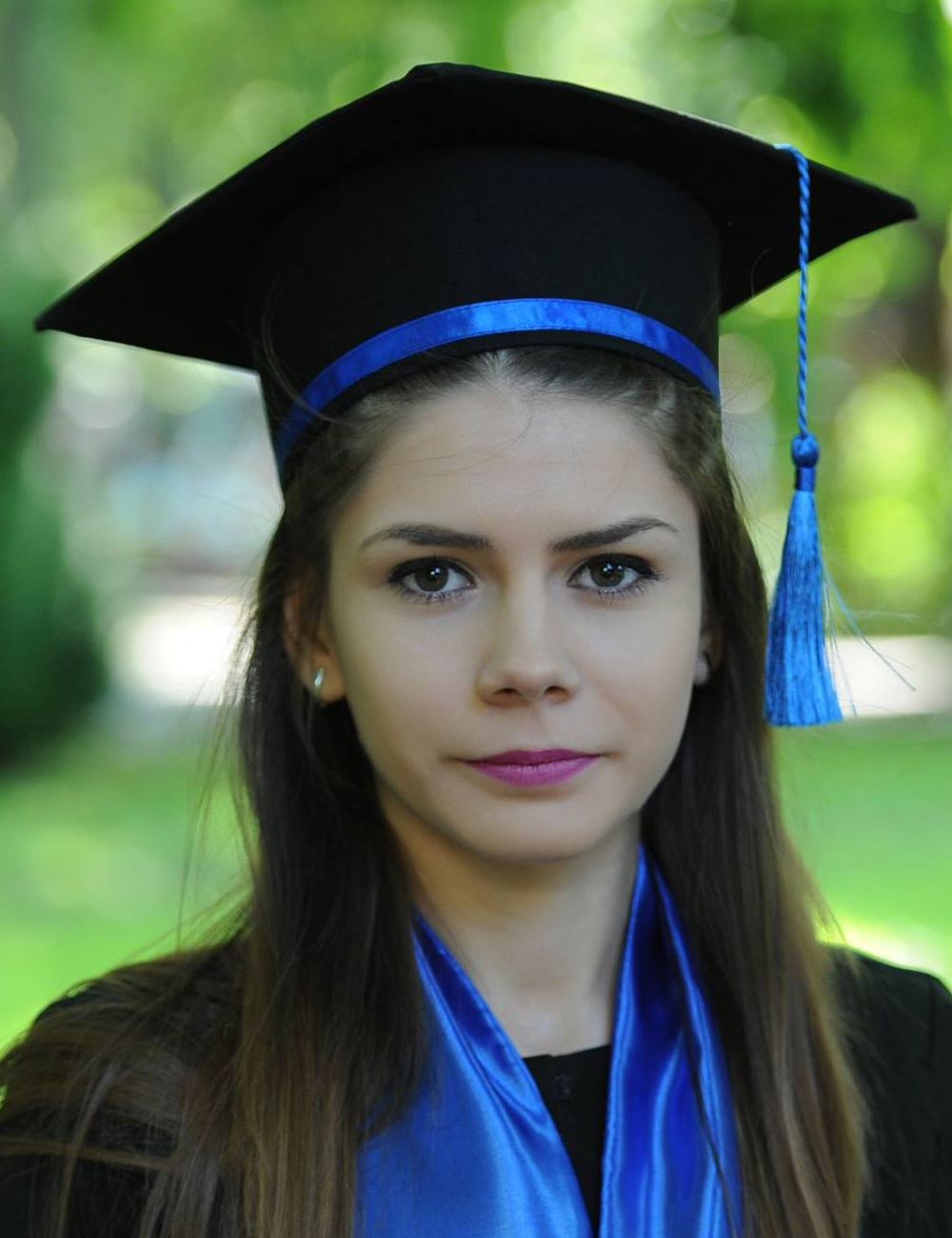}}]
{Iuliana Georgescu} is a PhD~student at the Faculty of Mathematics and Computer Science, University of Bucharest. She received the B.Sc.~degree from the Faculty of Mathematics and Computer Science, University of Bucharest, in 2017 and the M.Sc.~degree in Artificial Intelligence from the same university in 2019. Although she is early in her research career, she is the first author of 6 papers published at conferences and journals. Her research interests include artificial intelligence, computer vision, machine learning, deep learning and medical image processing.
\end{IEEEbiography}

\begin{IEEEbiography}[{\includegraphics[width=0.6in,height=0.78in,clip,keepaspectratio]{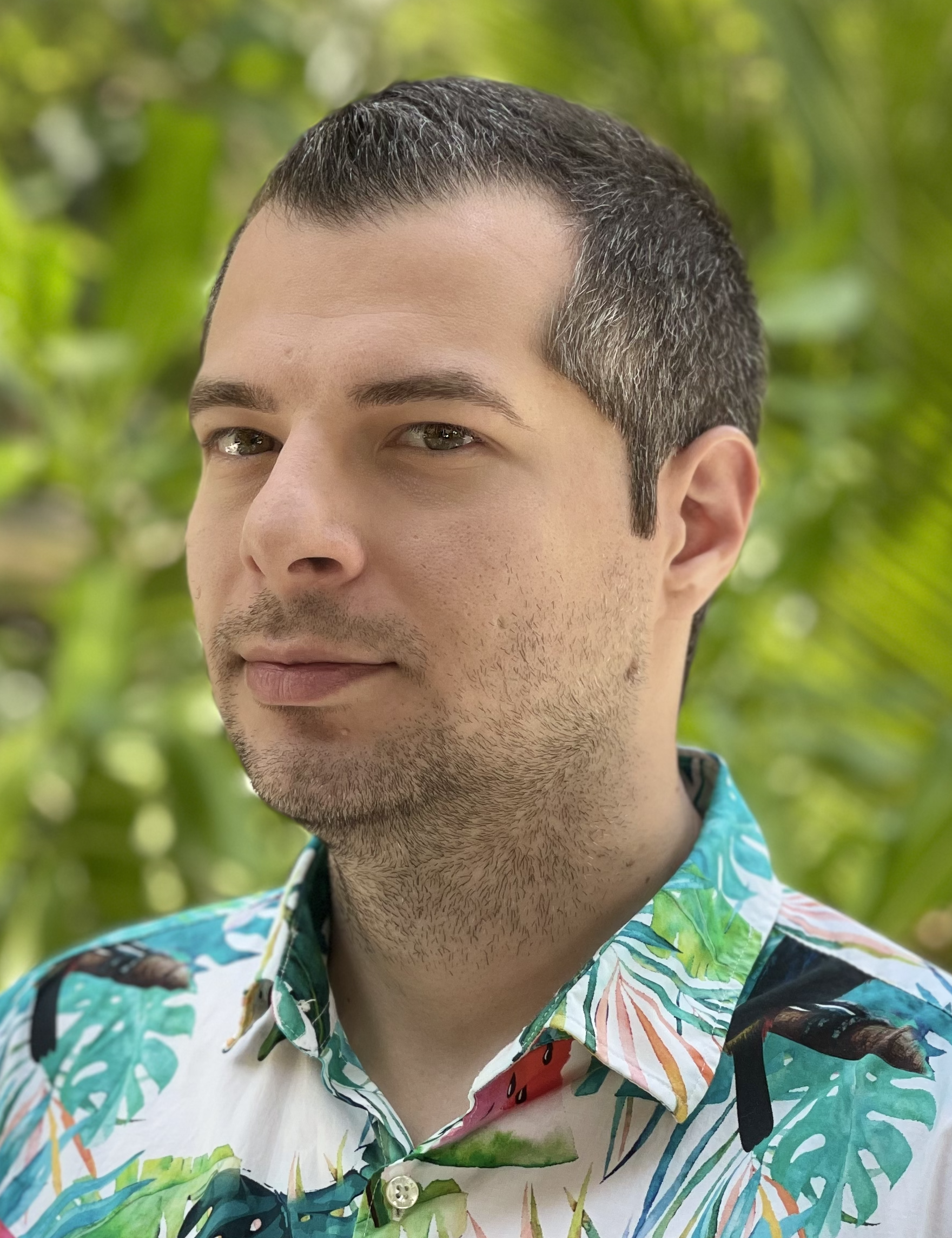}}]
{Radu Ionescu} is professor at the University of Bucharest, Romania. He completed his PhD at the University of Bucharest in 2013, receiving the 2014 Award for Outstanding Doctoral Research from the Romanian Ad Astra Association.
His research interests include machine learning, computer vision, image processing, computational linguistics and medical imaging. He published over 90 articles at international venues (including CVPR, NeurIPS, ICCV, ACL, EMNLP, NAACL), and a research monograph with Springer. Radu received the ``Caianiello Best Young Paper Award'' at ICIAP 2013. 
Radu also received the 2017 ``Young Researchers in Science and Engineering'' Prize for young Romanian researchers and the ``Danubius Young Scientist Award 2018 for Romania''. He participated at several international competitions obtaining top ranks: 4th place in the Facial Expression Recognition Challenge of WREPL 2013, 3rd place in the Native Language Identification Shared Task of BEA-8 2013, 2nd place in the Arabic Dialect Identification Shared Task of VarDial 2016, 
1st place in the Arabic Dialect Identification Shared Tasks of VarDial 2017 and 2018, 1st place in the Native Language Identification Shared Task of BEA-12 2017. Radu is also editor of the journal Mathematics.
\end{IEEEbiography}

\begin{IEEEbiography}[{\includegraphics[width=0.6in,height=0.78in,clip,keepaspectratio]{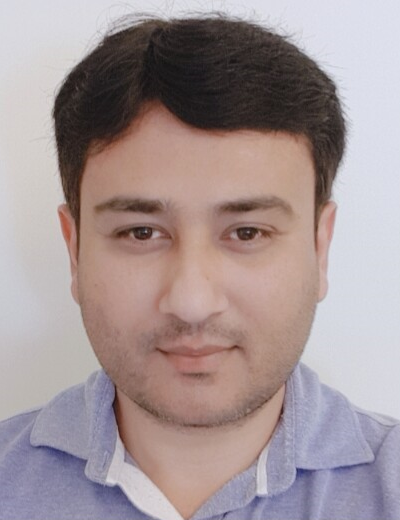}}]{Fahad Khan} is a faculty member at MBZ University of AI (MBZUAI), UAE and Link\"{o}ping University, Sweden. Prior to joining MBZUAI, he worked as a Lead Scientist at the Inception Institute of Artificial Intelligence (IIAI), UAE. He received the M.Sc.~degree in Intelligent Systems Design from Chalmers University of Technology, Sweden and a Ph.D.~degree in Computer Vision from Autonomous University of Barcelona, Spain. He has achieved top ranks on various international challenges (Visual Object Tracking VOT: 1st 2014, 2016 and 2018, 2nd 2015; VOT-TIR: 1st 2015 and 2016; OpenCV Tracking: 1st 2015; PASCAL VOC: 1st 2010) and a best paper award at ICPR 2016. His research interests include a wide range of topics within computer vision, such as object recognition, object detection, action recognition and visual tracking. He has published articles in high-impact computer vision journals and conferences in these areas.
\end{IEEEbiography}

\begin{IEEEbiography}[{\includegraphics[width=0.6in,height=0.78in,clip,keepaspectratio]{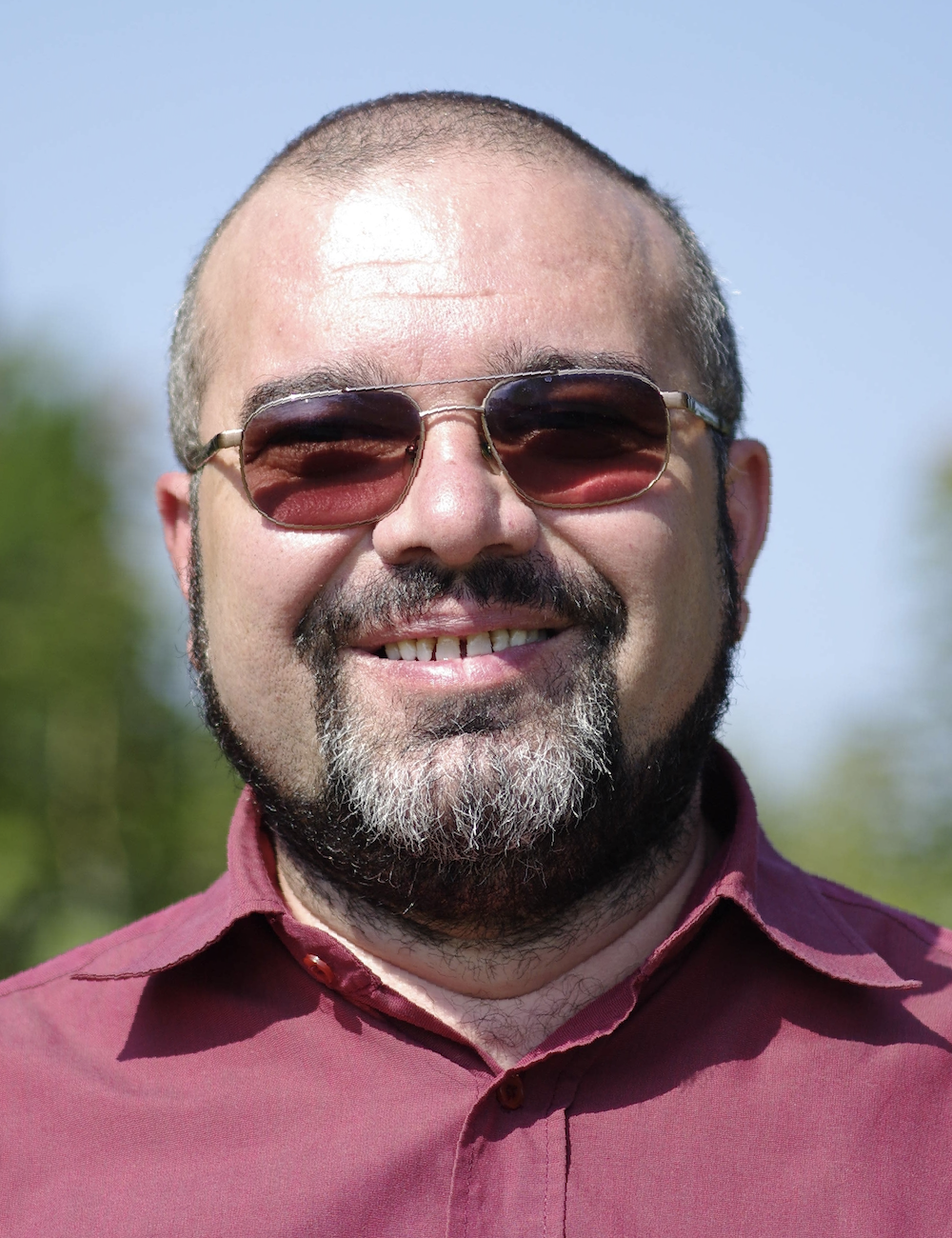}}]
{Marius Popescu} is associate professor at the University of Bucharest, Department of Computer Science. He defended his PhD in 2004 with the thesis ``Machine Learning Applied in Natural Language Processing''. 
His domains of interest are: artificial intelligence, machine learning,
computational linguistics, information retrieval, authorship identification, computer vision. His achievements in these fields include: a method for word sense disambiguation that was awarded third prize at Senseval 3 in 2004; the ENCOPLOT method for plagiarism detection that won the first international competition in plagiarism detection in 2009, followed by ranking 4th at PAN@CLEF 2010 and 2nd at PAN@CLEF 2011; 
a method for authorship analysis obtaining the best results in author identification at PAN@CLEF 2012; methods that ranked on 3rd place in the Native Language Identification Shared Task of BEA-8,
on 4th place in the Facial Expression Recognition Challenge of WREPL 2013, 2nd place in the Arabic Dialect Identification Shared Task of VarDial 2016 and 1st place in the Native Language Identification Shared Task of BEA-12.
\end{IEEEbiography}

%
\begin{IEEEbiography}[{\includegraphics[width=0.6in,height=0.78in,clip,keepaspectratio]{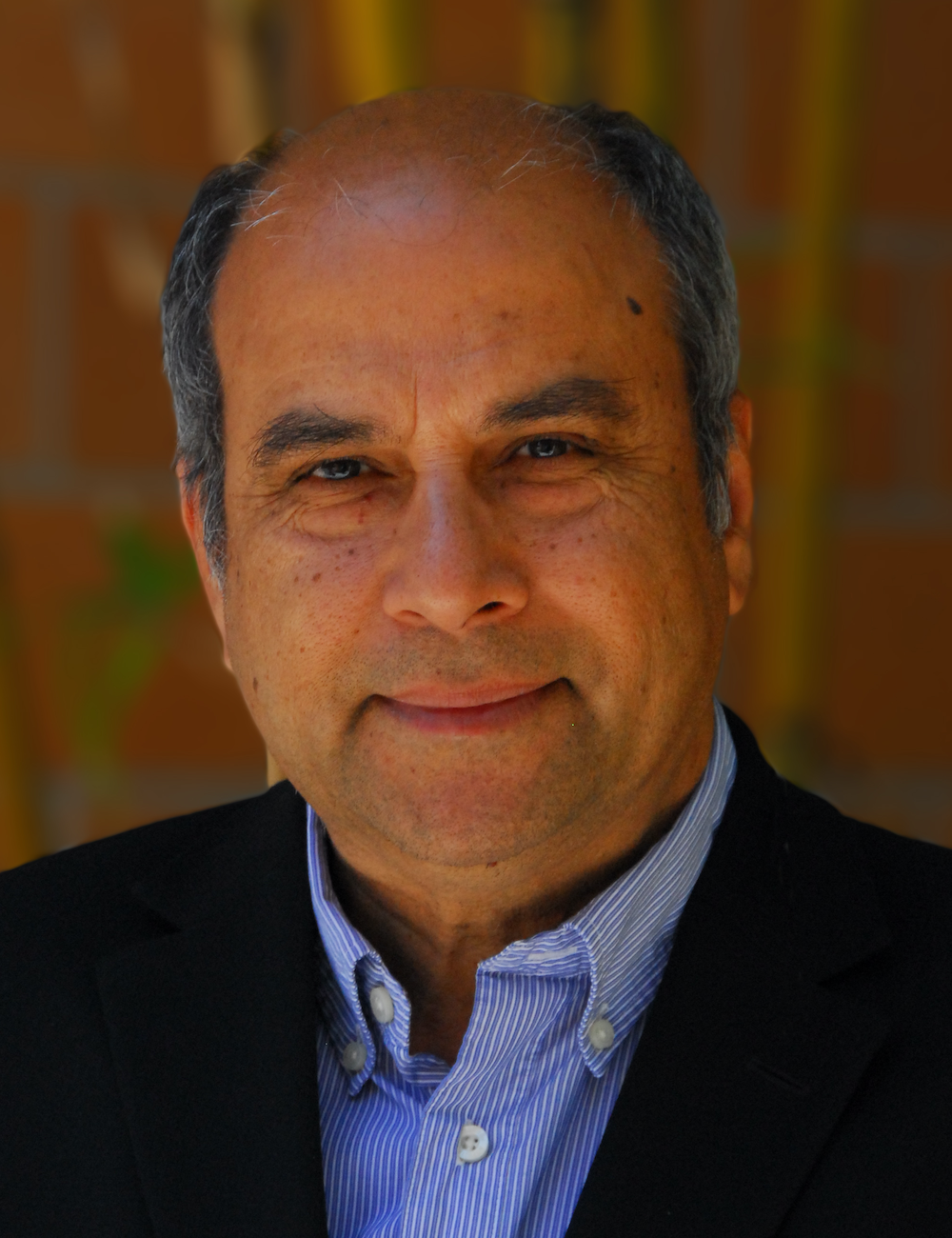}}]
{Mubarak Shah}, the UCF Trustee chair professor, is the founding director of the Center for Research in Computer Vision at the University of Central Florida (UCF). He is a fellow of the NAI, IEEE, AAAS, IAPR and SPIE. He is an editor of an international book series on video computing, was editor-in-chief of Machine Vision and Applications and an associate editor of ACM Computing Surveys and IEEE T-PAMI. He was the program cochair of CVPR 2008, an associate editor of the IEEE T-PAMI and a guest editor of the special issue of the International Journal of Computer Vision on Video Computing. 
His research interests include video surveillance, visual tracking, human activity recognition, visual analysis of crowded scenes, video registration, UAV video analysis, among others. He has served as an ACM distinguished speaker and IEEE distinguished visitor speaker. He is a recipient of ACM SIGMM Technical Achievement award; IEEE Outstanding Engineering Educator Award; Harris Corporation Engineering Achievement Award; an honorable mention for the ICCV 2005 ``Where Am I?'' Challenge Problem; 2013 NGA Best Research Poster Presentation; 2nd place in Grand Challenge at the ACM Multimedia 2013 conference; and runner up for the best paper award in ACM Multimedia Conference in 2005 and 2010. At UCF he has received Pegasus Professor Award; University Distinguished Research Award; Faculty Excellence in Mentoring Doctoral Students; Scholarship of Teaching and Learning award; Teaching Incentive Program award; Research Incentive Award.
\end{IEEEbiography}

\end{document}